\documentclass[11pt, a4paper, logo, copyright]{googledeepmind}

\usepackage[authoryear, sort&compress, round]{natbib}
\usepackage{defs}
\usepackage[capitalise]{cleveref}
\usepackage{fancybox}
\usepackage{makecell}
\usepackage{subcaption}
\usepackage{appendix}

\title{
AtP$^*$: An efficient and scalable method for localizing
LLM behaviour to components
}

\correspondingauthor{janosk@google.com}

\renewcommand{\today}{2024-02-23}

\author{J\'anos Kram\'ar}
\author{Tom Lieberum}
\author{Rohin Shah}
\author{Neel Nanda}

\affil{Google DeepMind}

\begin{abstract}

Activation Patching is a method of directly computing causal attributions of behavior to model components.
However, applying it exhaustively requires a sweep with cost scaling linearly in the number of model components, which can be prohibitively expensive for SoTA Large Language Models (LLMs). 
We investigate Attribution Patching (AtP) \citep{neel2022attribution}, a fast gradient-based approximation to Activation Patching and find two classes of failure modes of AtP which lead to significant false negatives.\\
We propose a variant of AtP called AtP$^*$, with two changes to address these failure modes while retaining scalability.
We present the first systematic study of AtP and alternative methods for faster activation patching and show that AtP significantly outperforms all other investigated methods, with AtP$^*$ providing further significant improvement.
Finally, we provide a method to bound the probability of remaining false negatives of AtP$^*$ estimates.

\end{abstract}

\begin{document}

\maketitle

\section{Introduction}

As LLMs become ubiquitous and integrated into numerous digital applications, it's an increasingly pressing research problem to understand the internal mechanisms that underlie their behaviour -- this is the problem of mechanistic interpretability. A fundamental subproblem is to causally attribute particular behaviours to individual parts of the transformer forward pass, corresponding to specific components (such as attention heads, neurons, layer contributions, or residual streams), often at specific positions in the input token sequence. This is important because in numerous case studies of complex behaviours, they are found to be driven by sparse subgraphs within the model~\citep{olsson2022context,wang2022interpretability,meng2023locating}.

A classic form of causal attribution uses zero-ablation, or knock-out, where a component is deleted and we see if this negatively affects a model's output -- a negative effect implies the component was causally important. More recent work has generalised this to replacing a component's activations with samples from some baseline distribution (with zero-ablation being a special case where activations are resampled to be zero). We focus on the popular and widely used method of Activation Patching (also known as causal mediation analysis)~\citep{geiger2022inducing,meng2023locating,chan2022causal} where the baseline distribution is a component's activations on some corrupted input, such as an alternate string with a different answer~\citep{pearl2001direct,Robins1992IdentifiabilityAE}.

Given a causal attribution method, it is common to sweep across all model components, directly evaluating the effect of intervening on each of them via resampling~\citep{meng2023locating}. However, when working with SoTA models it can be expensive to attribute behaviour especially to small components (e.g. heads or neurons) -- each intervention requires a separate forward pass, and so the number of forward passes can easily climb into the millions or billions. For example, on a prompt of length 1024, there are $2.7\cdot10^9$ neuron nodes in Chinchilla 70B~\citep{hoffmann2022chinchilla}.

We propose to accelerate this process by using Attribution Patching (AtP) \citep{neel2022attribution}, a faster, approximate, causal attribution method, as a prefiltering step: after running AtP, we iterate through the nodes in decreasing order of absolute value of the AtP estimate, then use Activation Patching to more reliably evaluate these nodes and filter out false positives -- we call this \emph{verification}. We typically care about a small set of top contributing nodes, so verification is far cheaper than iterating over all nodes.

\paragraph{Our contributions:}
\begin{itemize}
    \item We investigate the performance of AtP, finding two classes of failure modes which produce false negatives. We propose a variant of AtP called AtP$^*$, with two changes to address these failure modes while retaining scalability:
    \begin{itemize}
        \item When patching queries and keys, recomputing the attention softmax and using a gradient based approximation from then on, as gradients are a poor approximation to saturated attention.
        \item Using dropout on the backwards pass to fix brittle false negatives, where significant positive and negative effects cancel out.
\end{itemize}
    \item We introduce several alternative methods to approximate Activation Patching as baselines to AtP which outperform brute force Activation Patching.
    \item We present the first systematic study of AtP and these alternatives and show that AtP significantly outperforms all other investigated methods, with AtP$^*$ providing further significant improvement.
    \item To estimate the residual error of AtP$^*$ and statistically bound the sizes of any remaining false negatives we provide a diagnostic method, based on using AtP to filter out high impact nodes, and then patching random subsets of the remainder. Good diagnostics mean that practitioners may still gauge whether AtP is reliable in relevant domains without the costs of exhaustive verification. 
\end{itemize}

Finally, we provide some guidance in \Cref{sec:recommend} on how to successfully perform causal attribution in practice and what attribution methods are likely to be useful and under what circumstances.

\begin{figure}
    \centering
    \begin{subfigure}[b]{0.48\textwidth}
        \centering
        \includegraphics[width=\textwidth]{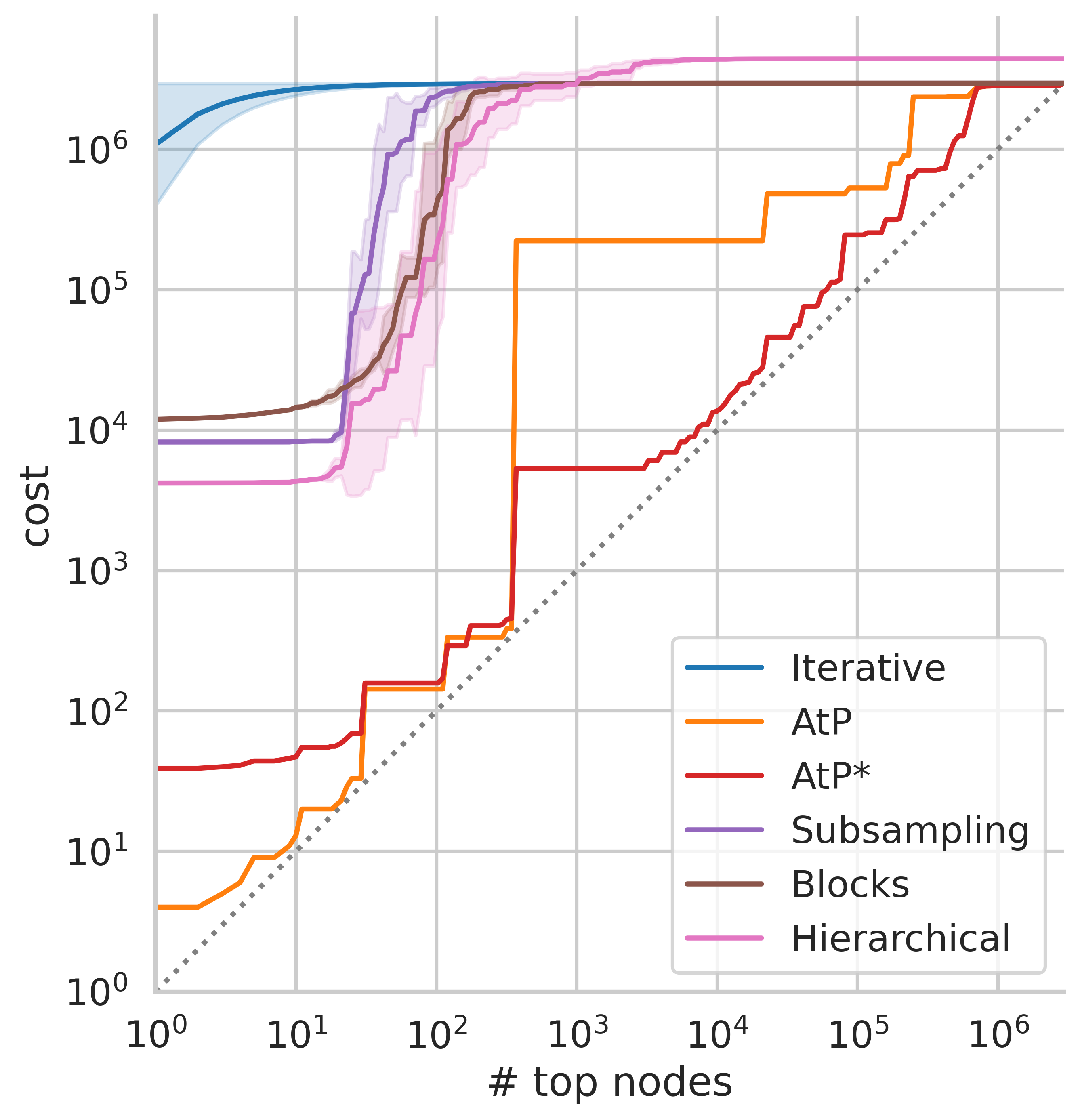}
        \caption{MLP neurons, on \texttt{CITY-PP}.}
    \end{subfigure}
    \hfill
    \begin{subfigure}[b]{0.48\textwidth}
        \centering
        \includegraphics[width=\textwidth]{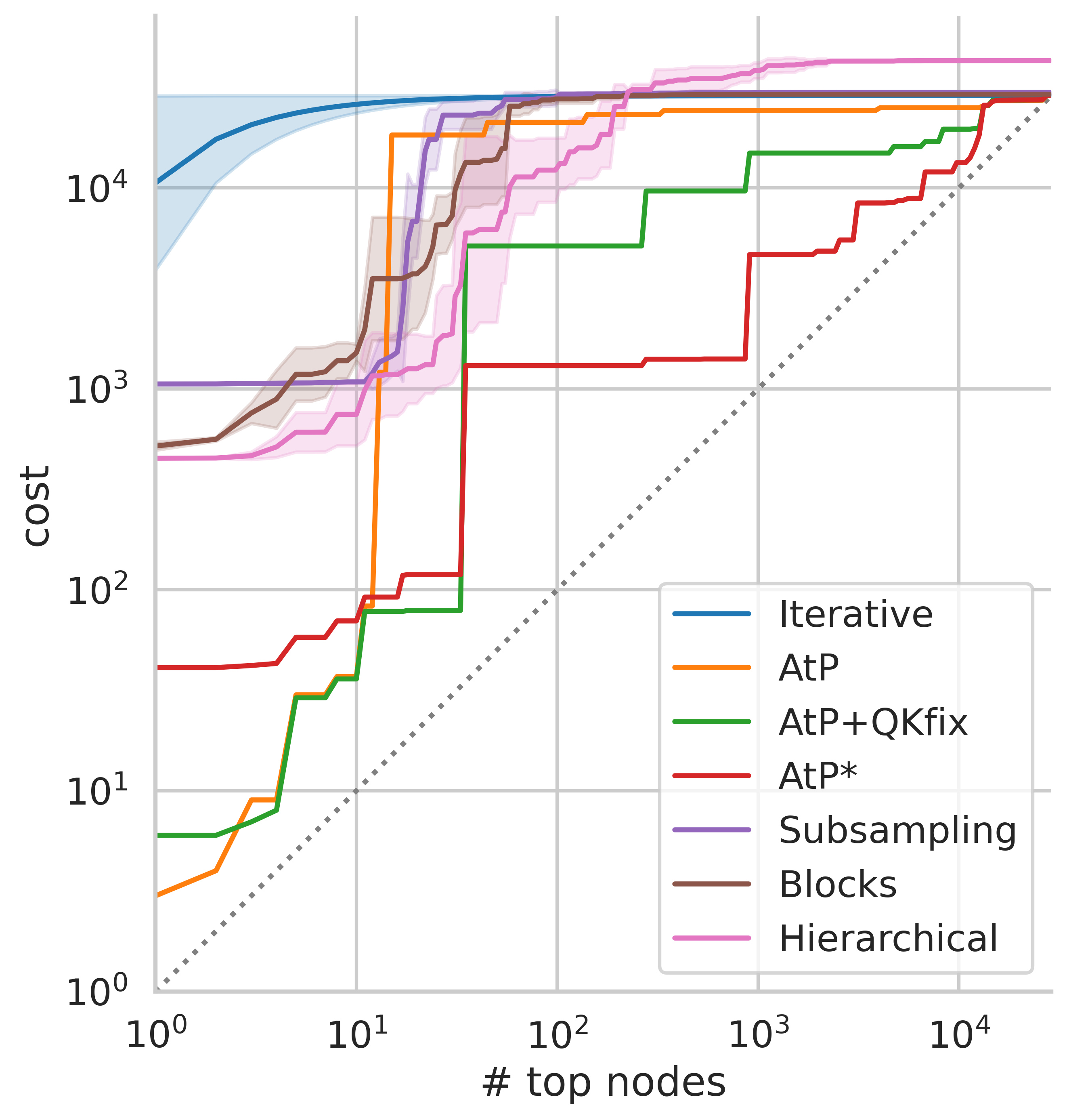}
        \caption{Attention nodes, on \texttt{IOI-PP}.}
    \end{subfigure}
    \caption{Costs of finding the most causally-important nodes in Pythia-12B using different methods, on sample prompt pairs (see \Cref{tab:single_prompt_pairs}). The shading indicates geometric standard deviation. Cost is measured in forward passes, thus each point's y-coordinate gives the number of forward passes required to find the top $x$ nodes. Note that each node must be verified, thus $y\geq x$, so all lines are above the diagonal, and an oracle for the verification order would produce the diagonal line. For a detailed description see~\Cref{sec:clean_prompt_pairs}.
    }
    \label{fig:main_result}
\end{figure}
\begin{figure}
    \centering
    \begin{subfigure}[b]{0.42\textwidth}
        \centering
        \includegraphics[width=\textwidth]{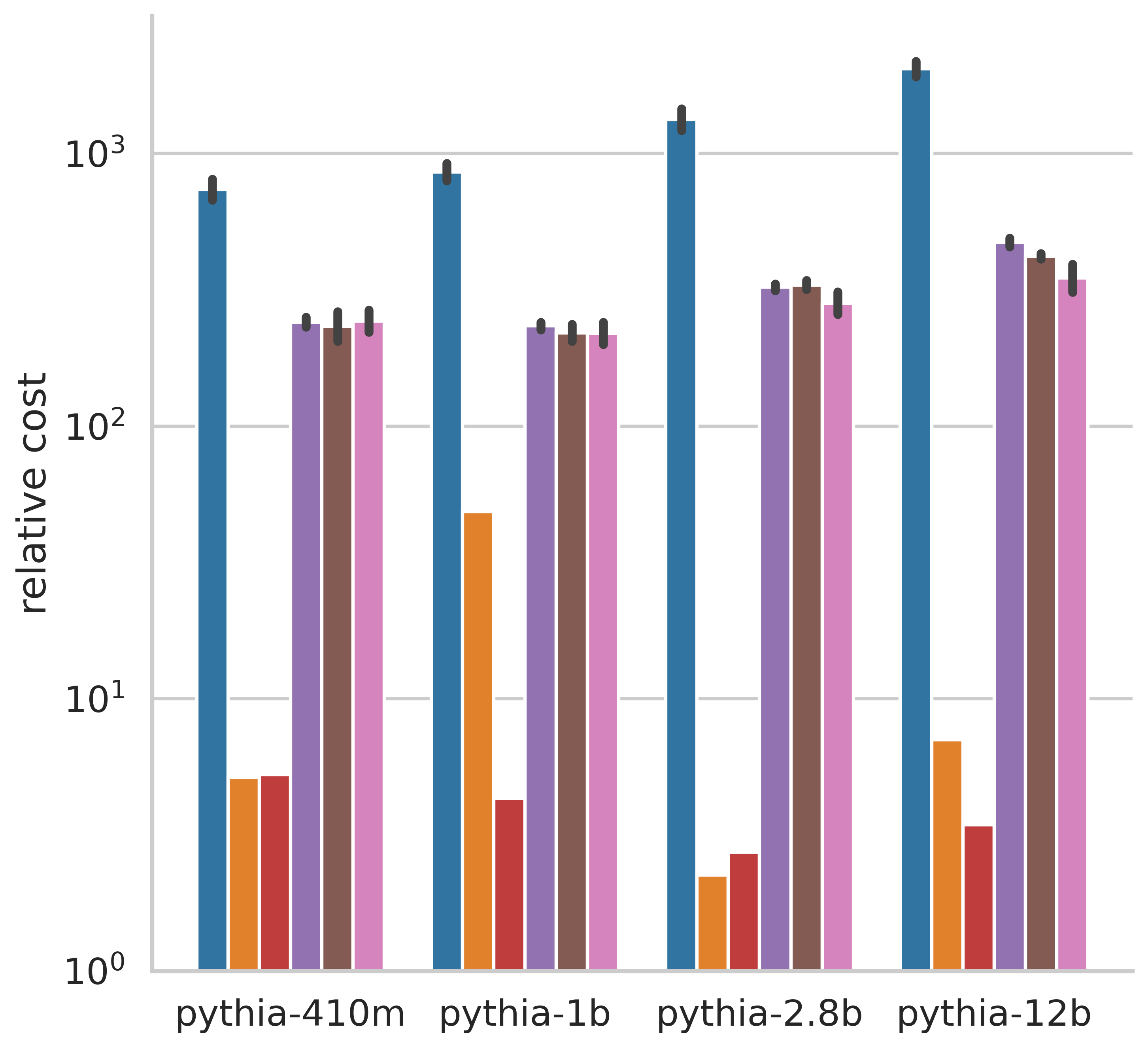}
        \caption{MLP neurons, on \texttt{CITY-PP}.}
    \end{subfigure}
    \hfill
    \begin{subfigure}[t]{0.14\textwidth}
        \centering
        \raisebox{4cm}{
        \includegraphics[width=\textwidth]{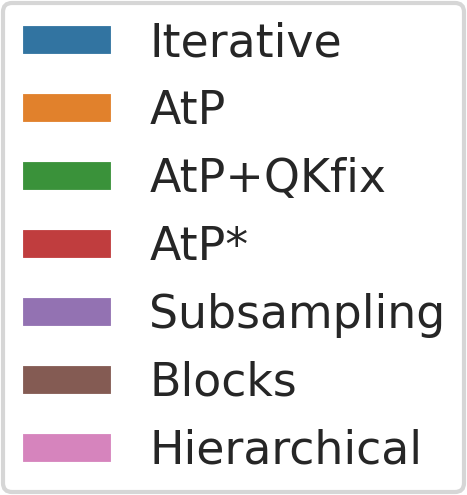}
        }
    \end{subfigure}
    \hfill
    \begin{subfigure}[b]{0.42\textwidth}
        \centering
        \includegraphics[width=\textwidth]{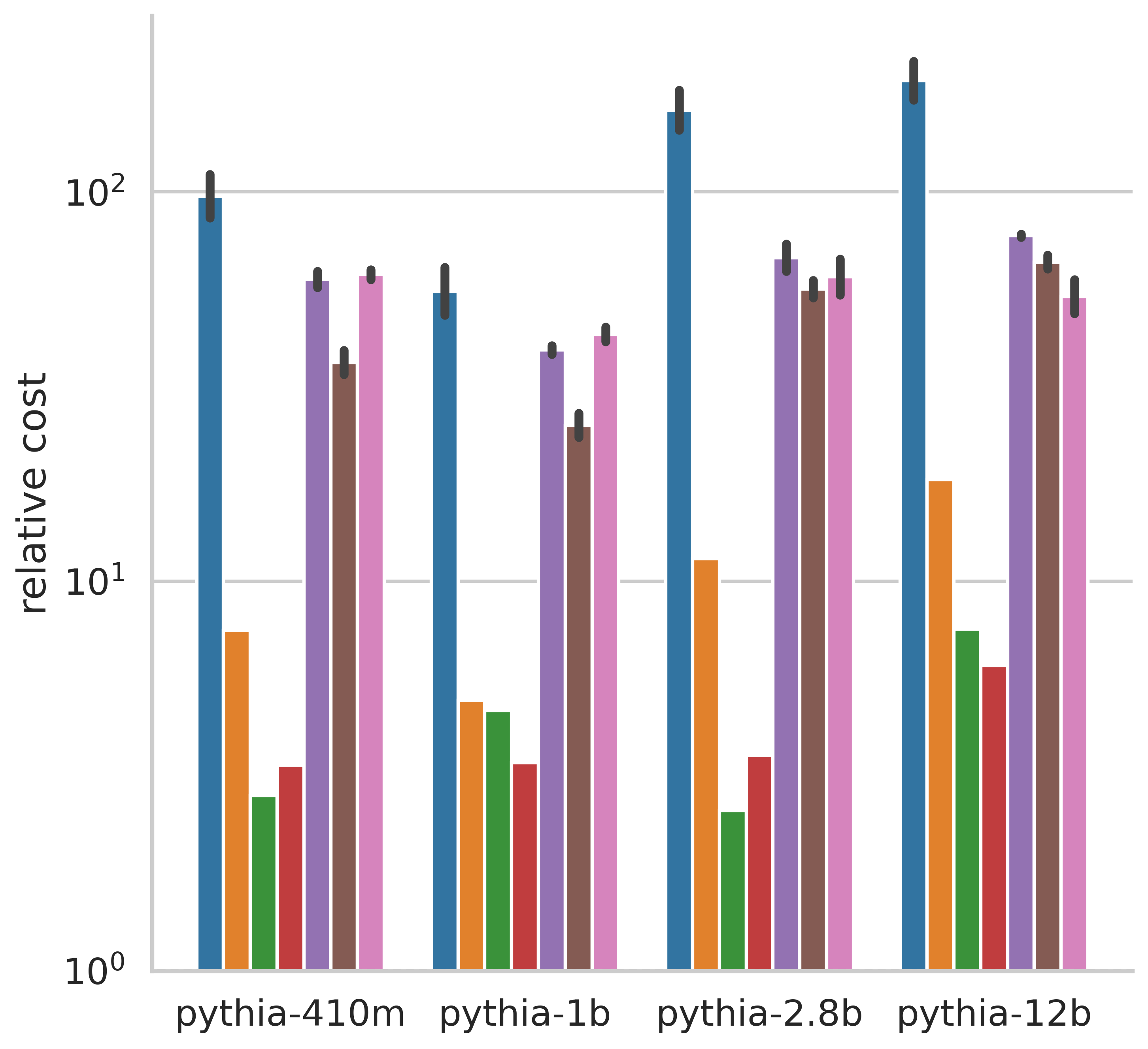}
        \caption{Attention nodes, on \texttt{IOI-PP}.}
    \end{subfigure}
    \caption{Relative costs of methods across models, on sample prompt pairs. The costs are relative to having an oracle, which would verify nodes in decreasing order of true contribution size. Costs are aggregated using an inverse-rank-weighted geometric mean. This means they correspond to the area above the diagonal for each curve in~\Cref{fig:main_result} and are relative to the area under the dotted (oracle) line. See~\Cref{sec:irwrgm} for more details on this metric. Note that GradDrop (difference between AtP+QKfix and AtP$^*$) comes with a noticeable upfront cost and so looks worse in this comparison while still helping avoid false negatives as shown in\Cref{fig:main_result}.
    }
    \label{fig:model_sweep}
\end{figure}

\section{Background} \label{sec:background}

\subsection{Problem Statement} \label{sec:problem-statement}
Our goal is to identify the contributions to model behavior by individual model components. We first formalize model components, then formalize model behaviour, and finally state the contribution problem in causal language. While we state the formalism in terms of a decoder-only transformer language model~\citep{vaswani2017attention,radford2018improving}, and conduct all our experiments on models of that class, the formalism is also straightforwardly applicable to other model classes.

\paragraph{Model components.} We are given a model $\model : \Prompts \rightarrow \reals^{\vocabSize}$ that maps a prompt (token sequence) $\prompt \in \Prompts := \{1,\ldots,V\}^T$ to output logits over a set of $\vocabSize$ tokens, aiming to predict the next token in the sequence. We will view the model $\model$ as a computational graph $(\Nodes, \Edges)$ where the node set $\Nodes$ is the set of model components, and a directed edge $\edge = (\node_1, \node_2) \in \Edges$ is present iff the output of $\node_1$ is a direct input into the computation of $\node_2$. We will use $\node(\prompt)$ to represent the \emph{activation} (intermediate computation result) of $\node$ when computing $\model(\prompt)$.

The choice of $\Nodes$ determines how fine-grained the attribution will be. For example, for transformer models, we could have a relatively coarse-grained attribution where each layer is considered a single node. In this paper we will primarily consider more fine-grained attributions that are more expensive to compute (see \Cref{sec:experiments} for details); we revisit this issue in \Cref{sec:discuss}.

\paragraph{Model behaviour.} 
Following past work~\citep{geiger2022inducing,chan2022causal,wang2022interpretability}, we assume a distribution $\promptPairDist$ over pairs of inputs $\xclean, \xnoise$, where $\xclean$ is a prompt on which the behaviour occurs, and $\xnoise$ is a reference prompt which we use as a source of noise to intervene with\footnote{This precludes interventions which use activation values that are never actually realized, such as zero-ablation or mean ablation. An alternative formulation via distributions of activation values is also possible.}.
We are also given a metric\footnote{Common metrics in language models are next token prediction loss, difference in log prob between a correct and incorrect next token, probability of the correct next token, etc.} $\metric : \reals^{\vocabSize} \rightarrow \reals$, which quantifies the behaviour of interest.

\paragraph{Contribution of a component.} Similarly to the work referenced above we define the contribution $\contrib(n)$ of a node $n$ to the model's behaviour as the counterfactual absolute\footnote{The sign of the impact may be of interest, but in this work we'll focus on the magnitude, as a measure of causal importance.} expected impact of replacing that node on the clean prompt with its value on the reference prompt $\xnoise$.

Using do-calculus notation~\citep{pearl2000causality} this can be expressed as $\contrib(\node):=|\inter(\node)|$, where
\begin{align}
\label{eq:ground_truth}
    \inter(\node) &:= \expect{(\xclean, \xnoise)\sim\promptPairDist}{\inter(\node; \xclean, \xnoise)},
\end{align}

\noindent where we define the intervention effect $\inter$ for $\xclean,\xnoise$ as

\begin{align}
\label{eq:inter}
    \inter(\node; \xclean, \xnoise) &:= \metric(\model(\xclean \mid \doGets{n}{n(\xnoise)})) - \metric(\model(\xclean)).
\end{align}

Note that the need to average the effect across a distribution adds a potentially large multiplicative factor to the cost of computing $\contrib(\node)$, further motivating this work.

We can also intervene on a set of nodes $\subsetNodes = \{\node_i\}$. To do so, we overwrite the values of all nodes in $\subsetNodes$ with their values from a reference prompt. Abusing notation, we write $\subsetNodes(x)$ as the set of activations of the nodes in $\subsetNodes$, when computing $\model(x)$.

\begin{align}
    \inter(\subsetNodes; \xclean, \xnoise) &:= \metric(\model(\xclean \mid \doGets{\subsetNodes}{\subsetNodes(\xnoise)})) - \metric(\model(\xclean))
\end{align}

We note that it is also valid to define contribution as the expected impact of replacing a node on the reference prompt with its value on the clean prompt, also known as denoising or knock-in. We follow~\citet{chan2022causal,wang2022interpretability} in using noising, however denoising is also widely used in the literature~\citep{meng2023locating,lieberum2023does}. We briefly consider how this choice affects AtP in~\Cref{sec:extensions}.

\subsection{Attribution Patching}
On state of the art models, computing $\contrib(n)$ for all $n$ can be prohibitively expensive as there may be billions or more nodes. Furthermore, to compute this value precisely requires evaluating it on all prompt pairs, thus the runtime cost of~\Cref{eq:ground_truth} for each $n$ scales with the size of the support of $\promptPairDist$. 

We thus turn to a fast approximation of~\Cref{eq:ground_truth}. As suggested by~\citet{neel2022attribution,figurnov2016perforatedcnns,molchanov2017pruning}, we can make a first-order Taylor expansion to $\inter(\node; \xclean, \xnoise)$ around $\node(\xnoise)\approx \node(\xclean)$:

\begin{align}\label{eq:inter_atp}
    \hat\inter_{\text{AtP}}(\node;\xclean,\xnoise)&:=(\node(\xnoise) - \node(\xclean))^\intercal \frac{\partial\metric(\model(\xclean))}{\partial \node}\Big{|}_{\node=\node(\xclean)}
\end{align}

Then, similarly to \citet{syed2023attribution}, we apply this to a distribution by taking the absolute value inside the expectation in \Cref{eq:ground_truth} rather than outside; this decreases the chance that estimates across prompt pairs with positive and negative effects might erroneously lead to a significantly smaller estimate. (We briefly explore the amount of cancellation behaviour in the true effect distribution in \Cref{app:distribution_cancellation}.) As a result, we get an estimate

\begin{align}\label{eq:contrib_atp}
\hat\contrib_{\text{AtP}}(\node) &:= \expect{\xclean, \xnoise}{\left|\hat\inter_{\text{AtP}}(\node;\xclean,\xnoise)\right|}.
\end{align}

This procedure is also called \emph{Attribution Patching}~\citep{neel2022attribution} or \emph{AtP}. AtP requires two forward passes and one backward pass to compute an estimate score for \emph{all nodes} on a given prompt pair, and so provides a very significant speedup over brute force activation patching.

\section{Methods} \label{sec:methods}

We now describe some failure modes of AtP and address them, yielding an improved method AtP*. We then discuss some alternative methods for estimating $\contrib(\node)$, to put AtP(*)'s performance in context. Finally we discuss how to combine Subsampling, one such alternative method described in~\Cref{sec:baselines}, and AtP* to give a diagnostic to statistically test whether AtP* may have missed important false negatives.

\subsection{AtP improvements}\label{sec:improvements}

We identify two common classes of false negatives occurring when using AtP. 

\begin{figure}[t]
    \centering
    \includegraphics[width=0.7\textwidth]{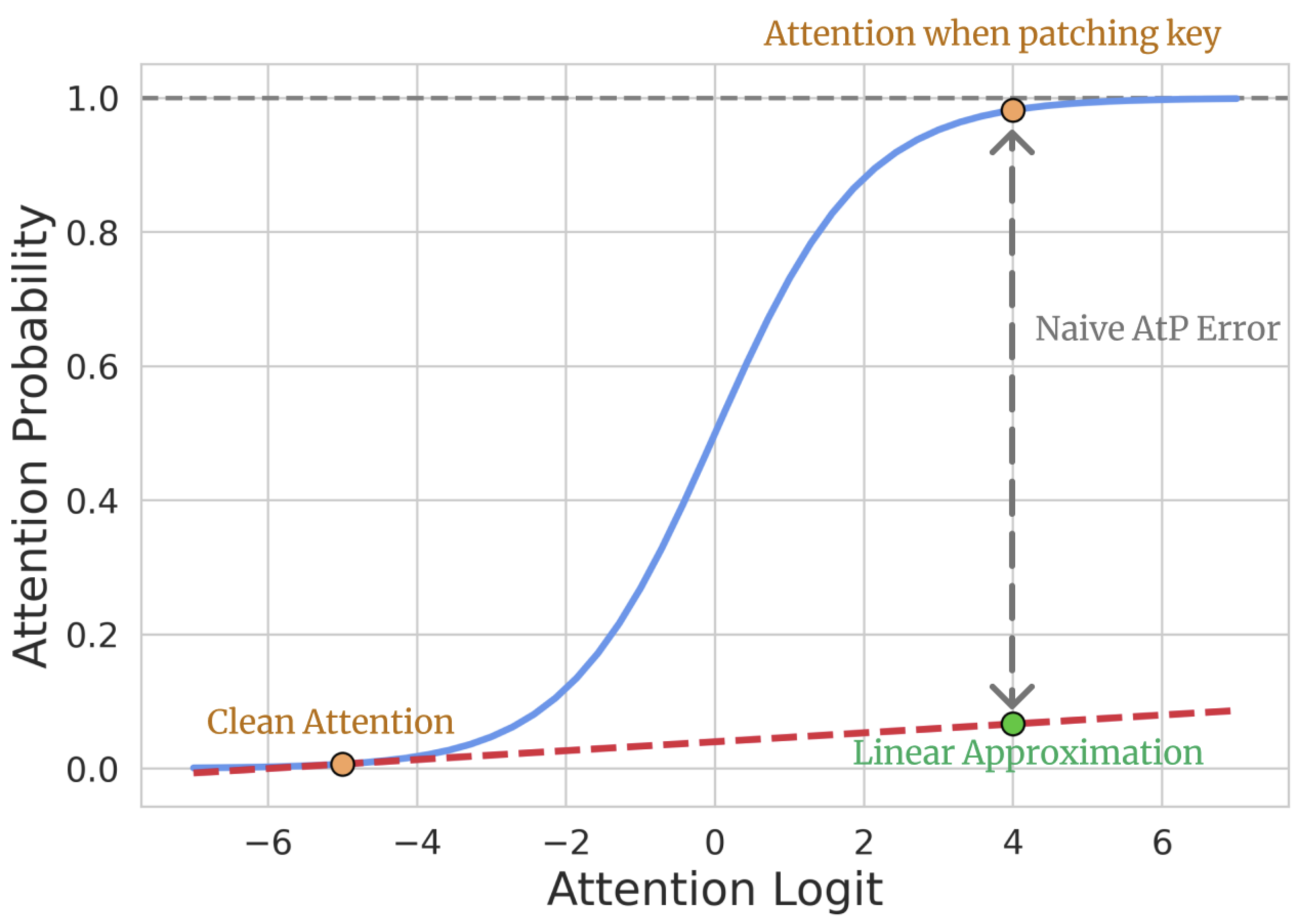}
    \caption{A linear approximation to the attention probability is a particularly poor approximation in cases where one or both of the endpoints are in a saturated region of the softmax. Note that when varying only a single key, the softmax becomes a sigmoid of the dot product of that key and the query.}
    \label{fig:saturation_cartoon}
\end{figure}

The first failure mode occurs when the preactivation on $\xclean$ is in a flat region of the activation function (e.g. produces a saturated attention weight), but the preactivation on $\xnoise$ is not in that region. As is apparent from~\Cref{eq:inter_atp}, AtP uses a linear approximation to the ground truth in~\Cref{eq:ground_truth}, so if the non-linear function is badly approximated by the local gradient, AtP ceases to be accurate -- see~\Cref{fig:saturation_cartoon} for an illustration and~\Cref{fig:ioi_qk_fix_scatter} which denotes in color the maximal difference in attention observed between prompt pairs, suggesting % illustrating
that this failure mode occurs in practice.

Another, unrelated failure mode occurs due to cancellation between direct and indirect effects: roughly, if the total effect (on some prompt pair) is a sum of direct and indirect effects~\citep{pearl2001direct} $\inter(\node)=\inter^{\text{direct}}(\node)+\inter^{\text{indirect}}(\node)$, and these are close to cancelling, then a small multiplicative approximation error in $\hat\inter_{\text{AtP}}^{\text{indirect}}(\node)$, due to non-linearities such as GELU and softmax, can accidentally cause $|\hat\inter_{\text{AtP}}^{\text{direct}}(\node)+\hat\inter_{\text{AtP}}^{\text{indirect}}(\node)|$ to be orders of magnitude smaller than $|\inter(\node)|$.

\subsubsection{False negatives from attention saturation} \label{sec:saturation}

\begin{figure}
    \centering
    \includegraphics[width=\textwidth]{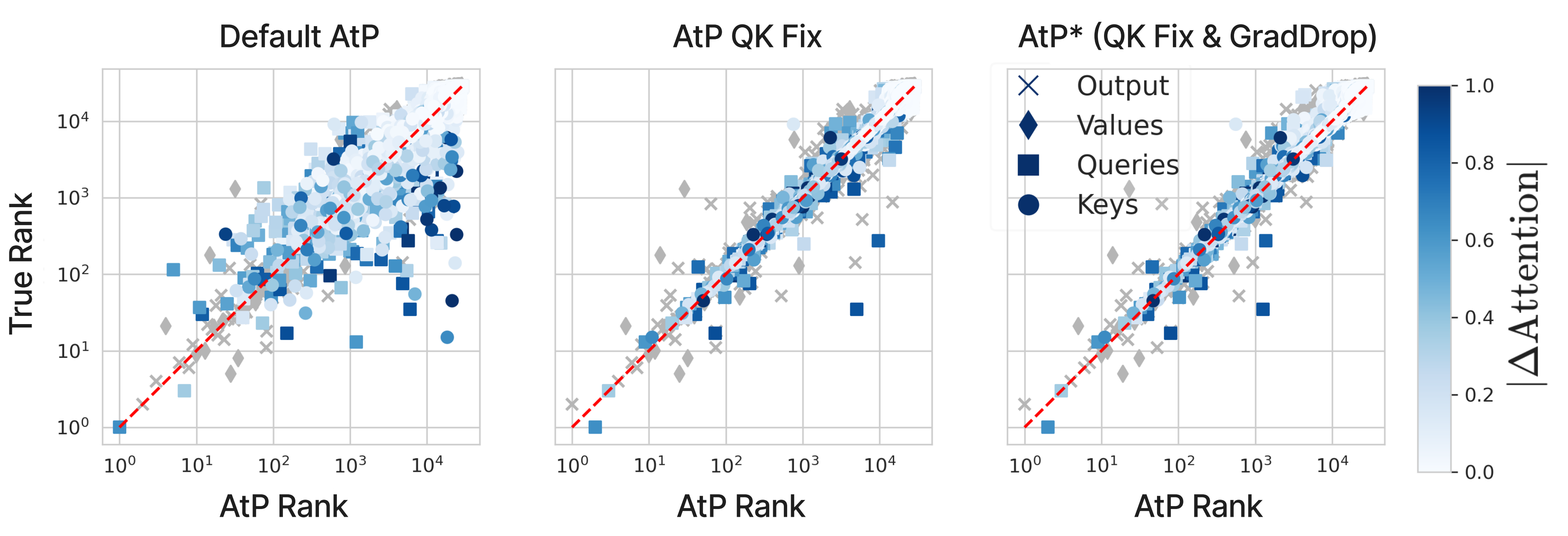}
    \caption{Ranks of $\contrib(\node)$ against ranks of $\hat\contrib_\text{AtP}(\node)$, on Pythia-12B on \texttt{CITY-PP}.  Both improvements to AtP reduce the number of false negatives (bottom right triangle area), where in this case most improvements come from the QK fix. Coloration indicates the maximum absolute difference in attention probability when comparing $\xclean$ and patching a given query or key. Many false negatives are keys and queries with significant maximum difference in attention probability, suggesting they are due to attention saturation as illustrated in~\Cref{fig:saturation_cartoon}. Output and value nodes are colored in grey as they do not contribute to the attention probability.}
    \label{fig:ioi_qk_fix_scatter}
\end{figure}

AtP relies on the gradient at each activation being reflective of the true behaviour of the function with respect to intervention at that activation. In some cases, though, a node may immediately feed into a non-linearity whose effect may not be adequately predicted by the gradient; for example, attention key and query nodes feeding into the attention softmax non-linearity. To showcase this, we plot the true rank of each node's effect against its rank assigned by AtP in~\Cref{fig:ioi_qk_fix_scatter} (left). The plot shows that there are many pronounced false negatives (below the dashed line), especially among keys and queries. 

Normal activation patching for queries and keys involves changing a query or key and then re-running the rest of the model, keeping all else the same. AtP takes a linear approximation to the entire rest of the model rather than re-running it. We propose explicitly re-computing the first step of the rest of the model, i.e. the attention softmax, and then taking a linear approximation to the rest. Formally, for attention key and query nodes, instead of using the gradient on those nodes directly, we take the difference in attention weight caused by that key or query, multiplied by the gradient on the attention weights themselves. This requires finding the change in attention weights from each key and query patch --- but that can be done efficiently using (for all keys and queries in total) less compute than two transformer forward passes. This correction avoids the problem of saturated attention, while otherwise retaining the performance of AtP. 

\paragraph{Queries}
For the queries, we can easily compute the adjusted effect by running the model on $\xnoise$ and caching the noise queries. We then run the model on $\xclean$ and cache the attention keys and weights. Finally, we compute the attention weights that result from combining all the keys from the $\xclean$ forward pass with the queries from the $\xnoise$ forward pass. This costs approximately as much as the unperturbed attention computation of the transformer forward pass. For each query node $\node$ we refer to the resulting weight vector as $\attn(\node)_{\text{patch}}$, in contrast with the weights $\attn(\node)(\xclean)$ from the clean forward pass. The improved attribution estimate for $\node$ is then

\begin{align} \label{eq:qfix}
    \hat\inter^Q_{\text{AtPfix}}(\node;\xclean,\xnoise) :={}& \sum_k \hat{\inter}_{\text{AtP}}(\text{attn}(n)_k; \xclean, \xnoise)\\ ={}&(\attn(\node)_{\text{patch}} - \attn(\node)(\xclean))^\intercal \frac{\partial\metric(\model(\xclean))}{\partial \attn(\node)}\Big{|}_{\attn(\node)=\attn(\node)(\xclean)}
\end{align}

\paragraph{Keys} \label{par:kfix}

For the keys we first describe a simple but inefficient method. We again run the model on $\xnoise$, caching the noise
keys. We also run it on $\xclean$, caching the clean queries and attention probabilities.
Let key nodes for a single attention head be $\node^k_{1},\dots,\node^k_{\numTokens}$ and let $\operatorname{queries}(\node^k_{t})=\{\node^q_{1},\dots,\node^q_{\numTokens}\}$
be the set of query nodes for the same head as node $\node^k_t$. We then define

\begin{align}
    \attn_{\text{patch}}^t(\node^q)&:=\attn(\node^q)(\xclean\mid\doGets{\node^k_{t}}{\node^k_{t}(\xnoise)})\label{eq:patch_attn_key}\\
    \Delta_{t}\attn(\node^q)&:=\attn_{\text{patch}}^t(\node^q)-\attn(\node^q)(\xclean)\label{eq:delta_attn_key}
\end{align}

The improved attribution estimate for $\node^k_{t}$ is then 

\begin{align} \label{eq:kfix}
\hat{\inter}_{\text{AtPfix}}^{K}(\node^k_{t};\xclean,\xnoise) & :=\sum_{\node^q\in\operatorname{queries}(\node^k_{t})}\Delta_{t}\attn(\node^q)^\intercal\frac{\partial\metric(\model(\xclean))}{\partial\attn(\node^q)}\Big{|}_{\attn(\node^q)=\attn(\node^q)(\xclean)}%\approx\contrib(\node^k_{t})
\end{align}

However, the procedure we just described is costly to execute as it requires $\operatorname{O}(\numTokens^3)$ flops to naively compute~\Cref{eq:delta_attn_key} for all $\numTokens$ keys. In~\Cref{app:kfix} we describe a more efficient variant that takes no more compute than the forward pass attention computation itself (requiring $\operatorname{O}(\numTokens^2)$ flops). Since~\Cref{eq:qfix} is also cheaper to compute than a forward pass, the full QK fix requires less than two transformer forward passes (since the latter also includes MLP computations).

For attention nodes we show the effects of applying the query and key fixes in~\Cref{fig:ioi_qk_fix_scatter} (middle). We observe that the propagation of Q/K effects has a major impact on reducing the false negative rate.

\subsubsection{False negatives from cancellation} \label{sec:cancellation}

This form of cancellation occurs when the backpropagated gradient from indirect effects is combined with the gradient from the direct effect. We propose a way to modify the backpropagation within the attribution patching to reduce this issue. If we artificially zero out the gradient at a downstream layer that contributes to the indirect effect, the cancellation is disrupted. (This is also equivalent to patching in clean activations at the outputs of the layer.) Thus we propose to do this iteratively, sweeping across the layers. Any node whose effect does not route through the layer being gradient-zeroed will have its estimate unaffected.

We call this method \emph{GradDrop}. For every layer $\ell\in\{1,\ldots,L\}$ in the model, GradDrop computes an AtP estimate for all nodes, where gradients on the residual contribution from $\ell$ are set to 0, including the propagation to earlier layers. This provides a different estimate for all nodes, for each layer that was dropped. We call the so-modified gradient $\frac{\partial\metric^{\ell}}{\partial \node}=\frac{\partial\metric}{\partial \node}(\model(\xclean\mid\doGets{\node^{\text{out}}_\ell}{\node^{\text{out}}_\ell(\xclean)}))$ when dropping layer $\ell$, where $\node^{\text{out}}_\ell$ is the contribution to the residual stream across all positions. Using $\frac{\partial\metric^{\ell}}{\partial \node}$ in place of $\frac{\partial\metric^{\ell}}{\partial \node}$ in the AtP formula produces an estimate $\hat \inter_{\text{AtP+GD}_\ell}(n)$. Then, the estimates are aggregated by averaging their absolute values, and then scaling by $\frac{L}{L-1}$ to avoid changing the direct-effect path's contribution (which is otherwise zeroed out when dropping the layer the node is in). 

\begin{align}\label{eq:contrib_atp_gd}
    \hat\contrib_{\text{AtP+GD}}(\node) &:= \expect{\xclean,\xnoise}{\frac{1}{L-1} \sum_{\ell=1}^L\left|\hat\inter_{\text{AtP+GD}_\ell}(\node;\xclean,\xnoise)\right|} %\approx |\contrib(\node)|
\end{align}

Note that the forward passes required for computing $\hat\inter_{\text{AtP+GD}_\ell}(\node;\xclean,\xnoise)$ don't depend on $\ell$, so the extra compute needed for GradDrop is $L$ backwards passes from the same intermediate activations on a clean forward pass. This is also the case with the QK fix: the corrected attributions $\hat\inter_{\text{AtPfix}}$ are dot products with the attention weight gradients, so the only thing that needs to be recomputed for $\hat\inter_{\text{AtPfix+GD}_\ell}(\node)$ is the modified gradient $\frac{\partial \metric^\ell}{\partial \attn(\node)}$. Thus, computing~\Cref{eq:contrib_atp_gd} takes $L$ backwards passes\footnote{This can be reduced to $(L+1)/2$ by reusing intermediate results.} on top of the costs for AtP.

\begin{figure}[t]
    \centering
    \includegraphics[width=0.5\textwidth]{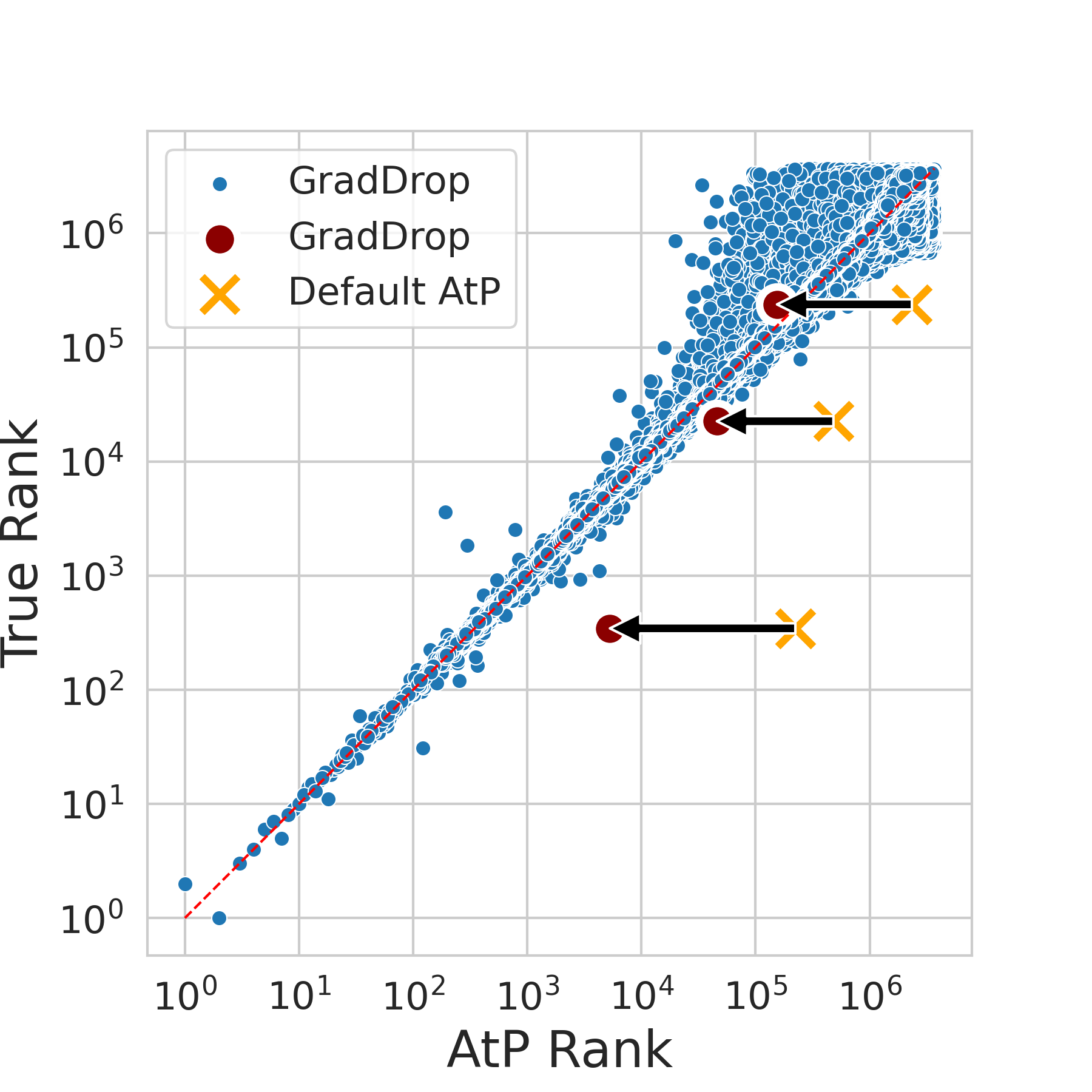}
    \caption{True rank and rank of AtP estimates with and without GradDrop, using Pythia-12B on the \texttt{CITY-PP} distribution with \NeuronNodes. GradDrop provides a significant improvement to the largest neuron false negatives (red circles) relative to Default AtP (orange crosses).}
    \label{fig:graddrop_neurons}
\end{figure}

We show the result of applying GradDrop on attention nodes in~\Cref{fig:ioi_qk_fix_scatter}~(right) and on MLP nodes in~\Cref{fig:graddrop_neurons}. In~\Cref{fig:graddrop_neurons}, we show the true effect magnitude rank against the AtP+GradDrop rank, while highlighting nodes which improved drastically by applying GradDrop. We give some arguments and intuitions on the benefit of GradDrop in~\Cref{app:graddrop}.

\paragraph{Direct Effect Ratio}
To provide some evidence that the observed false negatives are due to cancellation, we compute the ratio between the direct effect $\contrib^{\text{direct}}(\node)$ and the total effect $\contrib(\node)$. A higher direct effect ratio indicates more cancellation. We observe that the most significant false negatives corrected by GradDrop in~\Cref{fig:graddrop_neurons} (highlighted) have high direct effect ratios of $5.35$, $12.2$, and $0$ (no direct effect) , while the median direct effect ratio of all nodes is $0$ (if counting all nodes) or $0.77$ (if only counting nodes that have direct effect). Note that direct effect ratio is only applicable to nodes which in fact have a direct connection to the output, and not e.g. to MLP nodes at non-final token positions, since all disconnected nodes have a direct effect of 0 by definition. %So while high direct effect ratio is an indicator of cancellation, a ratio of 0 does not necessarily imply the absence of cancellation.

\subsection{Diagnostics}\label{sec:diagnostics}

Despite the improvements we have proposed in~\Cref{sec:improvements}, there is no guarantee that AtP* produces no false negatives. Thus, it is desirable to obtain an upper confidence bound on the effect size of nodes that might be missed by AtP*, i.e. that aren't in the top $K$ AtP* estimates, for some $K$. Let the top $K$ nodes be $\text{Top}^K_{AtP*}$. It so happens that we can use subset sampling to obtain such a bound.

As described in~\Cref{alg:subsampling} and~\Cref{par:subsampling}, the subset sampling algorithm returns summary statistics: $\bar i^\node_\pm$, $s^\node_\pm$ and $\text{count}^\node_\pm$ for each node $\node$: the average effect size $\bar i^\node_\pm$ of a subset conditional on the node being contained in that subset ($+$) or not ($-$), the sample standard deviations $s^\node_\pm$, and the sample sizes $\text{count}^\node_\pm$. Given these, consider a null hypothesis\footnote{This is an unconventional form of $H_0$ -- typically a null hypothesis will say that an effect is insignificant. However, the framework of statistical hypothesis testing is based on determining whether the data let us reject the null hypothesis, and in this case the hypothesis we want to reject is the presence, rather than the absence, of a significant false negative.} $H_0^\node$ that $|\inter(\node)|\geq \theta$, for some threshold $\theta$, versus the alternative hypothesis $H_1^\node$ that $|\inter(\node)|<\theta$. We use a one-sided Welch's t-test\footnote{This relies on the populations being approximately unbiased and normally distributed, and not skewed. This tended to be true on inspection, and it's what the additivity assumption (see~\cref{par:subsampling}) predicts for a single prompt pair --- but a nonparametric bootstrap test may be more reliable, at the cost of additional compute.} to test this hypothesis; the general practice with a compound null hypothesis is to select the simple sub-hypothesis that gives the greatest $p$-value, so to be conservative, the simple null hypothesis is that $\inter(\node)=\theta\operatorname{sign}(\bar i^\node_+-\bar i^\node_-)$, giving a test statistic of $t^\node=(\theta - |\bar i^\node_+-\bar i^\node_-|)/s^\node_{\text{Welch}}$, which gives a $p$-value of $p^\node=\mathbb{P}_{T\sim t_{\nu^\node_{\text{Welch}}}}(T>t^\node)$.

To get a combined conclusion across all nodes in $\Nodes\setminus\text{Top}^K_{AtP*}$, let's consider the hypothesis $H_0=\bigvee_{\node\in\Nodes\setminus\text{Top}^K_{AtP*}} H_0^\node$ that \emph{any} of those nodes has true effect $|\inter(\node)|>\theta$. Since this is also a compound null hypothesis, $\max_\node p^\node$ is the corresponding $p$-value. Then, to find an upper confidence bound with specified confidence level $1-p$, we invert this procedure to find the lowest $\theta$ for which we still have at least that level of confidence. We repeat this for various settings of the sample size $\numSamples$ in~\Cref{alg:subsampling}. The exact algorithm is described in~\Cref{app:diagnostic_algorithm}.

\begin{figure}
    \centering
    \begin{subfigure}[b]{0.49\textwidth}
        \centering
        \includegraphics[width=\textwidth]{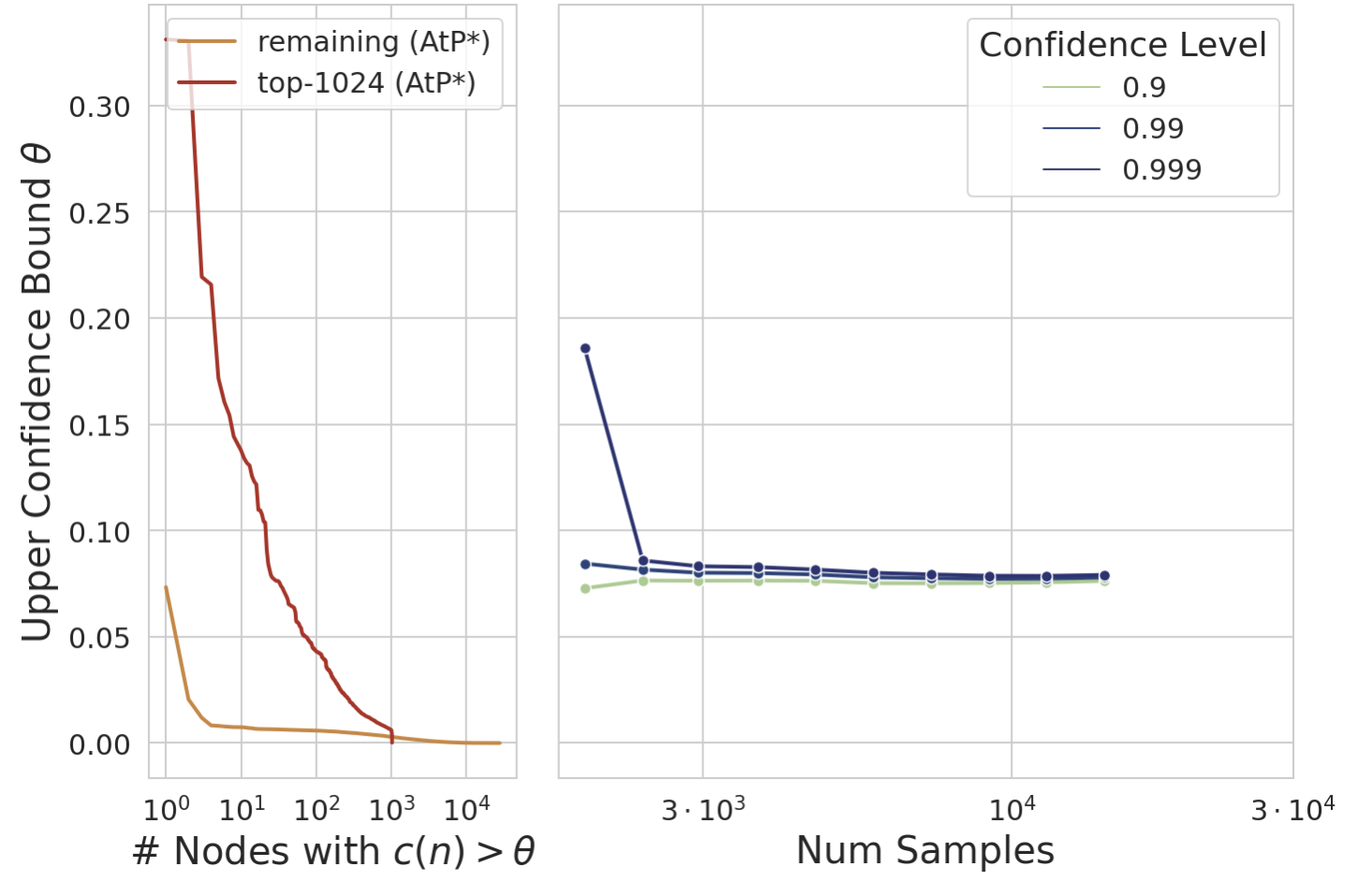}
        \caption{\texttt{IOI-PP}}
    \end{subfigure}
    \hfill
    \begin{subfigure}[b]{0.49\textwidth}
        \label{fig:diagnostic_main_distr}
        \includegraphics[width=\textwidth]{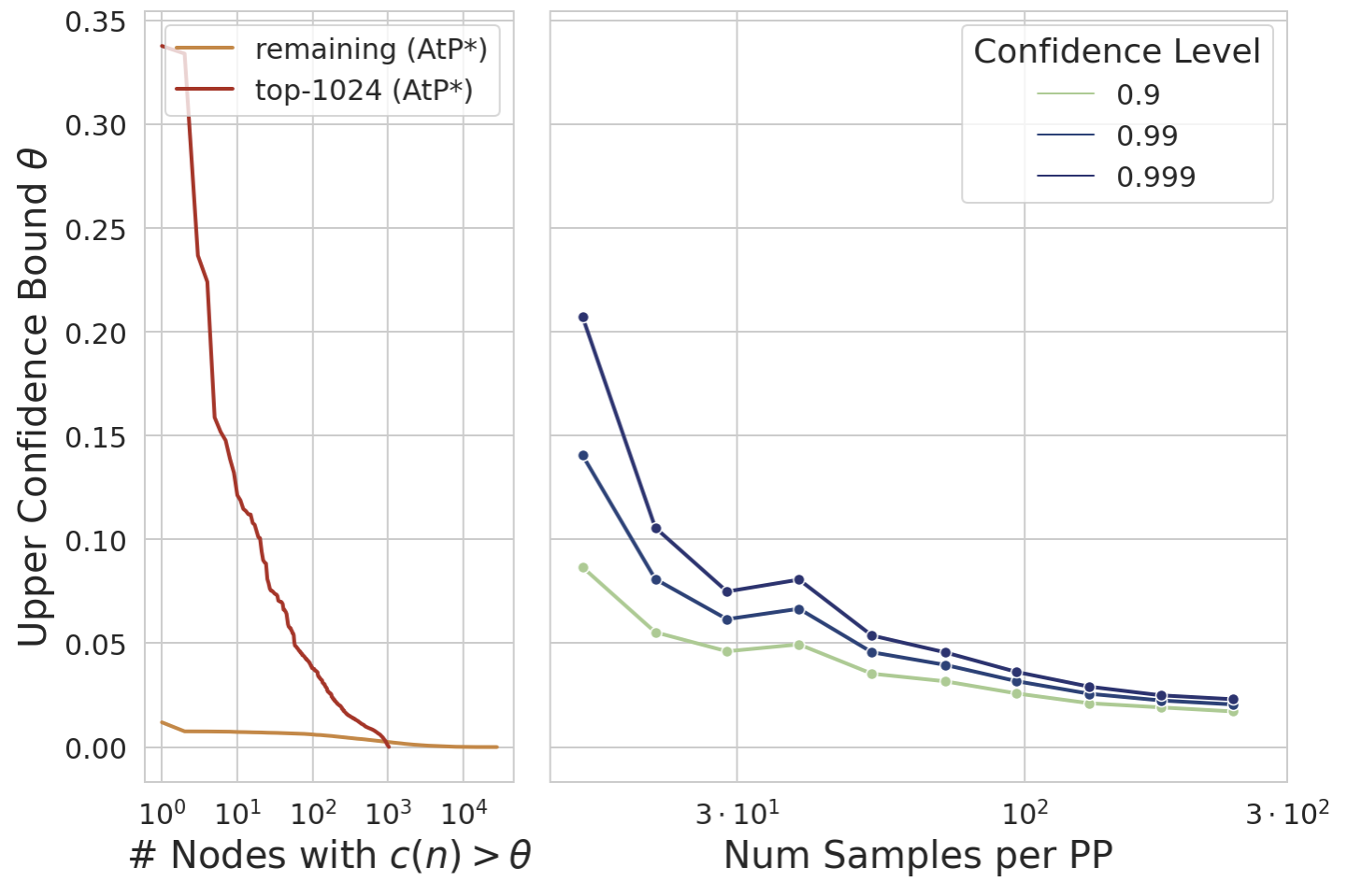}
        \caption{\texttt{IOI}}
    \end{subfigure}
    \caption{Upper confidence bounds on effect magnitudes of false negatives (i.e. nodes not in the top 1024 nodes according to AtP$^*$), at 3 confidence levels, varying the sampling budget. On the left we show in red the true effect of the nodes which are ranked highest by AtP$^*$. We also show the true effect magnitude at various ranks of the remaining nodes in orange.\\
    We can see that the bound for (a) finds the true biggest false negative reasonably early, while for (b), where there is no large false negative, we progressively keep gaining confidence with more data.\\
    Note that the costs involved per prompt pair are substantially different between the subplots, and in particular this diagnostic for the distributional case (b) is substantially cheaper to compute than the verification cost of 1024 samples per prompt pair.}
    \label{fig:diagnostic_main}
\end{figure}

In~\Cref{fig:diagnostic_main}, we report the upper confidence bounds at confidence levels 90\%, 99\%, 99.9\% from running~\Cref{alg:subsampling} with a given $\numSamples$ (right subplots), as well as the number of nodes that have a true contribution $\contrib(\node)$ greater than $\theta$ (left subplots).

\subsection{Baselines} \label{sec:baselines}

\paragraph{Iterative} \label{par:iterative}
The most straightforward method is to directly do Activation Patching to find the true effect $\contrib(n)$ of each node, in some uninformed random order. This is necessarily inefficient.

However, if we are scaling to a distribution, it is possible to improve on this, by alternating between phases of (i) for each unverified node, picking a not-yet-measured prompt pair on which to patch it, (ii) ranking the not-yet-verified nodes by the average observed patch effect magnitudes, taking the top $|\Nodes|/|\promptPairDist|$ nodes, and verifying them. This balances the computational expenditure on the two tasks, and allows us to find large nodes sooner, at least as long as their large effect shows up on many prompt pairs.

Our remaining baseline methods rely on an approximate \emph{node additivity assumption}: that when intervening on a set of nodes $\subsetNodes$, the measured effect $\inter(\subsetNodes; \xclean, \xnoise)$ is approximately equal to $\sum_{\node\in\subsetNodes} \inter(\node; \xclean, \xnoise)$.

\paragraph{Subsampling}\label{par:subsampling}

Under the approximate node additivity assumption, we can construct an approximately unbiased estimator of $\contrib(\node)$. We select the sets $\subsetNodes_k$ to contain each node independently with some probability $p$, and additionally sample prompt pairs $\xclean_k,\xnoise_k\sim \mathcal{D}$. For any node $\node$, and sets of nodes $\subsetNodes_k\subset\Nodes$, let $\subsetNodes^+(\node)$ be the collection of all those that contain $\node$, and $\subsetNodes^-(\node)$ be the collection of those that don't contain $\node$; we'll write these node sets as $\subsetNodes^+_k(\node)$ and $\subsetNodes^-_k(\node)$, and the corresponding prompt pairs as ${\xclean_k}^+(\node),{\xnoise_k}^+(\node)$ and ${\xclean_k}^-(\node),{\xnoise_k}^-(\node)$. The subsampling (or subset sampling) estimator is then given by

\begin{align}
    \label{eq:subsamplinginter}
    \hat{\inter}_{\text{SS}}(\node) &:= {\frac{1}{|\subsetNodes^+(\node)|}\sum_{k=1}^{|\subsetNodes^+(\node)|} \inter(\subsetNodes^+_k(\node); {\xclean_k}^+(\node), {\xnoise_k}^+(\node)) - \frac{1}{|\subsetNodes^-(\node)|}\sum_{k=1}^{|\subsetNodes^-(\node)|} \inter(\subsetNodes^-_k(\node); {\xclean_k}^-(\node), {\xnoise_k}^-(\node)) } \\
    \label{eq:subsampling}
    \hat\contrib_{\text{SS}}(\node) &:= |\hat\inter_{\text{SS}}(\node)|
\end{align}

The estimator $\hat{\inter}_{\text{SS}}(\node)$ is unbiased if there are no interaction effects, and has a small bias proportional to $p$ under a simple interaction model (see \Cref{app:subsampling} for proof).

In practice, we compute all the estimates $\hat\contrib_{\text{SS}}(\node)$ by sampling a binary mask over all nodes from i.i.d. Bernoulli$^{|\Nodes|}(p)$ -- each binary mask can be identified with a node set $\subsetNodes$. 
In~\Cref{alg:subsampling}, we describe how to compute summary statistics related to~\Cref{eq:subsampling} efficiently for all nodes $\node\in\Nodes$. The means $\bar i^\pm$ are enough to compute $\hat\contrib_{\text{SS}}(\node)$, while other summary statistics are involved in bounding the magnitude of a false negative (cf.~\Cref{sec:diagnostics}). (Note, $\rcount^\pm_\node$ is just an alternate notation for $|\eta^\pm(\node)|$.)

\begin{algorithm}
  \caption{Subsampling}
    \label{alg:subsampling}
  \begin{algorithmic}[1]
    \Require{$p\in(0,1)$, model $\model$, metric $\metric$, prompt pair distribution $\promptPairDist$, num samples $\numSamples$}

    \State $\rcount^\pm$, $\rsum^\pm$, $\rssum^\pm$ $\gets 0^{|\Nodes|}$ \Comment{Init counts and running sums to 0 vectors}
    \For{$i \gets 1 \textrm{ to } \numSamples$}
       \State $\xclean, \xnoise \sim \promptPairDist$
       \State $\mask^+ \gets \Bernoulli^{|\Nodes|}(p)$ \Comment{Sample binary mask for patching}
       \State $\mask^- \gets 1-\mask^+$
       \State $i \gets \inter(\{n\in\Nodes:\mask^+_\node=1\}; \xclean, \xnoise)$\Comment{$\subsetNodes^+=\{\node\in\Nodes:\mask^+_n=1\}$}
       \State $\rcount^\pm \,\gets\, \rcount^\pm + \mask^\pm$
       \State $\rsum^\pm \,\gets\, \rsum^\pm + i\cdot\mask^\pm $
       \State $\rssum^\pm \,\gets\, \rssum^\pm + i^2\cdot\mask^\pm$
      \EndFor
    \State $\bar i^\pm \gets \rsum^\pm / \rcount^\pm$
    \State $s^\pm \gets \sqrt{(\rssum^\pm - (\bar i^\pm)^2) / (\rcount^\pm - 1)}$
    \State \Return{$\rcount^\pm$, $\bar i^\pm$, $s^\pm$}\Comment{If diagnostics are not required, $\bar i^\pm$ is sufficient.}
  \end{algorithmic}
\end{algorithm}

\paragraph{Blocks \& Hierarchical}\label{par:blocks}
Instead of sampling each $\subsetNodes$ independently, we can group nodes into fixed ``blocks'' $\subsetNodes$ of some size, and patch each block to find its aggregated contribution $\contrib(\subsetNodes)$; we can then traverse the nodes, starting with high-contribution blocks and proceeding from there. 

There is a tradeoff in terms of the block size: using large blocks increases the compute required to traverse a high-contribution block, but using small blocks increases the compute required to finish traversing all of the blocks. We refer to the fixed block size setting as \emph{Blocks}. Another way to handle this tradeoff is to add recursion: the blocks can be grouped into higher-level blocks, and so forth. We call this method \emph{Hierarchical}.

We present results from both methods in our comparison plots, but relegate details to~\Cref{app:baselines}. Relative to subsampling, these grouping-based methods have the disadvantage that on distributions, their cost scales linearly with size of $\promptPairDist$'s support, in addition to scaling with the number of nodes\footnote{AtP* also scales linearly in the same way, but with far fewer forward passes per prompt pair.}.

\section{Experiments} \label{sec:experiments}

\subsection{Setup} \label{sec:setup}
\paragraph{Nodes}
When attributing model behavior to components, an important choice is the partition of the model's computational graph into units of analysis or `nodes' $\Nodes \ni \node$ (cf.~\Cref{sec:problem-statement}). We investigate two settings for the choice of $\Nodes$, \emph{\AttentionNodes{}} and \emph{\NeuronNodes{}}. For \NeuronNodes{}, each MLP neuron\footnote{We use the neuron post-activation for the node; this makes no difference when causally intervening, but for AtP it's beneficial, because it makes the $\node\mapsto\metric(\node)$ function more linear.} is a separate node. For \AttentionNodes{}, we consider the query, key, and value vector for each head as distinct nodes, as well as the pre-linear per-head attention output\footnote{We include the output node because it provides additional information about what function an attention head is serving, particularly in the case where its queries have negligible patch effects relative to its keys and/or values. This may happen as a result of choosing $\xclean,\,\xnoise$ such that the query does not differ across the prompts.}. We also refer to these units as `sites'. For each site, we consider each copy of that site at different token positions as a separate node. As a result, we can identify each node $\node \in \Nodes$ with a pair $(T, S)$ from the product TokenPosition $\times$ Site. Since our two settings for $\Nodes$ are using a different level of granularity and are expected to have different per-node effect magnitudes, we present results on them separately.

\paragraph{Models}
We investigate transformer language models from the Pythia suite~\citep{biderman2023pythia} of sizes between 410M and 12B parameters. This allows us to demonstrate that our methods are applicable across scale. Our cost-of-verified-recall plots in~\Cref{fig:main_result,fig:recall_90,fig:rand_and_distr_result} refer to Pythia-12B. Results for other model sizes are presented via the relative-cost~(cf.~\Cref{sec:irwrgm}) plots in the main body~\Cref{fig:rand_and_distr_model_sweep} and disaggregated via cost-of-verified recall in~\Cref{app:detailed_results}.

\paragraph{Effect Metric $\metric$}
All reported results use the negative log probability\footnote{Another popular metric is the difference in logits between the clean and noise target. As opposed to the negative logprob, the logit difference is linear in the final logits and thus might favor AtP. A downside of logit difference is that it is sensitive to the noise target, which may not be meaningful if there are multiple plausible completions, such as in \texttt{IOI}.} as their loss function $\metric$. We compute $\metric$ relative to targets from the clean prompt $\xclean$. We briefly explore other metrics in \Cref{app:metrics}.

\subsection{Measuring Effectiveness and Efficiency}\label{sec:measuring_effectiveness}
\paragraph{Cost of verified recall}
As mentioned in the introduction, we’re primarily interested in finding the largest-effect nodes -- see~\Cref{app:true_effect_distribution} for the distribution of $\contrib(\node)$ across models and distributions.
Once we have obtained node estimates via a given method, it is relatively cheap to directly measure true effects of top nodes one at a time; we refer to this as “verification”. Incorporating this into our methodology, we find that false positives are typically not a big issue; they are simply revealed during verification. In contrast, false negatives are not so easy to remedy without verifying all nodes, which is what we were trying to avoid.

We compare methods on the basis of total compute cost (in \# of forward passes) to verify the $K$ nodes with biggest true effect magnitude, for varying $K$. The procedure being measured is to first compute estimates (incurring an estimation cost), and then sweep through nodes in decreasing order of estimated magnitude, measuring their individual effects $\contrib(\node)$ (i.e. verifying them), and incurring a verification cost. Then the total cost is the sum of these two costs.

\paragraph{Inverse-rank-weighted geometric mean cost}\label{sec:irwrgm}
Sometimes we find it useful to summarize the method performance with a scalar; this is useful for comparing methods at a glance across different settings (e.g. model sizes, as in~\Cref{fig:model_sweep}), or for selecting hyperparameters (cf.~\Cref{app:hyper_selection}). The cost of verified recall of the top $K$ nodes is of interest for $K$ at varying orders of magnitude. In order to avoid the performance metric being dominated by small or large $K$, we assign similar total weight to different orders of magnitude: we use a weighted average with weight $1/K$ for the cost of the top $K$ nodes. Similarly, since the costs themselves may have different orders of magnitude, we average them on a log scale -- i.e., we take a geometric mean.

This metric is also proportional to the area under the curve in plots like~\Cref{fig:main_result}. To produce a more understandable result, we always report it relative to (i.e. divided by) the oracle verification cost on the same metric; the diagonal line is the oracle, with relative cost 1. We refer to this as the IRWRGM (inverse-rank-weighted relative geometric mean) cost, or the relative cost.

Note that the preference of the individual practitioner may be different such that this metric is no longer accurately measuring the important rank regime. For example, AtP* pays a notable upfront cost relative to AtP or AtP+QKfix, which sets it at a disadvantage when it doesn't manage to find additional false negatives; but this may or may not be practically significant. To understand the performance in more detail we advise to refer to the cost of verified recall plots, like \Cref{fig:main_result} (or many more in \Cref{app:detailed_results}).

\subsection{Single Prompt Pairs versus Distributions}
We focus many of our experiments on single prompt pairs. This is primarily because it's easier to set up and get ground truth data. It's also a simpler setting in which to investigate the question, and one that’s more universally applicable, since a distribution to generalize to is not always available.

\begin{figure}
    \centering
    \begin{subfigure}[b]{0.48\textwidth}
        \centering
        \includegraphics[width=\textwidth]{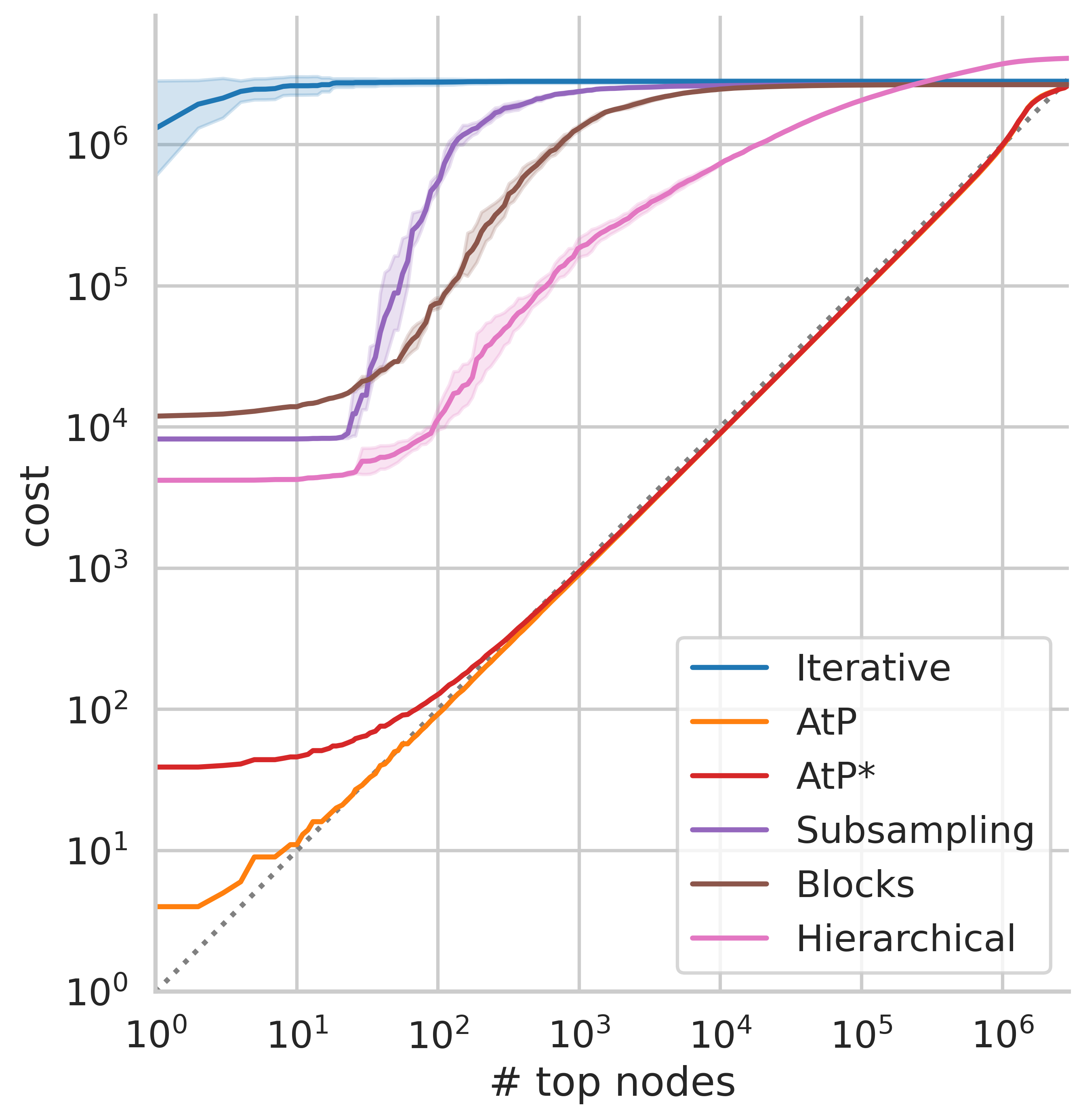}
        \caption{\NeuronNodes{} on \texttt{CITY-PP}}
    \end{subfigure}
    \hfill
    \begin{subfigure}[b]{0.48\textwidth}
        \centering
        \includegraphics[width=\textwidth]{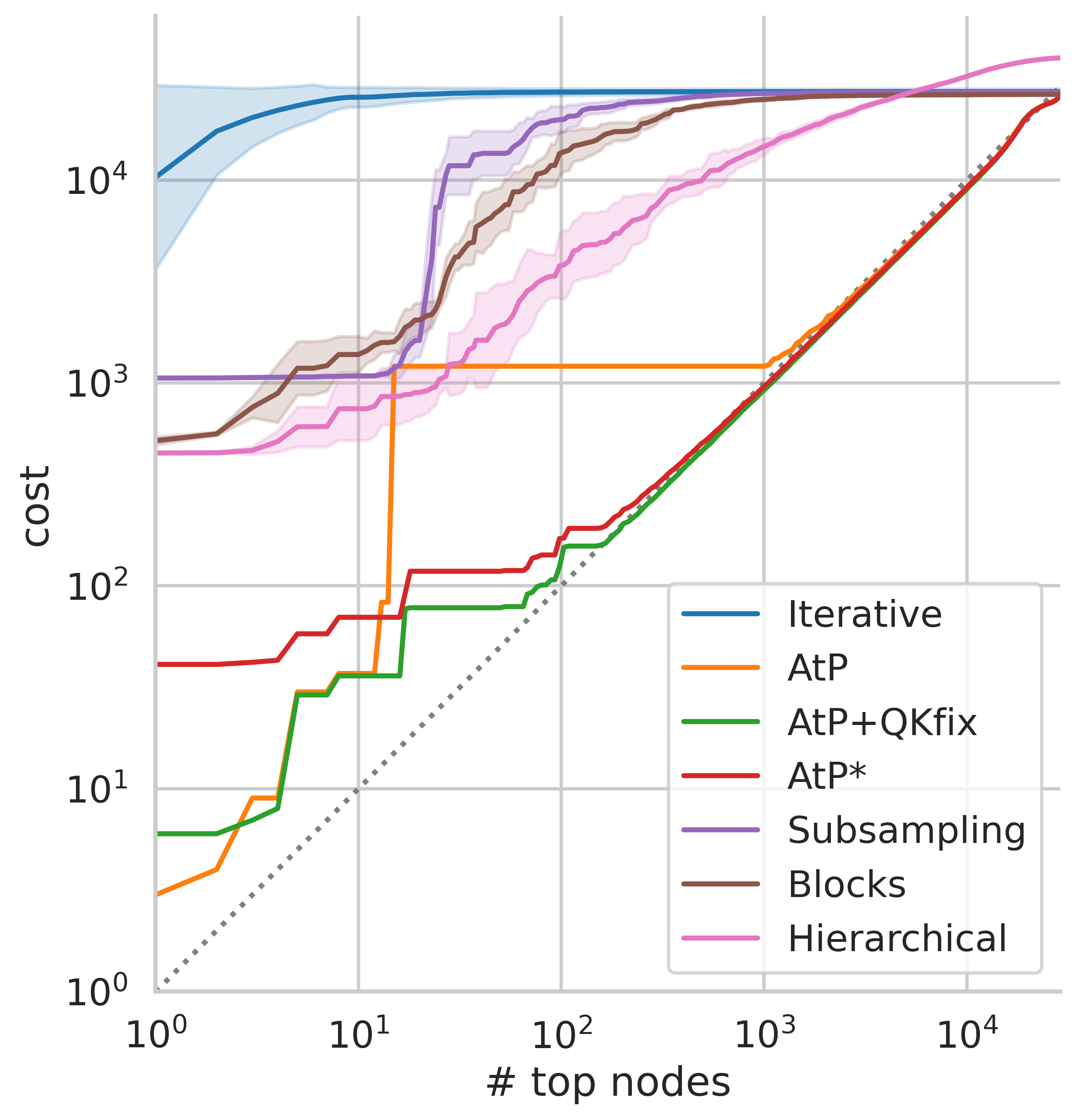}
        \caption{\AttentionNodes{} on \texttt{IOI-PP}}
    \end{subfigure}
    \caption{Costs of finding the most causally-important nodes in Pythia-12B using different methods on clean prompt pairs, with 90\% target recall. This highlights that the AtP* false negatives in~\Cref{fig:main_result} are a small minority of nodes.}
    \label{fig:recall_90}
\end{figure}

\paragraph{Clean single prompt pairs} \label{sec:clean_prompt_pairs} % ioi, barcelona/BJ, a_an
As a starting point we report results on single prompt pairs which we expect to have relatively clean circuitry\footnote{Formally, these represent prompt distributions via the delta distribution $p(\xclean, \xnoise) = \delta_{\xclean_1, \xnoise_1}(\xclean, \xnoise)$ where $\xclean_1, \xnoise_1$ is the singular prompt pair.}.
All singular prompt pairs are shown in~\Cref{tab:single_prompt_pairs}. \texttt{IOI-PP} is chosen to resemble an instance from the indirect object identification (IOI) task~\citep{wang2022interpretability}, a task predominantly involving attention heads. \texttt{CITY-PP} is chosen to elicit factual recall which previous research suggests involves early MLPs and a small number of late attention heads~\citep{meng2023locating,geva2023dissecting,nanda2023factfinding}. The country/city combinations were chosen such that Pythia-410M achieved low loss on both $\xclean$ and $\xnoise$ and such that all places were represented by a single token.

\begin{table}[t]
    \centering
    \begin{tabular}{c|c|c}
         Identifier & Clean Prompt & Noise Source Prompt \\
         \hline\hline
         \texttt{CITY-PP} & \makecell{\fbox{\strut{}\texttt{BOS}}\fbox{\strut{}City}\fbox{\strut{}:}\fbox{\strut{}\textvisiblespace{}Barcelona}\fbox{\strut{}\textbackslash{}n}\\
         \fbox{\strut{}Country}\fbox{\strut{}:}\doublebox{\strut{}\textvisiblespace{}Spain}} & \makecell{\fbox{\strut{}\texttt{BOS}}\fbox{\strut{}City}\fbox{\strut{}:}\fbox{\strut{}\textvisiblespace{}Beijing}\fbox{\strut{}\textbackslash{}n}\\
         \fbox{\strut{}Country}\fbox{\strut{}:}\fbox{\color{gray}\strut{}\textvisiblespace{}China}}\\
         \hline
         \texttt{IOI-PP} & \makecell{\fbox{\strut{}\texttt{BOS}}\fbox{\strut{}When}\fbox{\strut{}\textvisiblespace{}Michael}\fbox{\strut{}\textvisiblespace{}and}\fbox{\strut{}\textvisiblespace{}Jessica}\\
         \fbox{\strut{}\textvisiblespace{}went}\fbox{\strut{}\textvisiblespace{}to}\fbox{\strut{}\textvisiblespace{}the}\fbox{\strut{}\textvisiblespace{}bar}\fbox{\strut{},}\fbox{\strut{}\textvisiblespace{}Michael}\\
         \fbox{\strut{}\textvisiblespace{}gave}\fbox{\strut{}\textvisiblespace{}a}\fbox{\strut{}\textvisiblespace{}drink}\fbox{\strut{}\textvisiblespace{}to}\doublebox{\strut{}\textvisiblespace{}Jessica}}&
         \makecell{\fbox{\strut{}\texttt{BOS}}\fbox{\strut{}When}\fbox{\strut{}\textvisiblespace{}Michael}\fbox{\strut{}\textvisiblespace{}and}\fbox{\strut{}\textvisiblespace{}Jessica}\\
         \fbox{\strut{}\textvisiblespace{}went}\fbox{\strut{}\textvisiblespace{}to}\fbox{\strut{}\textvisiblespace{}the}\fbox{\strut{}\textvisiblespace{}bar}\fbox{\strut{},}\fbox{\strut{}\textvisiblespace{}Ashley}\\
         \fbox{\strut{}\textvisiblespace{}gave}\fbox{\strut{}\textvisiblespace{}a}\fbox{\strut{}\textvisiblespace{}drink}\fbox{\strut{}\textvisiblespace{}to}\fbox{\color{gray}\strut{}\textvisiblespace{}Michael}}\\
         \hline
         \texttt{RAND-PP} & \makecell{\fbox{\strut{}\texttt{BOS}}\fbox{\strut{}Her}\fbox{\strut{}\textvisiblespace{}biggest}\fbox{\strut{}\textvisiblespace{}worry}\fbox{\strut{}\textvisiblespace{}was}\fbox{\strut{}\textvisiblespace{}the}\\
         \fbox{\strut{}\textvisiblespace{}festival}\fbox{\strut{}\textvisiblespace{}might}\fbox{\strut{}\textvisiblespace{}suffer}\fbox{\strut{}\textvisiblespace{}and}\\
         \fbox{\strut{}\textvisiblespace{}people}\fbox{\strut{}\textvisiblespace{}might}\fbox{\strut{}\textvisiblespace{}erroneously}\doublebox{\strut{}\textvisiblespace{}think}} &
         \makecell{\fbox{\strut{}\texttt{BOS}}\fbox{\strut{}also}\fbox{\strut{}\textvisiblespace{}think}\fbox{\strut{}\textvisiblespace{}that}\fbox{\strut{}\textvisiblespace{}there}\\
         \fbox{\strut{}\textvisiblespace{}should}\fbox{\strut{}\textvisiblespace{}be}\fbox{\strut{}\textvisiblespace{}the}\fbox{\strut{}\textvisiblespace{}same}\fbox{\strut{}\textvisiblespace{}rules}\\
         \fbox{\strut{}\textvisiblespace{}or}\fbox{\strut{}\textvisiblespace{}regulations}\fbox{\strut{}\textvisiblespace{}when}\fbox{\color{gray}\strut{}\textvisiblespace{}it}}\\
    \end{tabular}
    \caption{Clean and noise source prompts for singular prompt pair distributions. Vertical lines denote tokenization boundaries. All prompts are preceded by the BOS (beginning of sequence) token. The last token is not part of the input. The last token of the clean prompt is used as the target in $\metric$.}
    \label{tab:single_prompt_pairs}
\end{table}

We show the cost of verified 100\% recall for various methods in~\Cref{fig:main_result}, where we focus on \NeuronNodes{} for \texttt{CITY-PP} and \AttentionNodes{} for \texttt{IOI-PP}. Exhaustive results for smaller Pythia models are shown in \Cref{app:detailed_results}. \Cref{fig:model_sweep} shows the aggregated relative costs for all models on \texttt{CITY-PP} and \texttt{IOI-PP}.

Instead of applying the strict criterion of recalling all important nodes, we can also relax this constraint. In~\Cref{fig:recall_90}, we show the cost of verified 90\% recall in the two clean prompt pair settings.

\paragraph{Random prompt pair}

\begin{figure}
    \centering
    \begin{subfigure}[b]{0.48\textwidth}
        \centering
        \includegraphics[width=\textwidth]{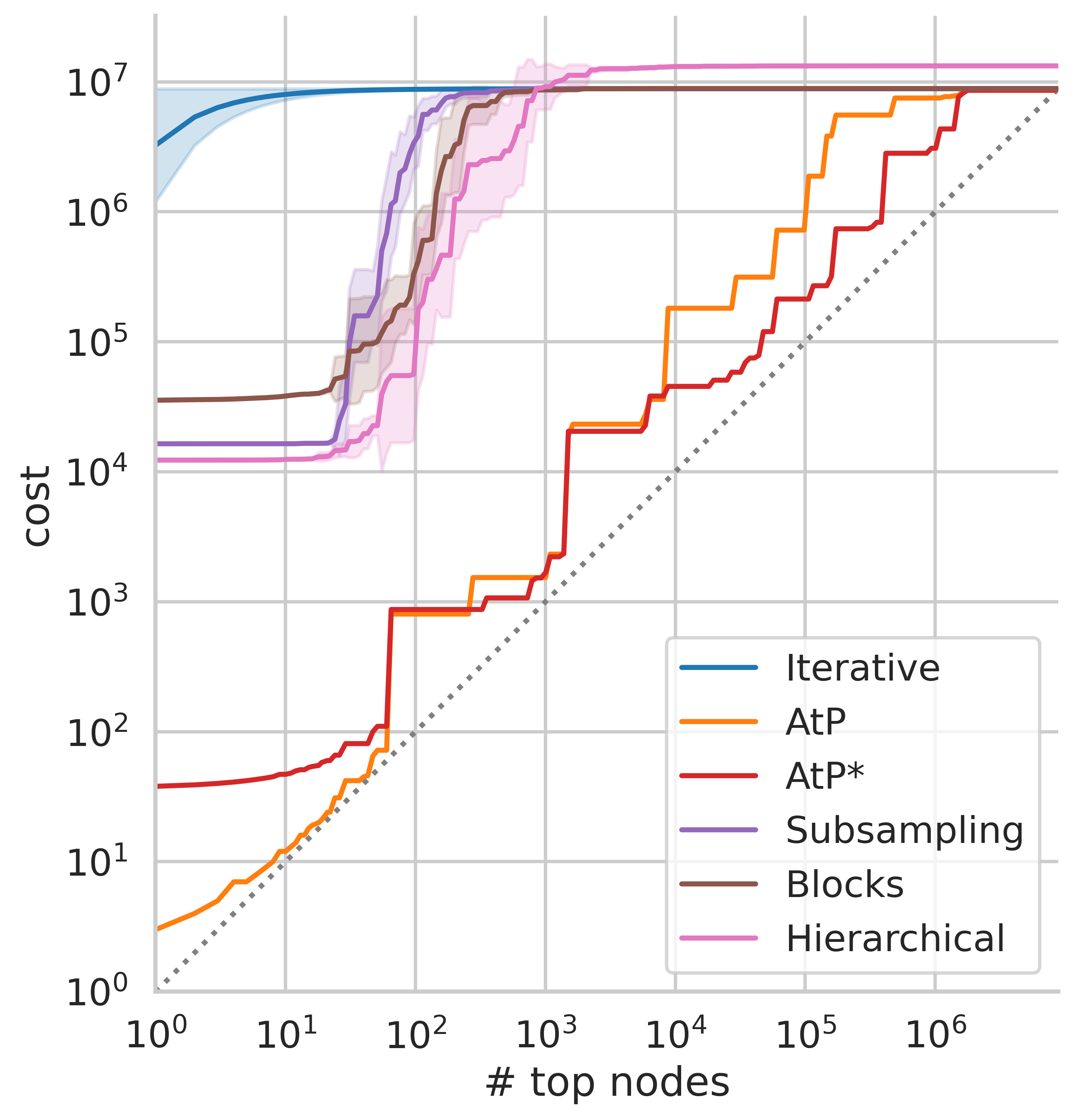}
        \caption{\texttt{RAND-PP} MLP neurons.}
    \end{subfigure}
    \hfill
    \begin{subfigure}[b]{0.48\textwidth}
        \centering
        \includegraphics[width=\textwidth]{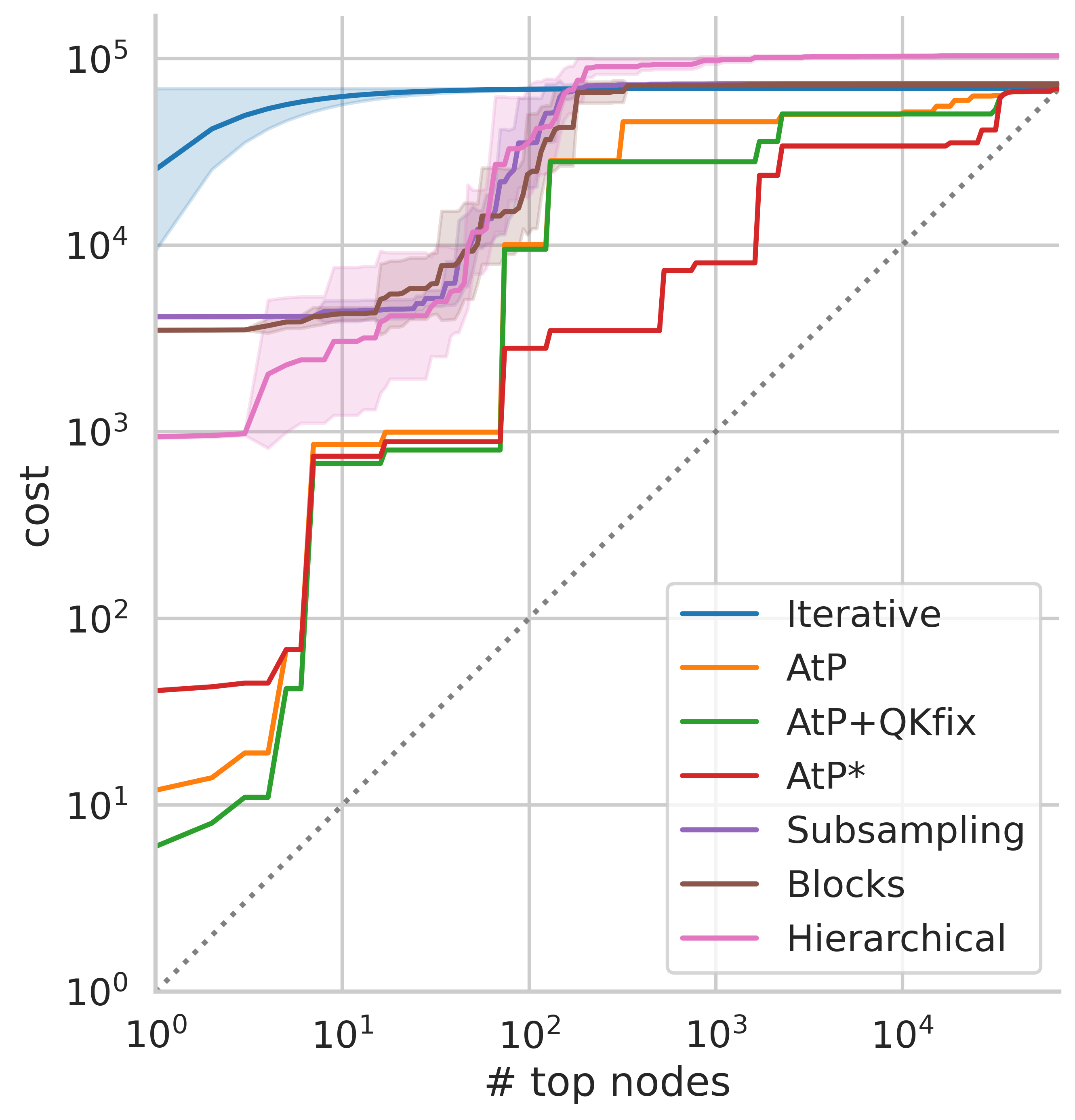}
        \caption{\texttt{RAND-PP} Attention nodes.}
    \end{subfigure}
    \hfill
    \begin{subfigure}[b]{0.48\textwidth}
        \centering

        \includegraphics[width=\textwidth]{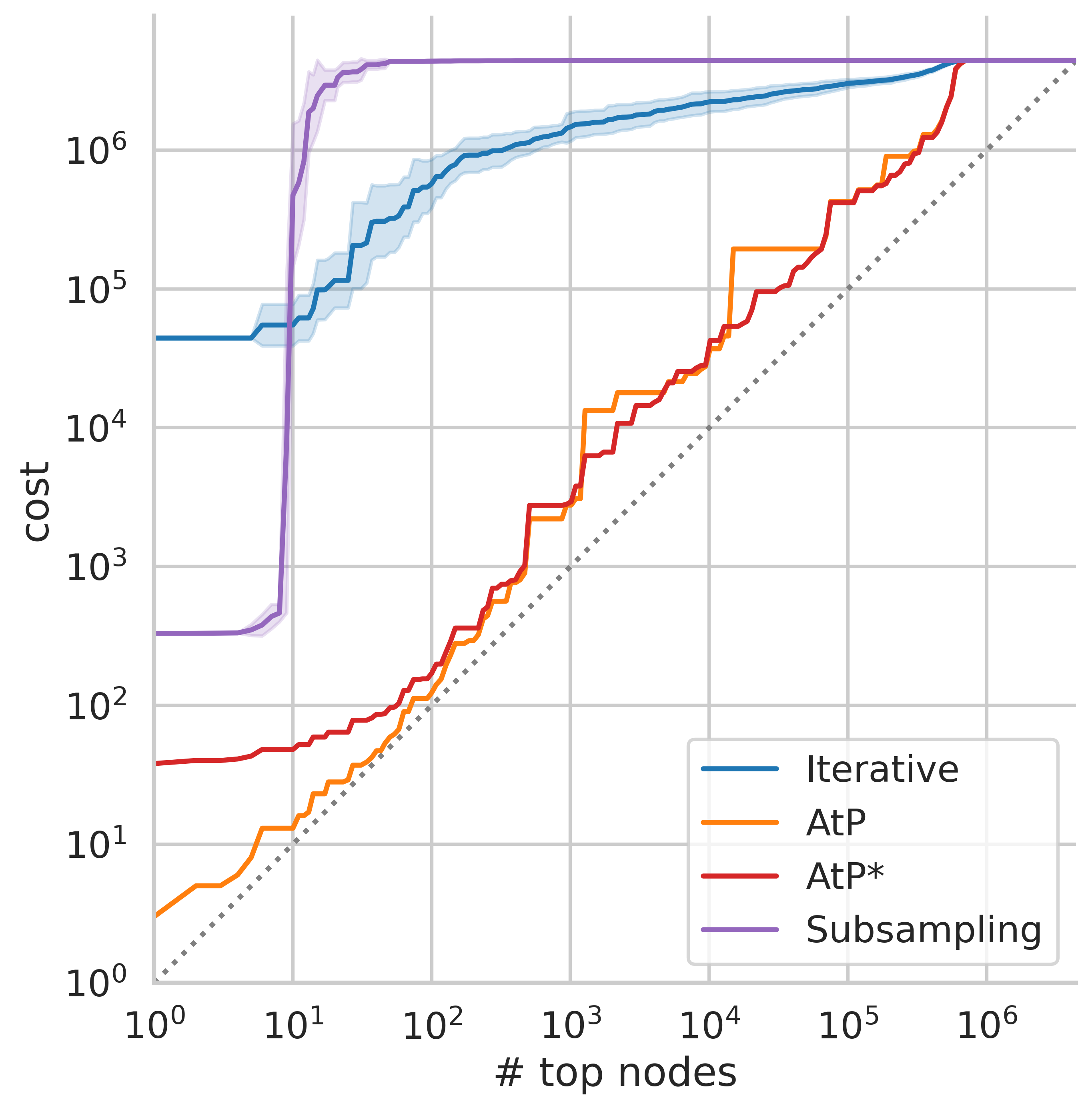}
        \caption{\texttt{A-AN} MLP neurons.}
    \end{subfigure}
    \hfill
    \begin{subfigure}[b]{0.48\textwidth}
        \centering
        \includegraphics[width=\textwidth]{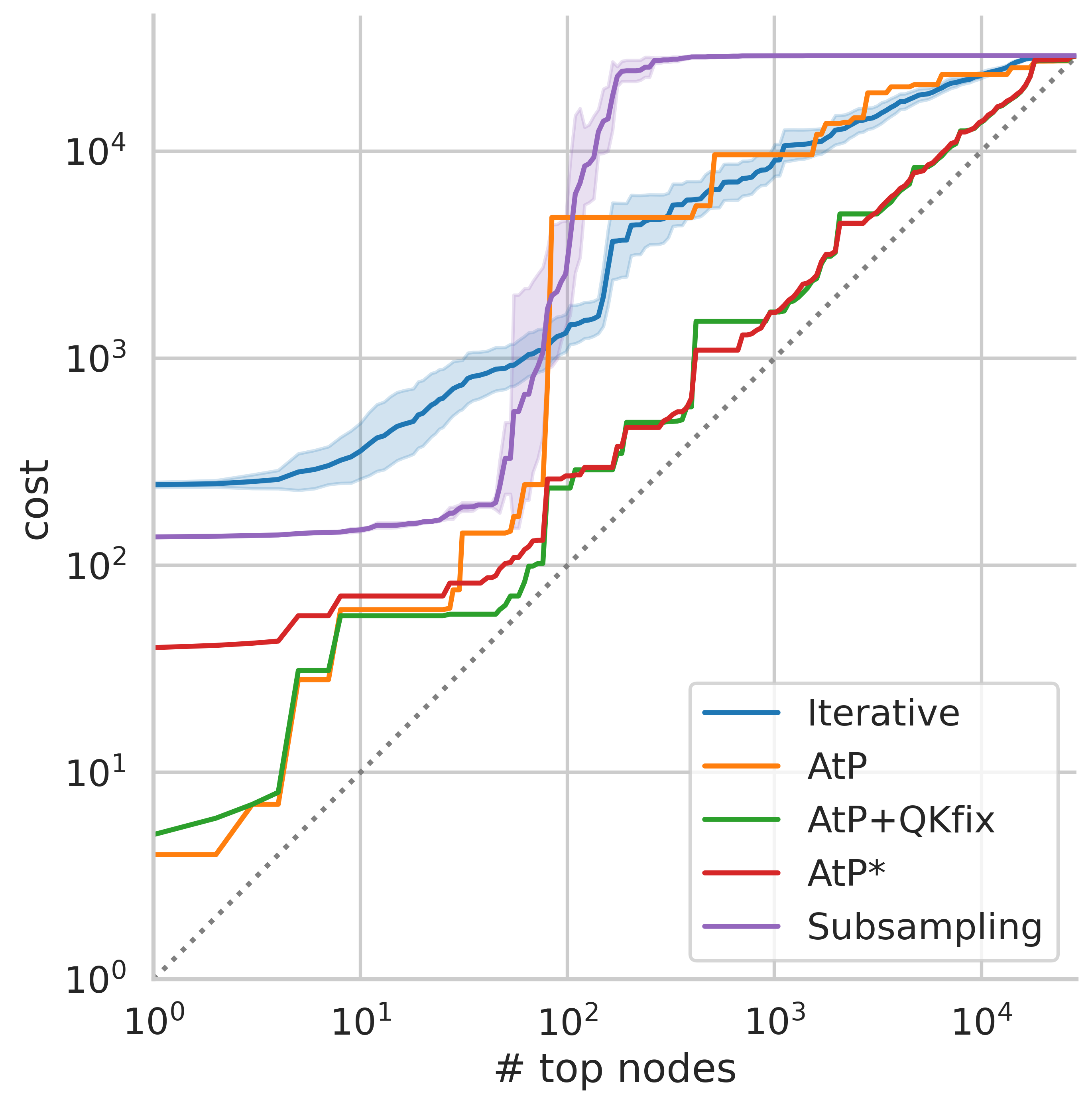}
        \caption{\texttt{IOI} Attention nodes.}
    \end{subfigure}
    \caption{Costs of finding the most causally-important nodes in Pythia-12B using different methods, on a random prompt pair (see \Cref{tab:single_prompt_pairs}) and on distributions. The shading indicates geometric standard deviation. Cost is measured in forward passes, or forward passes per prompt pair in the distributional case.
    }
    \label{fig:rand_and_distr_result}
\end{figure}

\begin{figure}
    % \ContinuedFloat
    \centering
    \begin{subfigure}[b]{0.42\textwidth}
        \centering
        \includegraphics[width=\textwidth]{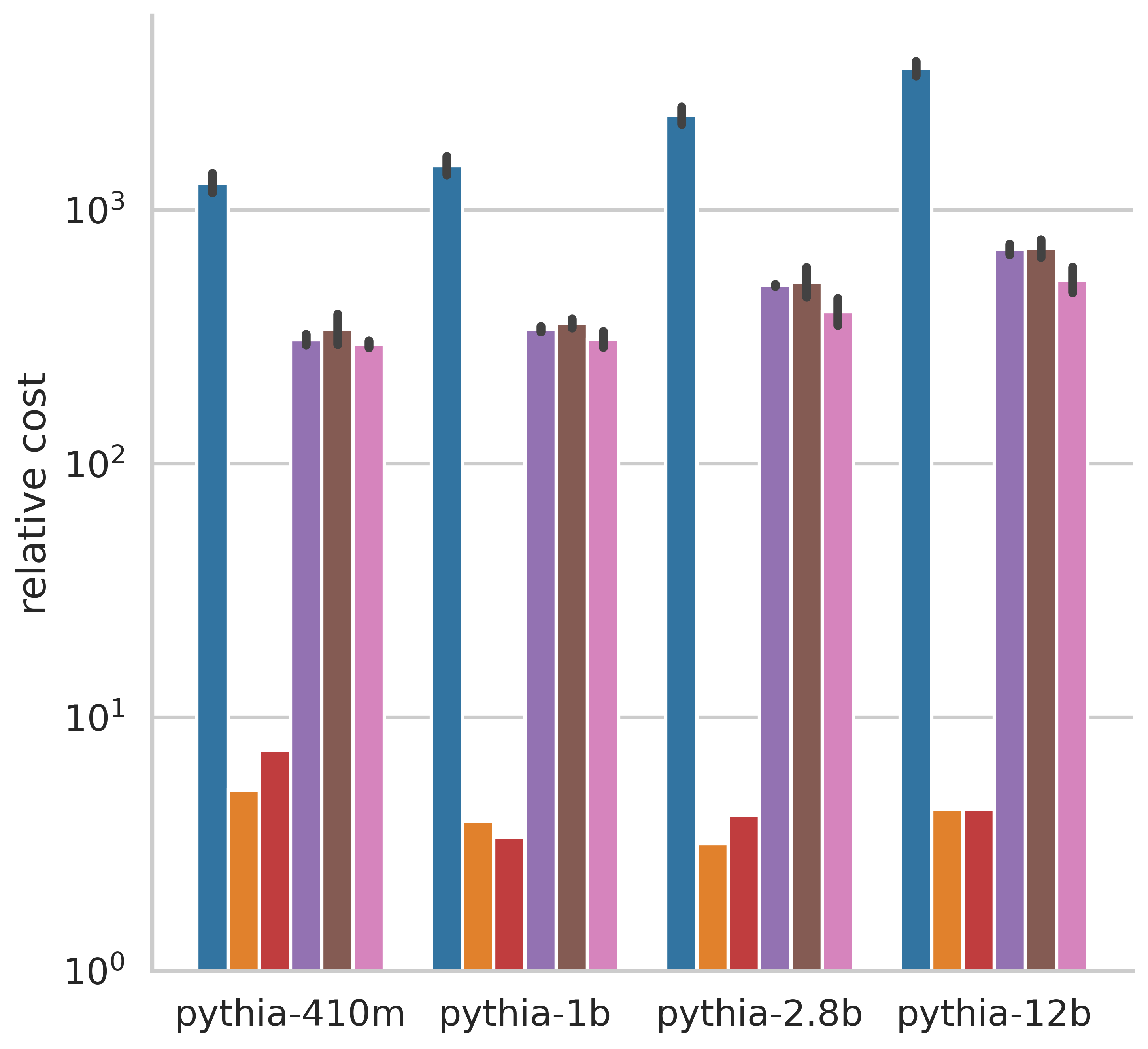}
        \caption{\texttt{RAND-PP} MLP neurons.}
    \end{subfigure}
    \hfill
    \begin{subfigure}[b]{0.14\textwidth}
        \centering
        \includegraphics[width=\textwidth]{legend.png}
    \end{subfigure}
    \hfill
    \begin{subfigure}[b]{0.42\textwidth}
        \centering
        \includegraphics[width=\textwidth]{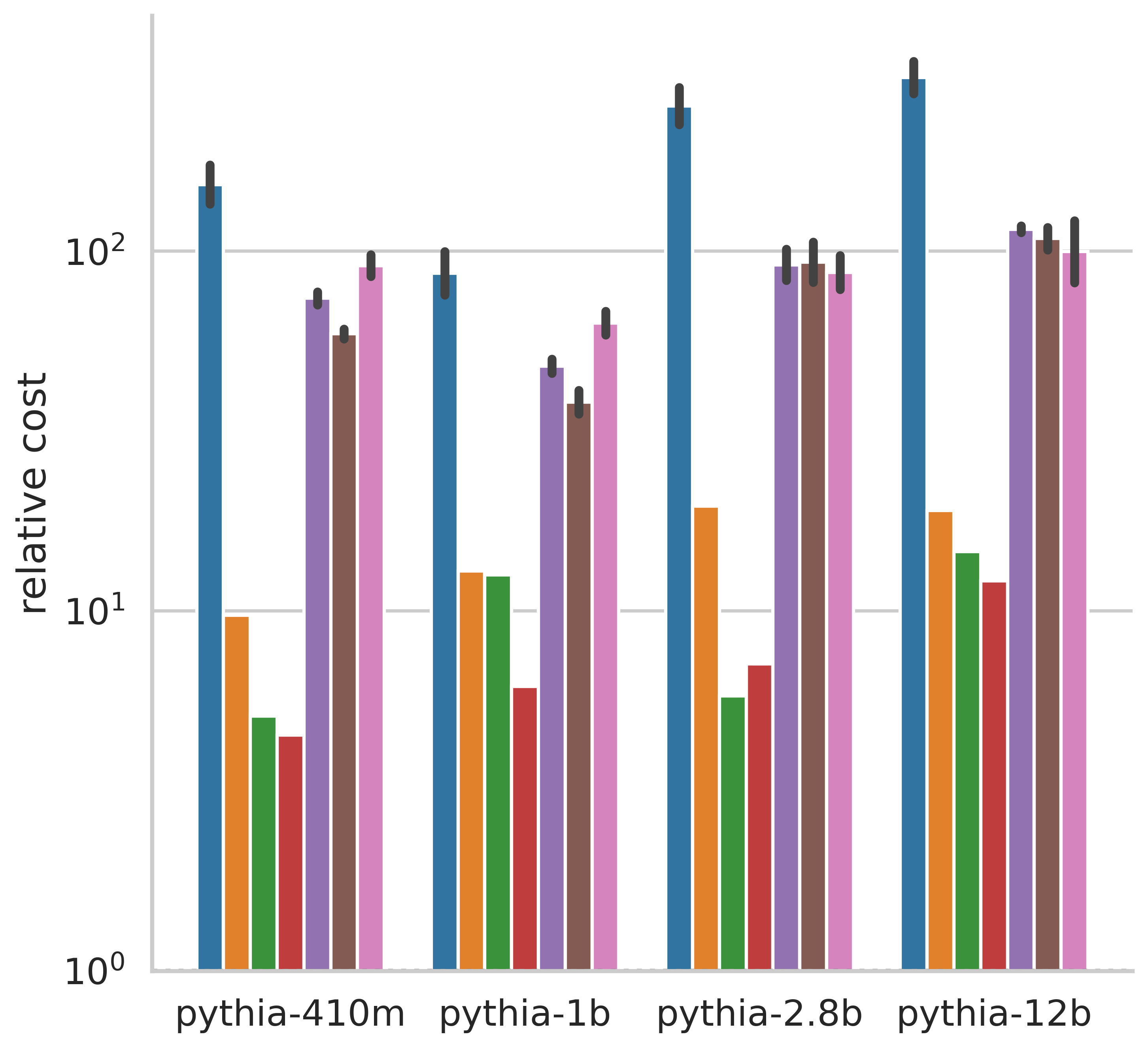}
        \caption{\texttt{RAND-PP} Attention nodes.}
    \end{subfigure}
    \hfill
    \begin{subfigure}[b]{0.42\textwidth}
        \centering
        \includegraphics[width=\textwidth]{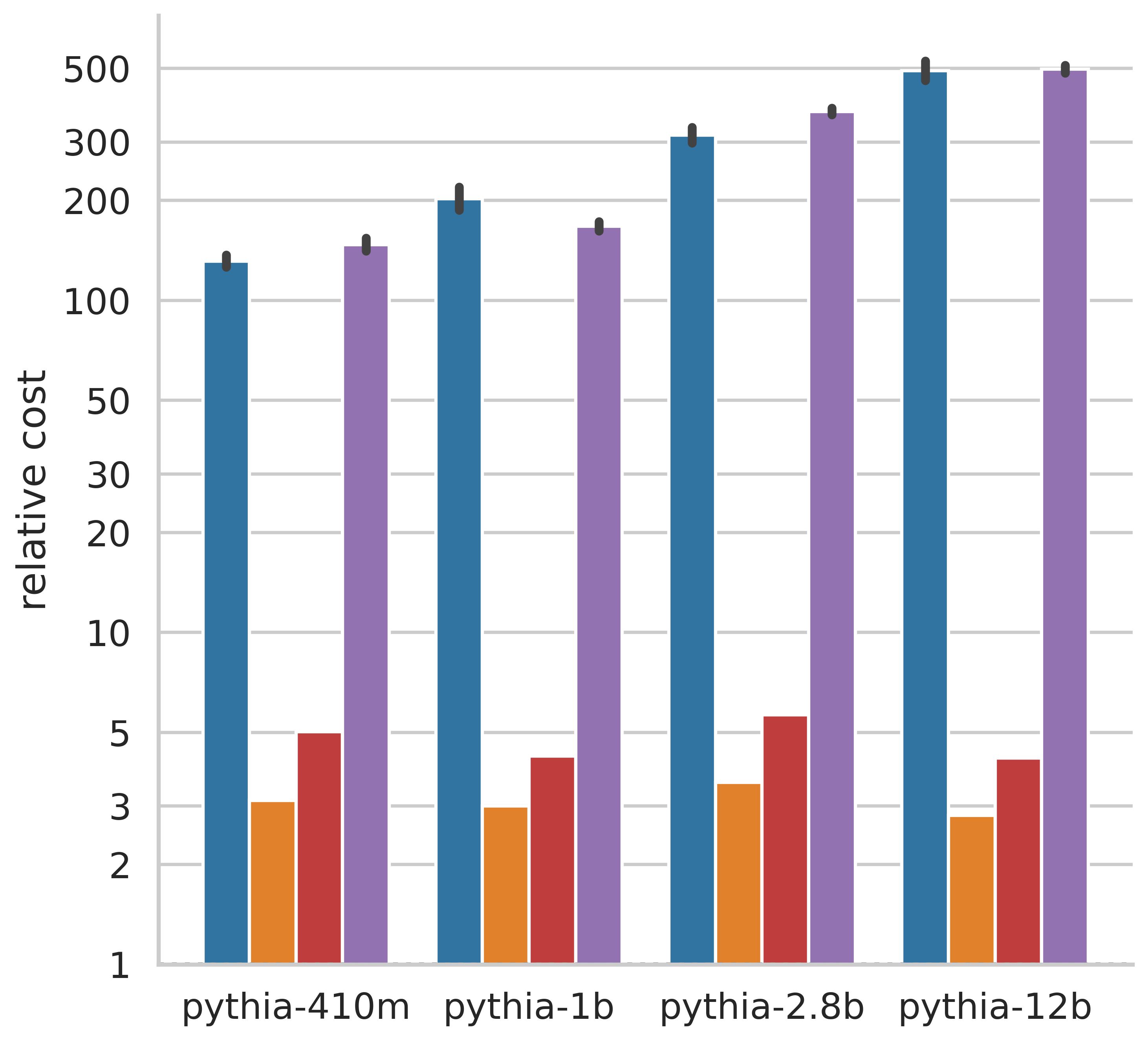}
        \caption{\texttt{A-AN} MLP neurons.}
    \end{subfigure}
    \hfill
    \begin{subfigure}[b]{0.42\textwidth}
        \centering
        \includegraphics[width=\textwidth]{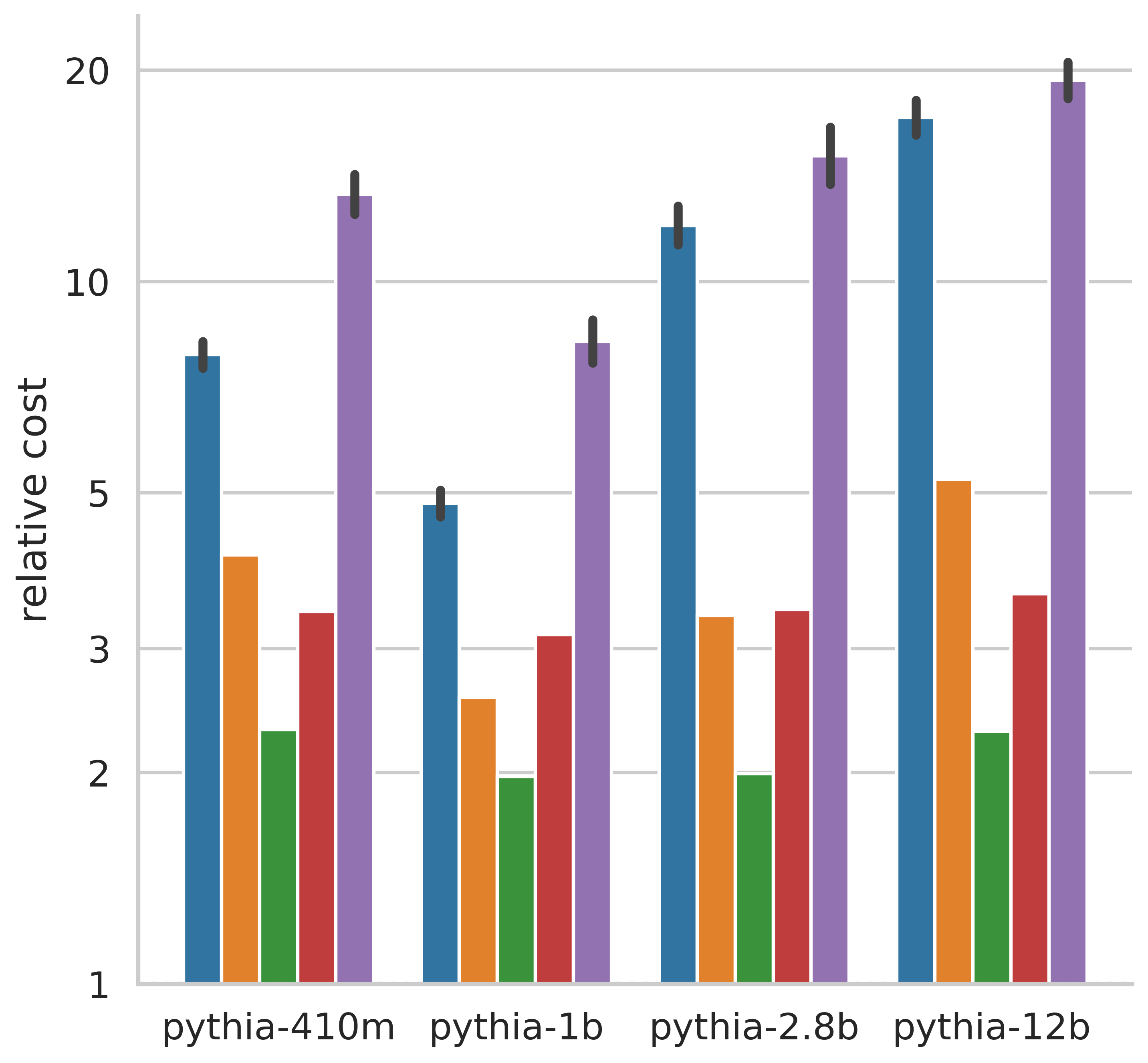}
        \caption{\texttt{IOI} Attention nodes.}
    \end{subfigure}
    \caption{Costs of methods across models, on random prompt pair and on distributions. The costs are relative to having an oracle (and thus verifying nodes in decreasing order of true contribution size); they're aggregated using an inverse-rank-weighted geometric mean. This means they correspond to the area above the diagonal for each curve in \Cref{fig:rand_and_distr_result}.
    }
    \label{fig:rand_and_distr_model_sweep}
\end{figure}

The previous prompt pairs may in fact be the best-case scenarios: the interventions they create will be fairly localized to a specific circuit, and this may make it easy for AtP to approximate the contributions. It may thus be informative to see how the methods generalize to settings where the interventions are less surgical. To do this, we also report results in~\Cref{fig:rand_and_distr_result} (top) and~\Cref{fig:rand_and_distr_model_sweep} on a random prompt pair chosen from a non-copyright-protected section of The Pile~\citep{pile} which we refer to as \texttt{RAND-PP}. The prompt pair was chosen such that Pythia-410M still achieved low loss on both prompts.


We find that AtP/AtP* is only somewhat less effective here; this provides tentative evidence that the strong performance of AtP/AtP* isn't reliant on the clean prompt using a particularly crisp circuit, or on the noise prompt being a precise control.

\paragraph{Distributions}\label{sec:distr}

Causal attribution is often of most interest when evaluated across a distribution, as laid out in \Cref{sec:background}. Of the methods, AtP, AtP*, and Subsampling scale reasonably to distributions; the former 2 because they're inexpensive so running them $|\promptPairDist|$ times is not prohibitive, and Subsampling because it intrinsically averages across the distribution and thus becomes proportionally cheaper relative to the verification via activation patching. In addition, having a distribution enables a more performant Iterative method, as described in \Cref{par:iterative}.

We present a comparison of these methods on 2 distributional settings. The first is a reduced version of IOI~\citep{wang2022interpretability} on 6 names, resulting in $6\times5\times4=120$ prompt pairs, where we evaluate \AttentionNodes{}. The other distribution prompts the model to output an indefinite article `\texttt{ a}' or `\texttt{ an}', where we evaluate \NeuronNodes{}. See~\Cref{app:prompt_distributions} for details on constructing these distributions. Results are shown in \Cref{fig:rand_and_distr_result} for Pythia 12B, and in \Cref{fig:rand_and_distr_model_sweep} across models. The results show that AtP continues to perform well, especially with the QK fix; in addition, the cancellation failure mode tends to be sensitive to the particular input prompt pair, and as a result, averaging across a distribution diminishes the benefit of GradDrops.

An implication of Subsampling scaling well to this setting is that diagnostics may give reasonable confidence in not missing false negatives with much less overhead than in the single-prompt-pair case; this is illustrated in \Cref{fig:diagnostic_main}.

\section{Discussion}\label{sec:discuss}

\subsection{Limitations}

\paragraph{Prompt pair distributions}
We only considered a small set of prompt pair distributions, which often were limited to a single prompt pair, since evaluating the ground truth can be quite costly. While we aimed to evaluate on distributions that are reasonably representative, our results may not generalize to other distributions.

\paragraph{Choice of Nodes $\Nodes$}
In the \NeuronNodes{} setting, we took MLP neurons as our fundamental unit of analysis. However, there is mounting evidence~\citep{bricken2023monosemanticity} that the decomposition of signals into neuron contributions does not correspond directly to a semantically meaningful decomposition. Instead, achieving such a decomposition seems to require finding the right set of directions in neuron activation space~\citep{bricken2023monosemanticity,gurnee2023finding} -- which we viewed as being out of scope for this paper. In~\cref{sec:extensions} we further discuss the applicability of AtP to sparse autoencoders, a method of finding these decompositions.

More generally, we only considered relatively fine-grained nodes, because this is a case where very exhaustive verification is prohibitively expensive, justifying the need for an approximate, fast method. \cite{neel2022attribution} speculate that AtP may perform worse on coarser components like full layers or entire residual streams, as a larger change may have more of a non-linear effect. There may still be benefit in speeding up such an analysis, particularly if the context length is long -- our alternative methods may have something to offer here, though we leave investigation of this to future work.

It is popular in the literature to do Activation Patching with these larger components, with short contexts -- this doesn't pose a performance issue, and so our work would not provide any benefit here.

\paragraph{Caveats of $\contrib(\node)$ as importance measure}
In this work we took the ground truth of activation patching, as defined in~\Cref{eq:ground_truth}, as our evaluation target.
As discussed by~\citet{mcgrath2023hydra}, \Cref{eq:ground_truth} often significantly disagrees with a different evaluation target, the ``direct effect'', by putting lower weight on some contributions when later components would shift their behaviour to compensate for the earlier patched component. In the worst case this could be seen as producing additional false negatives not accounted for by our metrics. To some degree this is likely to be mitigated by the GradDrop formula in \cref{eq:contrib_atp_gd}, which will include a term dropping out the effect of that downstream shift.

However, it is also questionable whether we need to concern ourselves with finding high-direct-effect nodes. For example, direct effect is easy to efficiently compute for all nodes, as explored by \citet{nostalgebraist2020logit} -- so there is no need for fast approximations like AtP if direct effect is the quantity of interest. This ease of computation is no free lunch, though, because direct effect is also more limited as a tool for finding causally important nodes: it would not be able to locate any nodes that contribute only instrumentally to the circuit rather than producing its output. For example, there is no direct effect from nodes at non-final token positions. We discuss the direct effect further in \cref{sec:cancellation} and \cref{app:graddrop}.

Another nuance of our ground--truth definition occurs in the distributional setting. Some nodes may have a real and significant effect, but only on a single clean prompt (e.g. they only respond to a particular name in \texttt{IOI}\footnote{We did observe this particular behavior in a few instances.} or object in \texttt{A-AN}). Since the effect is averaged over the distribution, the ground truth will not assign these nodes large causal importance. Depending on the goal of the practitioner this may or may not be desirable. %We briefly evaluate the impact of alternative metrics $\metric$ in~\Cref{app:metrics}.

\paragraph{Effect size versus rank estimation}
When evaluating the performance of various estimators, we focused on evaluating the relative rank of estimates, since our main goal was to identify important components (with effect size only instrumentally useful to this end), and we assumed a further verification step of the nodes with highest estimated effects one at a time, in contexts where knowing effect size is important. Thus, we do not present evidence about how closely the estimated effect magnitudes from AtP or AtP* match the ground truth. Similarly, we did not assess the prevalence of false positives in our analysis, because they can be filtered out via the verification process. Finally, we did not compare to past manual interpretability work to check whether our methods find the same nodes to be causally important as discovered by human researchers, as done in prior work~\citep{conmy2023automated,syed2023attribution}.

\paragraph{Other LLMs}
While we think it likely that our results on the Pythia model family~\citep{biderman2023pythia} will transfer to other LLM families, we cannot rule out qualitatively different behavior without further evidence, especially on SotA--scale models or models that significantly deviate from the standard decoder-only transformer architecture.

\subsection{Extensions/Variants}\label{sec:extensions}

\paragraph{Edge Patching}
While we focus on computing the effects of individual nodes, edge activation patching can give more fine-grained information about which paths in the computational graph matter. However, it suffers from an even larger blowup in number of forward passes if done naively. Fortunately, AtP is easy to generalize to estimating the effects of edges between nodes~\citep{neel2022attribution,syed2023attribution}, while AtP* may provide further improvement. We discuss edge-AtP, and how to efficiently carry over the insights from AtP*, in \cref{app:edge_atp}.

\paragraph{Coarser nodes $\Nodes$}
We focused on fine-grained attribution, rather than full layers or sliding windows~\citep{meng2023locating,geva2023dissecting}. In the latter case there's less computational blowup to resolve, but for long contexts there may still be benefit in considering speedups like ours; on the other hand, they may be less linear, thus favouring other methods over AtP*. We leave investigation of this to future work.

\paragraph{Layer normalization}
\citet{neel2022attribution} observed that AtP's approximation to layer normalization may be a worse approximation when it comes to patching larger/coarser nodes: on average the patched and clean activations are likely to have similar norm, but may not have high cosine-similarity. They recommend treating the denominator in layer normalization as fixed, e.g. using a stop-gradient operator in the implementation. In \cref{app:layernorm} we explore the effect of this, and illustrate the behaviour of this alternative form of AtP. It seems likely that this variant would indeed produce better results particularly when patching residual-stream nodes -- but we leave empirical investigation of this to future work. % In general, we recommend freezing the LayerNorm scale before differentiating, if possible. 

\paragraph{Denoising}
Denoising~\citep{meng2023locating,lieberum2023does} is a different use case for patching, which may produce moderately different results: the difference is that each forward pass is run on $\xnoise$ with the activation to patch taken from $\xclean$ --- colloquially, this tests whether the patched activation is sufficient to recover model performance on $\xclean$, rather than necessary. We provide some preliminary evidence to the effect of this choice in~\Cref{app:metrics} but leave a more thorough investigation to future work.

\paragraph{Other forms of ablation}
Further, in some settings it may be of interest to do mean-ablation, or even zero-ablation, and our tweaks remain applicable there; the random-prompt-pair result suggests AtP* isn't overly sensitive to the noise distribution, so we speculate the results are likely to carry over.

\subsection{Applications}\label{sec:applications}

\paragraph{Automated Circuit Finding}
A natural application of the methods we discussed in this work is the automatic identification and localization of sparse subgraphs or `circuits'~\citep{cammarata2020thread}. A variant of this was already discussed in concurrent work by~\citet{syed2023attribution} who combined edge attribution patching with the ACDC algorithm~\citep{conmy2023automated}. As we mentioned in the edge patching discussion, AtP* can be generalized to edge attribution patching, which may bring additional benefit for automated circuit discovery.

Another approach is to learn a (probabilistic) mask over nodes, similar to~\citet{louizos2018learning,decao2021sparse}, where the probability scales with the currently estimated node contribution~$\contrib(\node)$. For that approach, a fast method to estimate all node effects given the current mask probabilities could prove vital.

\paragraph{Sparse Autoencoders}
Recently there has been increased interest by the community in using sparse autoencoders (SAEs) to construct disentangled sparse representations with potentially more semantic coherence than transformer-native units such as neurons~\citep{cunningham2023sparse,bricken2023monosemanticity}. 
SAEs usually have a lot more nodes than the corresponding transformer block they are applied to. This could pose a larger problem in terms of the activation patching effects, making the speedup of AtP* more valuable. However, due to the sparseness of the SAE, on a given forward pass the effect of most features will be zero. For example, some successful SAEs by~\citet{bricken2023monosemanticity} have 10-20 active features for 500 neurons for a given token position, which reduces the number of nodes by 20-50x relative to the MLP setting, increasing the scale at which existing iterative methods remain practical. It is still an open research question, however, what degree of sparsity is feasible with tolerable reconstruction error for practically relevant or SOTA--scale models, where the methods discussed in this work may become more important again.

\paragraph{Steering LLMs}
AtP* could be used to discover single nodes in the model that can be leveraged for targeted inference time interventions to control the model's behavior. In contrast to previous work~\citep{li2023inferencetime,turner2023activation,zou2023representation} it might provide more localized interventions with less impact on the rest of the model's computation. One potential exciting direction would be to use AtP* (or other gradient-based approximations) to see which sparse autoencoder features, if activated, would have a significant effect.

\subsection{Recommendation}\label{sec:recommend}

Our results suggest that if a practitioner is trying to do fast causal attribution, there are 2 main factors to consider: (i) the desired granularity of localization, and (ii) the confidence vs compute tradeoff.

Regarding (i), the desired granularity, smaller components (e.g. MLP neurons or attention heads) are more numerous but more linear, likely yielding better results from gradient-based methods like AtP. We are less sure AtP will be a good approximation if patching layers or sliding windows of layers, and in this case practitioners may want to do normal patching. If the number of forward passes required remains prohibitive (e.g. a long context times many layers, when doing per token $\times$ layer patching), our other baselines may be useful. For a single prompt pair we particularly recommend trying Blocks, as it's easy to make sense of; for a distribution we recommend Subsampling because it scales better to many prompt pairs.

Regarding (ii), the confidence vs compute tradeoff, depending on the application, it may be desirable to run AtP as an activation patching prefilter followed by running the diagnostic to increase confidence. On the other hand, if false negatives aren't a big concern then it may be preferable to skip the diagnostic -- and if false positives aren't either, then in certain cases practitioners may want to skip activation patching verification entirely. In addition, if the prompt pair distribution does not adequately highlight the specific circuit/behaviour of interest, this may also limit what can be learned from any localization methods.

If AtP is appropriate, our results suggest the best variant to use is probably AtP* for single prompt pairs, AtP+QKFix for \AttentionNodes{} on distributions, and AtP for \NeuronNodes{} (or other sites that aren't immediately before a nonlinearity) on distributions.

Of course, these recommendations are best-substantiated in settings similar to those we studied: focused prompt pairs / distribution, attention node or neuron sites, nodewise attribution, measuring cross-entropy loss on the clean-prompt next token. If departing from these assumptions we recommend looking before you leap.

\section{Related work}

\paragraph{Localization and Mediation Analysis}
This work is concerned with identifying the effect of all (important) nodes in a causal graph~\citep{pearl2000causality}, in the specific case where the graph represents a language model's computation. A key method for finding important intermediate nodes in a causal graph is intervening on those nodes and observing the effect, which was first discussed under the name of causal mediation analysis by~\citet{Robins1992IdentifiabilityAE,pearl2001direct}.

\paragraph{Activation Patching}

In recent years there has been increasing success at applying the ideas of causal mediation analysis to identify causally important nodes in deep neural networks, in particular via the method of activation patching, where the output of a model component is intervened on. This technique has been widely used by the community and successfully applied in a range of contexts~\citep{olsson2022context,vig2020investigating,soulos-etal-2020-discovering,meng2023locating,wang2022interpretability, hase2023does, lieberum2023does, conmy2023automated, hanna2023does, geva2023dissecting, huang2023rigorously, tigges2023linear, merullo2023circuit, mcdougall2023copy, goldowskydill2023localizing, stolfo2023mechanistic, feng2023language, hendel2023incontext, todd2023function, cunningham2023sparse,finlayson-etal-2021-causal, nanda2023factfinding}. 

\citet{chan2022causal} introduce causal scrubbing, a generalized algorithm to verify a hypothesis about the internal mechanism underlying a model's behavior, and detail their motivation behind performing noising and resample ablation rather than denoising or using mean or zero ablation -- they interpret the hypothesis as implying the computation is invariant to some large set of perturbations, so their starting-point is the clean unperturbed forward pass.\footnote{Our motivation for focusing on noising rather than denoising was a closely related one -- we were motivated by automated circuit discovery, where gradually noising more and more of the model is the basic methodology for both of the approaches discussed in \cref{sec:applications}.} 

Another line of research concerning formalizing causal abstractions focuses on finding and verifying high-level causal abstractions of low-level variables~\citep{geiger2020neural,geiger2021causal,geiger2022inducing,geiger2023causal}. See \citet{jenner2023comparison} for more details on how these different frameworks agree and differ. In contrast to those works, we are chiefly concerned with identifying the important low-level variables in the computational graph and are not investigating their semantics or potential groupings of lower-level into higher-level variables.

In addition to causal mediation analysis, intervening on node activations in the model forward pass has also been studied as a way of steering models towards desirable behavior~\citep{rimsky2023steering,zou2023representation,turner2023activation,jorgensen2023improving,li2023inferencetime,belrose2023leace}.

\paragraph{Attribution Patching / Gradient-based Masking}
While we use the resample--ablation variant of AtP as formulated in~\citet{neel2022attribution}, similar formulations have been used in the past to successfully prune deep neural networks~\citep{figurnov2016perforatedcnns,molchanov2017pruning,michel2019sixteen}, or even identify causally important nodes for interpretability~\citep{decao2021sparse}. Concurrent work by~\citet{syed2023attribution} also demonstrates AtP can help with automatically finding causally important circuits in a way that agrees with previous manual circuit identification work. In contrast to~\citet{syed2023attribution}, we provide further analysis of AtP's failure modes, give improvements in the form of AtP$^*$, and evaluate both methods as well as several baselines on a suite of larger models against a ground truth that is independent of human researchers' judgement.

\section{Conclusion}
In this paper, we have explored the use of attribution patching for node patch effect evaluation. We have compared attribution patching with alternatives and augmentations, characterized its failure modes, and presented reliability diagnostics. We have also discussed the implications of our contributions for other settings in which patching can be of interest, such as circuit discovery, edge localization, coarse-grained localization, and causal abstraction.

Our results show that AtP* can be a more reliable and scalable approach to node patch effect evaluation than alternatives. However, it is important to be aware of the failure modes of attribution patching, such as cancellation and saturation. We explored these in some detail, and provided mitigations, as well as recommendations for diagnostics to ensure that the results are reliable.

We believe that our work makes an important contribution to the field of mechanistic interpretability and will help to advance the development of more reliable and scalable methods for understanding the behavior of deep neural networks.

\section{Author Contributions}

J\'anos Kram\'ar was research lead, and Tom Lieberum was also a core contributor – both were highly involved in most aspects of the project. Rohin Shah and Neel Nanda served as advisors and gave feedback and guidance throughout.

\bibliography{main}
\clearpage
\appendix

\section{Method details}

\subsection{Baselines}

\subsubsection{Properties of Subsampling}\label{app:subsampling}
Here we prove that the subsampling estimator $\hat{\inter}_{\text{SS}}(\node)$ from~\Cref{par:subsampling} is unbiased in the case of no interaction effects. Furthermore, assuming a simple interaction model, we show the bias of $\hat{\inter}_{\text{SS}}(\node)$ is $p$ times the total interaction effect of $\node$ with other nodes.
We assume a pairwise interaction model. That is, given a set of nodes $\subsetNodes$, we have
\begin{align}
    \inter(\subsetNodes;x) &= \sum_{\node\in\subsetNodes} \inter(\node;x) + \sum_{\substack{\node, \node'\in \subsetNodes\\\node\neq \node}} \sigma_{\node,\node'}(x)
\end{align}

\noindent with fixed constants $\sigma_{\node, \node'}(x)\in\mathbb{R}$ for each prompt pair $x\in\operatorname{support}(\promptPairDist)$. Let $\sigma_{\node,\node'}=\expect{x\sim \promptPairDist}{\sigma_{\node,\node'}(x)}$.

Let $p$ be the probability of including each node in a given $\subsetNodes$ and let $M$ be the number of node masks sampled from $\operatorname{Bernoulli}^{|\Nodes|}(p)$ and prompt pairs $x$ sampled from $\promptPairDist$. Then,

\begin{subequations}
\begin{align}
    \expect{}{\hat{\inter}_{\text{SS}}(\node)}
    &= \expect{}{\frac{1}{|\subsetNodes^+(\node)|}\sum_{k=1}^{|\subsetNodes^+(\node)|} \inter(\subsetNodes^+_k(\node);x_k^+) - \frac{1}{|\subsetNodes^-(\node)|}\sum_{k=1}^{|\subsetNodes^-(\node)|} \inter(\subsetNodes^-_k(\node);x_k^-) }\\
    &= \expect{}{\expect{}{\frac{1}{|\subsetNodes^+(\node)|}\sum_{k=1}^{|\subsetNodes^+(\node)|} \inter(\subsetNodes^+_k(\node);x_k^+) - \frac{1}{|\subsetNodes^-(\node)|}\sum_{k=1}^{|\subsetNodes^-(\node)|} \inter(\subsetNodes^-_k(\node);x_k^-)\middle||\subsetNodes^+(\node)|}}\\
    &= \expect{}{\expect{}{\frac{|\subsetNodes^+(\node)|}{|\subsetNodes^+(\node)|}\expect{}{ \inter(\subsetNodes_1;x_1)\middle|\node\in\subsetNodes_1} - \frac{|\subsetNodes^-(\node)|}{|\subsetNodes^-(\node)|}\expect{}{ \inter(\subsetNodes_1;x_1)\middle|\node\not\in\subsetNodes_1}\middle||\subsetNodes^+(\node)|}}\\
    &= \expect{}{ \inter(\subsetNodes_1;x_1)\middle|\node\in\subsetNodes_1} - \expect{}{ \inter(\subsetNodes_1;x_1)\middle|\node\not\in\subsetNodes_1}\\
    &= \contrib(\node) + \expect{}{\sum_{n'\neq\node}\ind{n'\in\subsetNodes_1}\left(\contrib(\node') + \sigma_{\node\node'} + \frac12 \sum_{\node'' \not\in \{\node', \node\}} \ind{\node'\in\subsetNodes_1}\sigma_{\node'\node''}\middle|\node\in\subsetNodes_1\right)} \\
    &\quad- \expect{}{\sum_{n'\neq\node}\ind{n'\in\subsetNodes_1}\left(\contrib(\node') + \frac12 \sum_{\node'' \not\in \{\node', \node\}} \ind{\node'\in\subsetNodes_1}\sigma_{\node'\node''}\right)\middle|\node\not\in\subsetNodes_1}\\
    &= \contrib(\node) + p\sum_{\node'\neq\node} \sigma_{\node\node'} \label{eq:ss_math_result}
\end{align}
\end{subequations}

In~\Cref{eq:ss_math_result}, we observe that if the interaction terms $\sigma_{\node\node'}$ are all zero, the estimator is unbiased. Otherwise, the bias scales both with the sum of interaction effects and with $p$, as expected.

\subsubsection{Pseudocode for Blocks and Hierarchical baselines}\label{app:baselines}

In \Cref{alg:blocks} we detail the Blocks baseline algorithm. As explained in \Cref{par:blocks}, it comes with a tradeoff in its ``block size'' hyperparameter $B$: a small block size requires a lot of time to evaluate all the blocks, while a large block size means many irrelevant nodes to evaluate in each high-contribution block.

\begin{algorithm}
\caption{Blocks algorithm for causal attribution.}
\label{alg:blocks}
\begin{algorithmic}[1]
\Require{block size $B$, compute budget $M$, nodes $\Nodes=\{\node_i\}$, prompts $\xclean,\,\xnoise$, intervention function $\Tilde\inter:\subsetNodes \mapsto \inter(\subsetNodes;\xclean, \xnoise)$}
\State $\mathrm{numBlocks}\gets \lceil|\Nodes|/B\rceil$
\State $\pi\gets\operatorname{shuffle}\left(\left\{\left\lfloor \mathrm{numBlocks}\cdot i B/|\Nodes|\right\rfloor \mid i\in\{0, \dots, |N|-1\}\right\}\right)$\Comment{Assign each node to a block.}
\For{$i\gets 0 \textrm{ to numBlocks}-1$}
    \State $\textrm{blockContribution}[i]\gets |\Tilde\inter(\pi^{-1}(\{i\}))|$ \Comment{$\pi^{-1}(\{i\}) := \{\node:\, \pi(\node) = i\mid\node\in\Nodes\})$}
\EndFor
\State $\mathrm{spentBudget}\gets M - \mathrm{numBlocks}$
\State $\mathrm{topNodeContribs}\gets \operatorname{CreateEmptyDictionary}()$
\ForAll{$i\in \{0 \textrm{ to numBlocks}-1\}$ in decreasing order of $\textrm{blockContribution}[i]$}
    \ForAll{$\node\in\pi^{-1}(\{i\})$} \Comment{Eval all nodes in block.}
        \If{$\mathrm{spentBudget}< M$}
            \State $\mathrm{topNodeContribs}[\node]\gets \mid\Tilde\inter(\{\node\})|$
            \State $\mathrm{spentBudget}\gets \mathrm{spentBudget} + 1$
        \Else
            \State \Return{topNodeContribs}
        \EndIf
    \EndFor
\EndFor
\State\Return{topNodeContribs}
\end{algorithmic}
\end{algorithm}

The Hierarchical baseline algorithm aims to resolve this tradeoff, by using small blocks, but grouped into superblocks so it's not necessary to traverse all the small blocks before finding the key nodes. In \Cref{alg:hier} we detail the hierarchical algorithm in its iterative form, corresponding to batch size 1.

One aspect that might be surprising is that on line \ref{alg:hier:prio_min}, we ensure a subblock is never added to the priority queue with higher priority than its ancestor superblocks. The reason for doing this is that in practice we use batched inference rather than patching a single block at a time, so depending on the batch size, we do evaluate blocks that aren't the highest-priority unevaluated blocks, and this might impose a significant delay in when some blocks are evaluated. In order to reduce this dependence on the batch size hyperparameter, line \ref{alg:hier:prio_min} ensures that every block is evaluated at most $L$ batches later than it would be with batch size 1. 

\begin{algorithm}
\caption{Hierarchical algorithm for causal attribution, in iterative form. In practice we do additional batching rather than evaluating a single block at a time on line \ref{alg:hier:inter}.}
\label{alg:hier}
\begin{algorithmic}[1]
\Require{branching factor $B$, num levels $L$, compute budget $M$, nodes $\Nodes=\{\node_i\}$, intervention function $\inter$}
\State $\mathrm{numTopLevelBlocks}\gets \lceil|\Nodes|/B^L\rceil$
\State $\pi\gets\operatorname{shuffle}\left(\left\{\left\lfloor \mathrm{numTopLevelBlocks}\cdot i B^L/|\Nodes|\right\rfloor \middle| i\in\{0, \dots, |N|-1\}\right\}\right)$
\ForAll{$\node_i\in\Nodes$}
  \State $(d_{L-1},d_{L-2},\dots,d_0)\gets \text{zero-padded final }L$ base-$B$ digits of $\pi_i$
  \State $\mathrm{address}(\node_i)=(\lfloor\pi_i/B^L\rfloor, d_{L-1},\dots,d_0)$
\EndFor
\State $Q\gets\operatorname{CreateEmptyPriorityQueue}()$
\For{$i \gets 0 \textrm{ to numTopLevelBlocks}-1$}
    \State $\operatorname{PriorityQueueInsert}(Q, [i], \infty)$
\EndFor
\State $\mathrm{spentBudget}\gets 0$
\State $\mathrm{topNodeContribs}\gets \operatorname{CreateEmptyDictionary}()$
\Repeat
    \State $(\mathrm{addressPrefix},\mathrm{priority})\gets \operatorname{PriorityQueuePop}(Q)$
    \State $\mathrm{blockNodes} \gets \left\{\node\in\Nodes\middle|\operatorname{StartsWith}(\mathrm{address}(\node),\mathrm{addressPrefix})\right\}$
    \State $\mathrm{blockContribution} \gets |\inter\left(\mathrm{blockNodes}\right)|$\label{alg:hier:inter}
    \State $\mathrm{spentBudget}\gets \mathrm{spentBudget} + 1$
    \If{$\mathrm{blockNodes}=\{\node\}$ for some $\node\in\Nodes$}
        \State $\mathrm{topNodeContribs}[\node] \gets \mathrm{blockContribution}$
    \Else
        \For{$i\gets 0 \textrm{ to }B-1$}
            \If{$\{\node\in\mathrm{blockNodes}|\operatorname{StartsWith}(\mathrm{address}(\node), \mathrm{addressPrefix}+[i]\} \not=\emptyset$}
                \State $\operatorname{PriorityQueueInsert}(Q, \mathrm{addressPrefix}+[i], \min(\mathrm{blockContribution},\mathrm{priority}))$\label{alg:hier:prio_min}
            \EndIf
        \EndFor
    \EndIf
\Until{$\mathrm{spentBudget}=M$ or $\operatorname{PriorityQueueEmpty}(Q)$}
\State \Return{topNodeContribs}
\end{algorithmic}
\end{algorithm}

\subsection{AtP improvements}

\subsubsection{Pseudocode for corrected AtP on attention keys}\label{app:kfix}

As described in~\Cref{par:kfix}, computing~\Cref{eq:kfix} naïvely for all nodes requires $\operatorname{O}(\numTokens^{3})$
flops at each attention head and prompt pair. Here we give a more efficient algorithm running in $\operatorname{O}(\numTokens^{2})$.
In addition to keys, queries and attention probabilities, we now also cache attention logits (pre-softmax scaled key-query dot products).

\begin{sloppypar}
We define $\attnLogits_{\text{patch}}^t(\node^q)$ and $\Delta_{t}\attnLogits(\node^q)$ analogously to~\Cref{eq:patch_attn_key,eq:delta_attn_key}. For brevity we can also define $\attnLogits_{\text{patch}}(\node^q)_t:=\attnLogits^t_{\text{patch}}(\node^q)_t$ and $\Delta \attnLogits(\node^q)_t:=\Delta_t \attnLogits(\node^q)_t$, since the aim with this algorithm is to avoid having to separately compute effects of $\doGets{\node^k_{t}}{\node^k_{t}(\xnoise)}$ on any other component of $\attnLogits$ than the one for key node $\node^k_t$.
\end{sloppypar}

Note that, for a key $\node^k_t$ at position $t$ in the sequence, the proportions of the non-$t$ components
of $\attn(\node^q)_{t}$ do not change when $\attnLogits(\node^q)_{t}$
is changed, so $\Delta_{t}\attn(\node^q)$ is actually $\mathrm{onehot}(t)-\attn(\node^q)$ multiplied by some scalar $s_t$; specifically, to get the
right attention weight on $n^k_t$, the scalar must be $s_t:=\frac{\Delta\attn(\node^q)_{t}}{1-\attn(\node^q)_{t}}$.
Additionally, we have $\log\left(\frac{\attn_{\text{patch}}^t(\node^q)_t}{1-\attn_{\text{patch}}^t(\node^q)_t}\right)=\log\left(\frac{\attn(\node^q)_{t}}{1-\attn(\node^q)_{t}}\right)+\Delta\attnLogits(\node^q)_{t}$;
note that the logodds function $p\mapsto\log\left(\frac{p}{1-p}\right)$ is the inverse
of the sigmoid function, so $\attn_{\text{patch}}^t(\node^q)=\operatorname{\sigma}\left(\log\left(\frac{\attn_{\text{patch}}^t(\node^q)_{t}}{1-\attn_{\text{patch}}^t(\node^q)_{t}}\right)\right)$.
Putting this together, we can compute all $\attnLogits_{\text{patch}}(\node^q)$
by combining all keys from the $\xnoise$ forward pass with all queries
from the $\xclean$ forward pass, and proceed to compute $\Delta\attnLogits(\node^q)$, and all $\Delta_{t}\attn(\node^q)_{t}$,
and thus all $\hat{\inter}_{\text{{AtPfix}}}^{K}(n_{t};\xclean,\xnoise)$, using $\operatorname{O}(T^{2})$
flops per attention head.

\Cref{alg:kfix} computes the contribution of some query node $\node^q$ and prompt pair $\xclean,\xnoise$ to the corrected AtP estimates $\hat{\contrib}_{\text{AtPfix}}^{K}(\node^k_{t})$ for key nodes $\node^k_1,\dots,\node^k_\numTokens$ from a single attention head, using $O(\numTokens)$ flops, while avoiding numerical overflows. We reuse the notation $\attn(\node^q)$, $\attn_{\text{patch}}^t(\node^q)$, $\Delta_t \attn(\node^q)$, $\attnLogits(\node^q)$, $\attnLogits_{\text{patch}}(\node^q)$, and $s_t$ from \Cref{par:kfix}, leaving the prompt pair implicit.

\begin{algorithm}
\caption{AtP correction for attention keys}
% \noindent\begin{minipage}[H]{\linewidth}
\label{alg:kfix}
\begin{algorithmic}[1]
\Require{$\mathbf{a}:=\attnLogits(\node^q)$, $\mathbf{a}^{\text{patch}}:=\attnLogits_{\text{patch}}(\node^q)$, $\mathbf{g}:=\frac{\partial\metric(\model(\xclean))}{\partial\attn(\node^q)}$}
\State $t^*\gets \operatorname{argmax}_t(a_t)$
\State $\ell \gets \mathbf{a} - a_{t^*} - \log\left(\sum_t e^{a_t - a_{t^*}}\right)$ \Comment{Clean log attn weights, $\ell=\log(\attn(\node^q))$}
\State $\mathbf{d} \gets \ell - \log(1-e^\ell)$ \Comment{Clean logodds, $d_t=\log\left(\frac{\attn(\node^q)_t}{1-\attn(\node^q)_t}\right)$}
\State $d_{t^*}\gets a_{t^*} - \max_{t\not=t^*} a_{t} - \log\left(\sum_{t'\not=t^*}e^{a_{t'}-\max_{t\not=t^*} a_t}\right)$ \Comment{Adjust $\mathbf{d}$; more stable for $a_{t^*}\gg\max_{t\not=t^*} a_t$}
\State $\ell^{\text{patch}} \gets \operatorname{logsigmoid}(\mathbf{d} + \mathbf{a}^{\text{patch}} - \mathbf{a})$ \Comment{Patched log attn weights, $\ell^{\text{patch}}_t=\log(\attn_{\text{patch}}^t(\node^q)_t)$}
\State $\Delta \ell \gets \ell^{\text{patch}} - \ell$ \Comment{$\Delta \ell_t = \log\left(\frac{\attn_{\text{patch}}^t(\node^q)_t}{\attn(\node^q)_t}\right)$}
\State $b\gets \operatorname{softmax}(\mathbf{a})^\intercal\mathbf{g}$ \Comment{$b=\attn(\node^q)^\intercal \mathbf{g}$}

\For{$t \gets 1 \textrm{ to } \numTokens$}
    \Statex \LeftComment{1}{Compute scaling factor $s_t:=\frac{\Delta_t\attn(\node^q)_t}{1-\attn(\node^q)_t}$}
    \If{$\ell^{\text{patch}}_t > \ell_t$}\Comment{Avoid overflow when $\ell^{\text{patch}}_t \gg \ell_t$}
        \State $s_t\gets e^{d_t + \Delta \ell_t + \log(1-e^{-\Delta \ell_t})}$ \Comment{$s_t=\frac{\attn(\node^q)_t}{1-\attn(\node^q)_t}\frac{\attn_{\text{patch}}^t(\node^q)_t}{\attn(\node^q)_t}\left(1-\frac{\attn(\node^q)_t}{\attn_{\text{patch}}^t(\node^q)_t}\right)$}
    \Else \Comment{Avoid overflow when $\ell^{\text{patch}}_t \ll \ell_t$}
        \State $s_t\gets -e^{d_t+ \log(1-e^{\Delta \ell_t})}$ \Comment{$s_t=-\frac{\attn(\node^q)_t}{1-\attn(\node^q)_t}\left(1-\frac{\attn_{\text{patch}}^t(\node^q)_t}{\attn(\node^q)_t}\right)$}
    \EndIf
    \State $r_t \gets s_t (g_t - b)$ \Comment{$r_t=s_t(\mathrm{onehot}(t)-\attn(\node^q))^\intercal \mathbf{g}=\Delta_t \attn(\node^q)\cdot \frac{\partial\metric(\model(\xclean))}{\partial\attn(\node^q)}$}
\EndFor
\State \Return{$\mathbf{r}$}
\end{algorithmic}
\end{algorithm}

The corrected AtP estimates $\hat{\contrib}_{\text{AtPfix}}^{K}(\node^k_{t})$ can then be computed using \Cref{eq:kfix}; in other words, by summing the returned $r_t$ from \Cref{alg:kfix} over queries $n^q$ for this attention head, and averaging over $\xclean,\xnoise\sim \promptPairDist$.

\subsubsection{Properties of GradDrop}\label{app:graddrop}

In~\Cref{sec:cancellation} we introduced GradDrop to address an AtP failure mode arising from cancellation between direct and indirect effects: roughly, if the total effect (on some prompt pair) is $\inter(\node)=\inter^{\text{direct}}(\node)+\inter^{\text{indirect}}(\node)$, and these are close to cancelling, then a small multiplicative approximation error in $\hat\inter_{\text{AtP}}^{\text{indirect}}(\node)$, due to nonlinearities, can accidentally cause $|\hat\inter_{\text{AtP}}^{\text{direct}}(\node)+\hat\inter_{\text{AtP}}^{\text{indirect}}(\node)|$ to be orders of magnitude smaller than $|\inter(\node)|$.

To address this failure mode with an improved estimator $\hat\contrib_{\text{AtP+GD}}(\node)$, there's 3 desiderata for GradDrop: \begin{enumerate}
\item $\hat\contrib_{\text{AtP+GD}}(\node)$ shouldn't be much smaller than $\hat\contrib_{\text{AtP}}(\node)$, because that would risk creating more false negatives.
\item $\hat\contrib_{\text{AtP+GD}}(\node)$ should usually not be much larger than $\hat\contrib_{\text{AtP}}(\node)$, because that would create false positives, which also slows down verification and can effectively create false negatives at a given budget.
\item If $\hat\contrib_{\text{AtP}}(\node)$ is suffering from the cancellation failure mode, then $\hat\contrib_{\text{AtP+GD}}(\node)$ should be significantly larger than $\hat\contrib_{\text{AtP}}(\node)$.
\end{enumerate}

Let's recall how GradDrop was defined in~\Cref{sec:cancellation}, using a virtual node $\node_\ell^{\text{out}}$ to represent the residual-stream contributions of layer $\ell$:

\begin{align*}
    \hat\contrib_{\text{AtP+GD}}(\node) :={}& \expect{\xclean,\xnoise}{\frac{1}{L-1} \sum_{\ell=1}^L\left|\hat\inter_{\text{AtP+GD}_\ell}(\node;\xclean,\xnoise)\right|}\\
    ={}&\expect{\xclean,\xnoise}{\frac{1}{L-1} \sum_{\ell=1}^L\left|(\node(\xnoise)-\node(\xclean))^\intercal\frac{\partial\metric^{\ell}}{\partial \node}\right|}\\
    ={}&\expect{\xclean,\xnoise}{\frac{1}{L-1} \sum_{\ell=1}^L\left|(\node(\xnoise)-\node(\xclean))^\intercal\frac{\partial\metric}{\partial \node}(\model(\xclean\mid\doGets{\node^{\text{out}}_\ell}{\node^{\text{out}}_\ell(\xclean)}))\right|}\\
\end{align*}

To better understand the behaviour of GradDrop, let's look more carefully at the gradient $\frac{\partial \metric}{\partial \node}$.
The total gradient $\frac{\partial \metric}{\partial \node}$ can be expressed as a sum of all path gradients from the node $\node$ to the output. Each path is characterized by the set of layers $s$ it goes through (in contrast to routing via the skip connection). We write the gradient along a path $s$ as $\frac{\partial \metric_s}{\partial n}$.

Let $\mathcal S$ be the set of all subsets of layers after the layer $\node$ is in. For example, the direct-effect path is given by $\emptyset \in \mathcal S$. Then the total gradient can be expressed as

\begin{align}
    \frac{\partial \metric}{\partial \node} &= \sum_{s\in\mathcal S} \frac{\partial \metric_s}{\partial \node}.
\end{align}

We can analogously define $\hat\inter_{\text{AtP}}^s(\node)=(\node(\xnoise)-\node(\xclean))^\intercal \frac{\partial\metric_s}{\partial\node}$, and break down $\hat\inter_{\text{AtP}}(\node)=\sum_{s\in\mathcal S}\hat\inter_{\text{AtP}}^s(\node)$. The effect of doing GradDrop at some layer $\ell$ is then to drop all terms $\hat\inter_{\text{AtP}}^s(\node)$ with $\ell\in s$: in other words, \begin{align}\hat\inter_{\text{AtP+GD}_\ell}(\node)&=\sum_{\substack{s\in\mathcal S\\\ell\not\in s}} \hat\inter_{\text{AtP}}^s(\node).\end{align}

Now we'll use this understanding to discuss the 3 desiderata.

Firstly, most node effects are approximately independent of most layers (see e.g. \citet{veit2016residual}); for any layer $\ell$ that $\node$'s effect is independent of, we'll have $\hat\inter_{\text{AtP+GD}_\ell}(\node)=\hat\inter_{\text{AtP}}(\node)$. Letting $K$ be the set of downstream layers that matter, this guarantees $\frac{1}{L-1}\sum_{\ell=1}^L\left|\hat\inter_{\text{AtP+GD}_\ell}(\node;\xclean,\xnoise)\right|\geq \frac{L-|K|-1}{L-1}\left|\hat\inter_{\text{AtP}}(\node;\xclean,\xnoise)\right|$, which meets the first desideratum.

Regarding the second desideratum: for each $\ell$ we have $\left|\hat \inter_{\text{AtP+GD}_\ell} (\node)\right|\leq \sum_{s\in\mathcal{S}}\left|\hat\inter_{\text{AtP}}^s(\node)\right|$, so overall we have $\frac{1}{L-1}\sum_{\ell=1}^L\left|\hat\inter_{\text{AtP+GD}_\ell}(\node)\right|\leq \frac{L-|K|-1}{L-1}\left|\hat\inter_{\text{AtP}}(\node)\right|+\frac{|K|}{L-1}\sum_{s\in\mathcal{S}}\left|\hat\inter_{\text{AtP}}^s(\node)\right|$. For the RHS to be much larger (e.g. $\alpha$ times larger) than $\left|\sum_{s\in\mathcal{S}}\hat\inter_{\text{AtP}}^s(\node)\right|=|\hat\inter_{\text{AtP}}(\node)|$, there must be quite a lot of cancellation between different paths, enough so that $\sum_{s\in\mathcal{S}}\left|\hat\inter_{\text{AtP}}^s(\node)\right|\geq \frac{(L-1)\alpha}{|K|} \left|\sum_{s\in\mathcal{S}}\hat\inter_{\text{AtP}}^s(\node)\right|$. This is possible, but seems generally unlikely for e.g. $\alpha>3$.

Now let's consider the third desideratum, i.e. suppose $\node$ is a cancellation false negative, with $|\hat\inter_{\text{AtP}}(\node)|\ll |\inter(\node)| \ll |\inter^{\text{direct}}(\node)|\approx |\hat\inter_{\text{AtP}}^{\text{direct}}(\node)|$.
Then, $\left|\sum_{s\in\mathcal S\setminus\emptyset}\hat\inter_{\text{AtP}}^{s}(\node)\right|=\left|\hat\inter_{\text{AtP}}(\node)-\hat\inter_{\text{AtP}}^{\text{direct}}(\node)\right|\gg|\inter(\node)|$.
The summands in $\sum_{s\in\mathcal S\setminus\emptyset}\hat\inter_{\text{AtP}}^{s}(\node)$ are the union of the summands in $\sum_{\substack{s\in\mathcal S\\\ell\in s}} \hat\inter_{\text{AtP}}^{s}(\node)=\hat\inter_{\text{AtP}}(\node)-\hat\inter_{\text{AtP+GD}_\ell}(\node)$ across layers $\ell$.

\begin{sloppypar}
It's then possible but intuitively unlikely that $\sum_{\ell}\left|\hat\inter_{\text{AtP}}(\node)-\hat\inter_{\text{AtP+GD}_\ell}(\node)\right|$ would be much smaller than $\left|\hat\inter_{\text{AtP}}(\node)-\hat\inter_{\text{AtP}}^{\text{direct}}(\node)\right|$. Suppose the ratio is $\alpha$, i.e. suppose $\sum_{\ell}\left|\hat\inter_{\text{AtP}}(\node)-\hat\inter_{\text{AtP+GD}_\ell}(\node)\right|=\alpha\left|\hat\inter_{\text{AtP}}(\node)-\hat\inter_{\text{AtP}}^{\text{direct}}(\node)\right|$. For example, if all indirect effects use paths of length 1 then the union is a disjoint union, so $\sum_{\ell}\left|\hat\inter_{\text{AtP}}(\node)-\hat\inter_{\text{AtP+GD}_\ell}(\node)\right|\geq\left|\sum_{\ell}\left(\hat\inter_{\text{AtP}}(\node)-\hat\inter_{\text{AtP+GD}_\ell}(\node)\right)\right|=\left|\hat\inter_{\text{AtP}}(\node)-\hat\inter_{\text{AtP}}^{\text{direct}}(\node)\right|$, so $\alpha\geq 1$. Now:
\end{sloppypar}

\begin{align}
    \sum_{\ell\in K}\left|\hat\inter_{\text{AtP+GD}_\ell}(\node)\right|
    &\geq \sum_{\ell\in K}\left|\hat\inter_{\text{AtP}}(\node)-\hat\inter_{\text{AtP+GD}_\ell}(\node)\right|-|K|\left|\hat\inter_{\text{AtP}}(\node)\right|\\
    &=\alpha\left|\hat\inter_{\text{AtP}}(\node)-\hat\inter_{\text{AtP}}^{\text{direct}}(\node)\right|-|K|\left|\hat\inter_{\text{AtP}}(\node)\right|\\
    &\geq \alpha\left|\hat\inter_{\text{AtP}}^{\text{direct}}(\node)\right|-(|K|+\alpha)\left|\hat\inter_{\text{AtP}}(\node)\right|\\
    \therefore\frac{1}{L-1}\sum_{\ell=1}^L \left|\hat\inter_{\text{AtP+GD}_\ell}(\node)\right|
    &= \frac{1}{L-1}\sum_{\ell\in K}\left|\hat\inter_{\text{AtP+GD}_\ell}(\node)\right| + \frac{L-|K|-1}{L-1}\left|\hat\inter_{\text{AtP}}(\node)\right|\\
    &\geq \frac{\alpha}{L-1}\left|\hat\inter_{\text{AtP}}^{\text{direct}}(\node)\right| + \frac{L-2|K|-1-\alpha}{L-1}\left|\hat\inter_{\text{AtP}}(\node)\right|\\
\end{align}

And the RHS is an improvement over $\left|\hat\inter_{\text{AtP}}(\node)\right|$ so long as $\alpha\left|\hat\inter_{\text{AtP}}^{\text{direct}}(\node)\right|>(2|K|+\alpha)\left|\hat\inter_{\text{AtP}}(\node)\right|$, which is likely given the assumptions.

Ultimately, though, the desiderata are validated by the experiments, which consistently show GradDrops either decreasing or leaving untouched the number of false negatives, and thus improving performance apart from the initial upfront cost of the extra backwards passes.

\subsection{Algorithm for computing diagnostics}\label{app:diagnostic_algorithm}

Given summary statistics $\bar i_\pm$, $s_\pm$ and $\rcount_\pm$ for every node $\node$, obtained from~\Cref{alg:subsampling}, and a threshold $\theta > 0$ we can use Welch's $t$-test~\cite{welch1947test} to test the hypothesis that $|\bar i_+ - \bar i_-| \geq \theta$. Concretely we compute the $t$-statistic via

\begin{align}
    s_{\bar i_\pm} &= \frac{s_\pm}{\sqrt{\rcount_\pm}}\\
    t &= \frac{\theta - |\bar i_+ - \bar i_-|}{\sqrt{s_{\bar i_+}^2 + s_{\bar i_-}^2}}.
\end{align}

\noindent The effective degrees of freedom $\nu$ can be approximated with the Welch--Satterthwaite equation
\begin{align}
    \nu_{\text{Welch}} = \frac{\left(\frac{s_+^2}{\rcount_+} + \frac{s_-^2}{\rcount_-}\right)^2}{\frac{s_+^4}{\rcount^2_+(\rcount_+ - 1)} + \frac{s_-^4}{\rcount^2_-(\rcount_- - 1)}}
\end{align}

We then compute the probability ($p$-value) of obtaining a $t$ at least as large as observed, using the cumulative distribution function of Student's $t\Big{(}x; \nu_{\text{Welch}}\Big{)}$ at the appropriate points. We take the max of the individual $p$-values of all nodes to obtain an aggregate upper bound. Finally, we use binary search to find the largest threshold $\theta$ that still has an aggregate $p$-value smaller than a given target $p$ value. We show multiple such diagnostic curves in~\Cref{app:detailed_results}, for different confidence levels ($1 - p_{\text{target}}$).

\section{Experiments}

\subsection{Prompt Distributions}\label{app:prompt_distributions}

\subsubsection{\texttt{IOI}}

We use the following prompt template:

\noindent\fbox{\strut{}\texttt{BOS}}\fbox{\strut{}When}\fbox{\strut{}\textvisiblespace{}[A]}\fbox{\strut{}\textvisiblespace{}and}\fbox{\strut{}\textvisiblespace{}[B]}\fbox{\strut{}\textvisiblespace{}went}\fbox{\strut{}\textvisiblespace{}to}\fbox{\strut{}\textvisiblespace{}the}\fbox{\strut{}\textvisiblespace{}bar}\fbox{\strut{},}\fbox{\strut{}\textvisiblespace{}[A/C]}\fbox{\strut{}\textvisiblespace{}gave}\fbox{\strut{}\textvisiblespace{}a}\fbox{\strut{}\textvisiblespace{}drink}\fbox{\strut{}\textvisiblespace{}to}\doublebox{\strut{}\textvisiblespace{}[B/A]}

Each clean prompt $\xclean$ uses two names A and B with completion B, while a noise prompt $\xnoise$ uses names A, B, and C with completion A. We construct all possible such assignments where names are chosen from the set of \{\texttt{Michael}, \texttt{Jessica}, \texttt{Ashley}, \texttt{Joshua}, \texttt{David}, \texttt{Sarah}\}, resulting in 120 prompt pairs.

\subsubsection{\texttt{A-AN}}

We use the following prompt template to induce the prediction of an indefinite article.

\noindent\fbox{\strut{}\texttt{BOS}}\fbox{\strut{}I}\fbox{\strut{}\textvisiblespace{}want}\fbox{\strut{}\textvisiblespace{}one}\fbox{\strut{}\textvisiblespace{}pear}\fbox{\strut{}.}\fbox{\strut{}\textvisiblespace{}Can}\fbox{\strut{}\textvisiblespace{}you}\fbox{\strut{}\textvisiblespace{}pick}\fbox{\strut{}\textvisiblespace{}up}\fbox{\strut{}\textvisiblespace{}a}\fbox{\strut{}\textvisiblespace{}pear}\fbox{\strut{}\textvisiblespace{}for}\fbox{\strut{}\textvisiblespace{}me}\fbox{\strut{}?}\\
\fbox{\strut{}\textvisiblespace{}I}\fbox{\strut{}\textvisiblespace{}want}\fbox{\strut{}\textvisiblespace{}one}\fbox{\strut{}\textvisiblespace{}orange}\fbox{\strut{}.}\fbox{\strut{}\textvisiblespace{}Can}\fbox{\strut{}\textvisiblespace{}you}\fbox{\strut{}\textvisiblespace{}pick}\fbox{\strut{}\textvisiblespace{}up}\fbox{\strut{}\textvisiblespace{}an}\fbox{\strut{}\textvisiblespace{}orange}\fbox{\strut{}\textvisiblespace{}for}\fbox{\strut{}\textvisiblespace{}me}\fbox{\strut{}?}\\
\fbox{\strut{}\textvisiblespace{}I}\fbox{\strut{}\textvisiblespace{}want}\fbox{\strut{}\textvisiblespace{}one}\fbox{\strut{}\textvisiblespace{}[OBJECT]}\fbox{\strut{}.}\fbox{\strut{}\textvisiblespace{}Can}\fbox{\strut{}\textvisiblespace{}you}\fbox{\strut{}\textvisiblespace{}pick}\fbox{\strut{}\textvisiblespace{}up}\doublebox{\strut\textvisiblespace{}[a/an]}

We found that zero shot performance of small models was relatively low, but performance improved drastically when providing a single example of each case. Model performance was sensitive to the ordering of the two examples but was better than random in all cases. The magnitude and sign of the impact of the few-shot ordering was inconsistent.

Clean prompts $\xclean$ contain objects inducing `\textvisiblespace{}\texttt{a}', one of \{\texttt{boat}, \texttt{coat}, \texttt{drum}, \texttt{horn}, \texttt{map}, \texttt{pipe}, \texttt{screw}, \texttt{stamp}, \texttt{tent}, \texttt{wall}\}.
Noise prompts $\xnoise$ contain objects inducing `\textvisiblespace{}\texttt{an}', one of \{\texttt{apple}, \texttt{ant}, \texttt{axe}, \texttt{award}, \texttt{elephant}, \texttt{egg}, \texttt{orange}, \texttt{oven}, \texttt{onion}, \texttt{umbrella}\}.
This results in a total of 100 prompt pairs.

\subsection{Cancellation across a distribution}\label{app:distribution_cancellation}

\begin{figure}[t]
    \centering
    \begin{subfigure}[b]{0.49\textwidth}
        \centering
        \includegraphics[width=\textwidth]{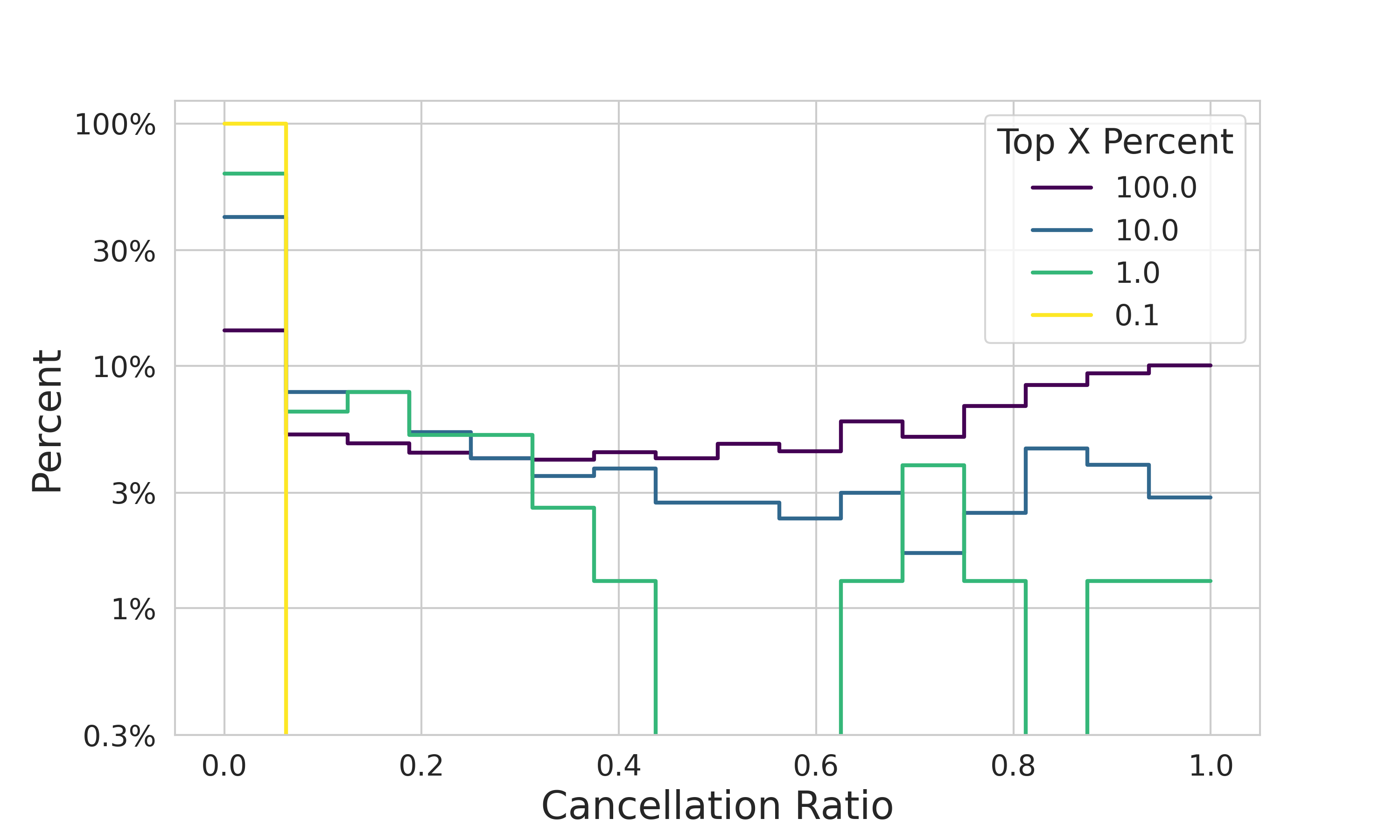}
        \caption{Pythia-410M}
    \end{subfigure}
    \hfill
    \begin{subfigure}[b]{0.49\textwidth}
        \centering
        \includegraphics[width=\textwidth]{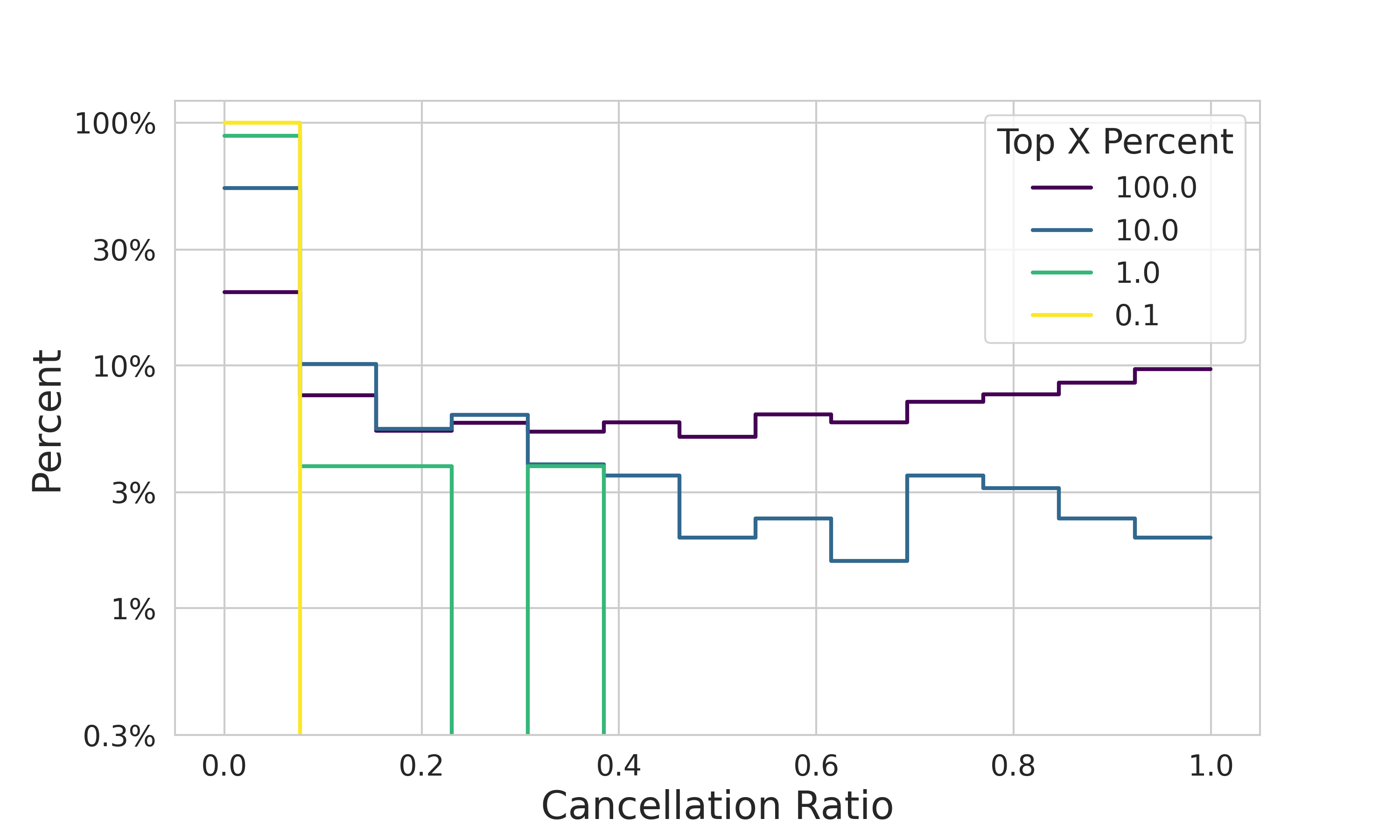}
        \caption{Pythia-1B}
    \end{subfigure}
    \hfill
    \begin{subfigure}[b]{0.49\textwidth}
        \centering
        \includegraphics[width=\textwidth]{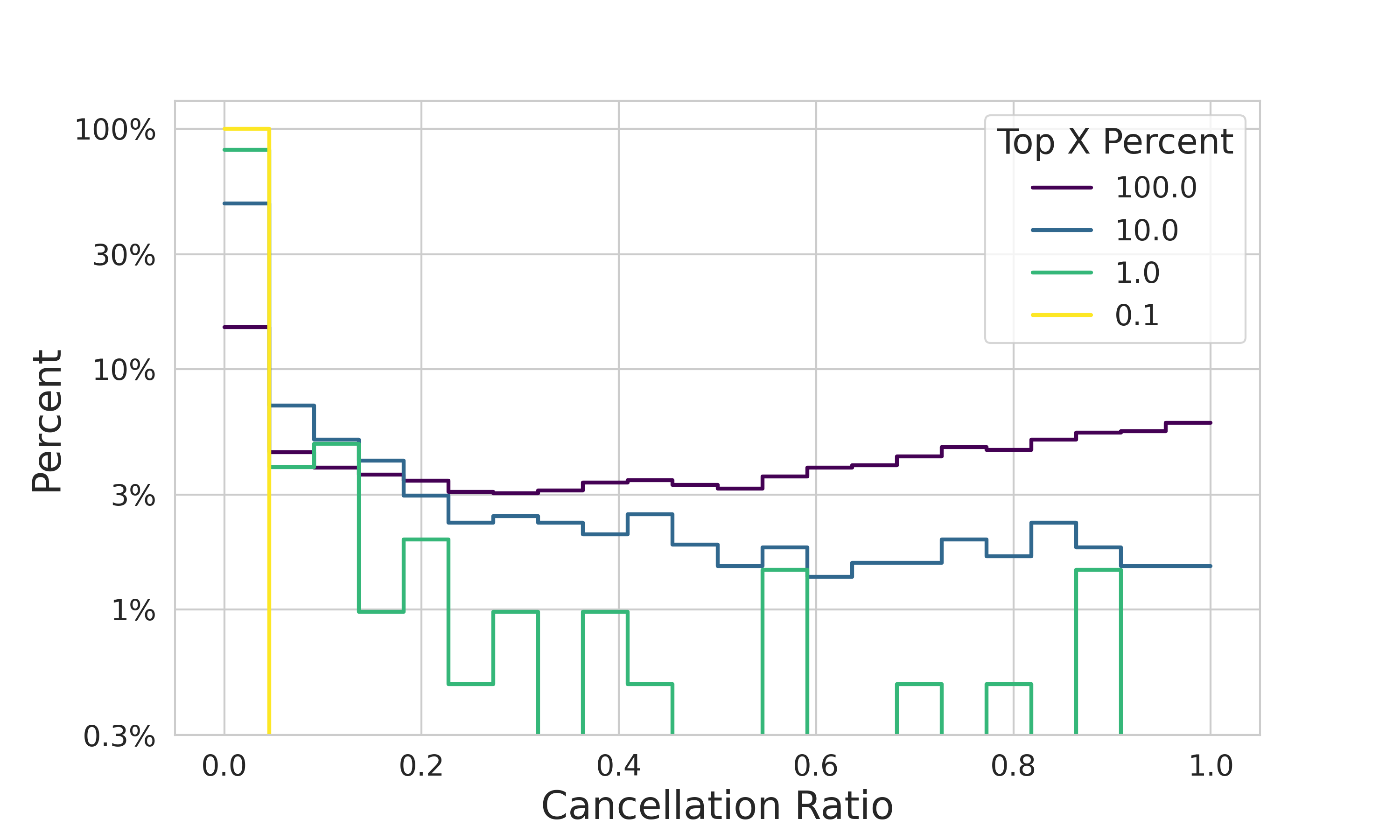}
        \caption{Pythia-2.8B}
    \end{subfigure}
    \hfill
    \begin{subfigure}[b]{0.49\textwidth}
        \centering
        \includegraphics[width=\textwidth]{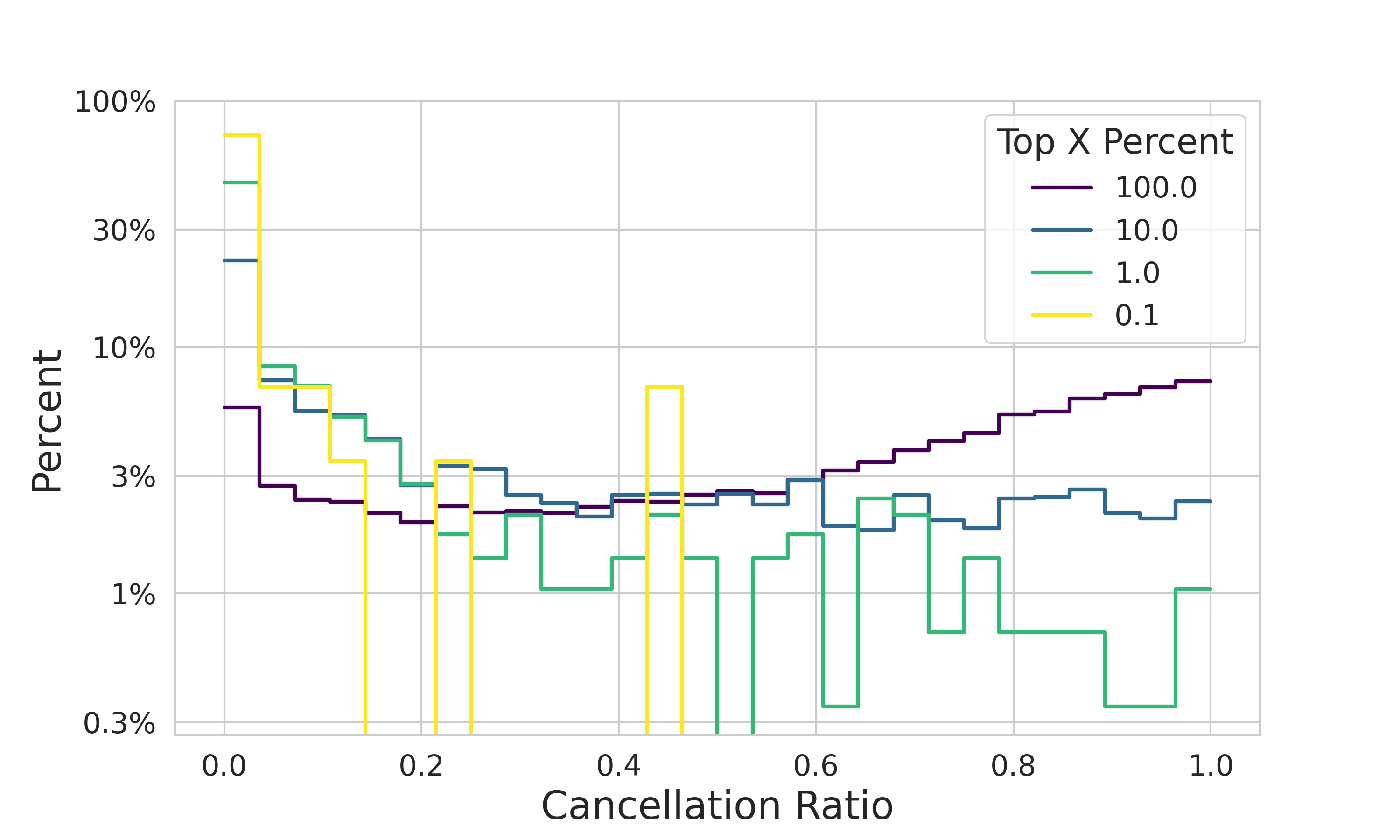}
        \caption{Pythia-12B}
    \end{subfigure}
    \caption{Cancellation ratio across \texttt{IOI} for various model sizes. A ratio of 1 means positive and negative effects cancel out across the distribution, whereas a ratio of 0 means only either negative or positive effects exist across the distribution. We report cancellation ratio for different percentiles of nodes based on $\sum_{\xclean, \xnoise} \left|\inter(\node; \xclean, \xnoise)\right|$.}
    \label{fig:distribution_cancellation}
\end{figure}

As mention in~\Cref{sec:background}, we average the magnitudes of effects across a distribution, rather than taking the magnitude of the average effect. We do this because cancellation of effects is happening frequently across a distribution, which, together with imprecise estimates, could lead to significant false negatives. A proper ablation study to quantify this effect exactly is beyond the scope of this work. In~\Cref{fig:distribution_cancellation}, we show the degree of cancellation across the IOI distribution for various model sizes. For this we define the \emph{Cancellation Ratio} of node $\node$ as
\begin{align*}
    1 - \frac{\left|\sum_{\xclean, \xnoise} \inter(\node; \xclean, \xnoise)\right|}{\sum_{\xclean, \xnoise} \left|\inter(\node; \xclean, \xnoise)\right|}.
\end{align*}

\subsection{Additional detailed results}
\label{app:detailed_results}

\begin{figure}[t]
    \centering
    \caption{Diagnostic of false negatives for 12B across distributions.}
    
    \begin{subfigure}[b]{\textwidth}
    \setcounter{subsubfigure}{0}
        \begin{subsubfigure}[b]{0.32\textwidth}
            \includegraphics[width=\textwidth]{assets/diagnostic_ioipp_attn.png}
            \caption{\texttt{IOI-PP}}
        \end{subsubfigure}
        \begin{subsubfigure}[b]{0.32\textwidth}
            \includegraphics[width=\textwidth]{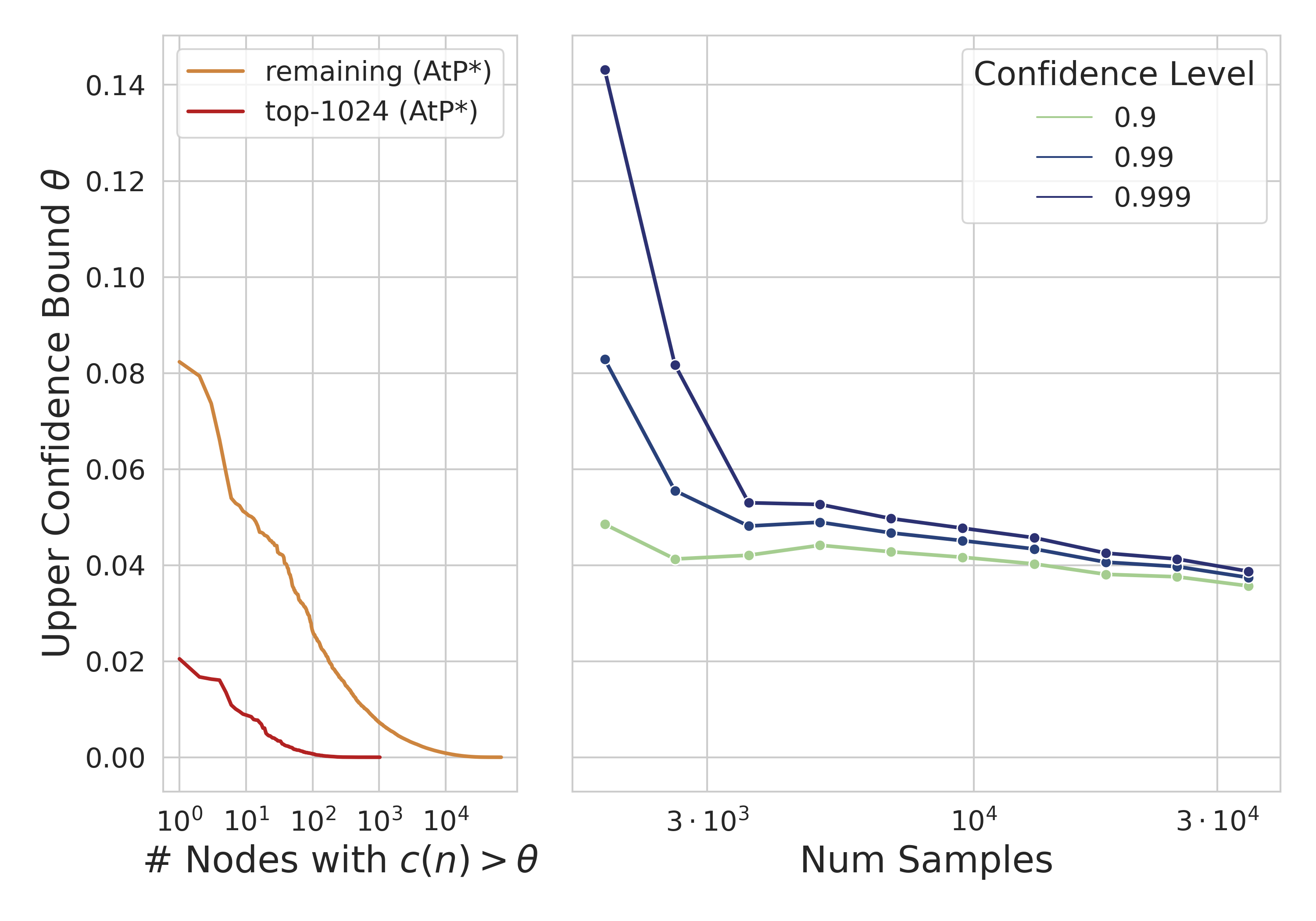}
            \caption{\texttt{RAND-PP}}
        \end{subsubfigure}
        \begin{subsubfigure}[b]{0.32\textwidth}
            \includegraphics[width=\textwidth]{assets/diagnostic_ioi_dist_attn.png}
            \caption{\texttt{IOI}}
        \end{subsubfigure}
        \caption{\AttentionNodes{}}
    \end{subfigure}

    \begin{subfigure}[b]{\textwidth}
    \setcounter{subsubfigure}{0}
        \begin{subsubfigure}[b]{0.32\textwidth}
            \includegraphics[width=\textwidth]{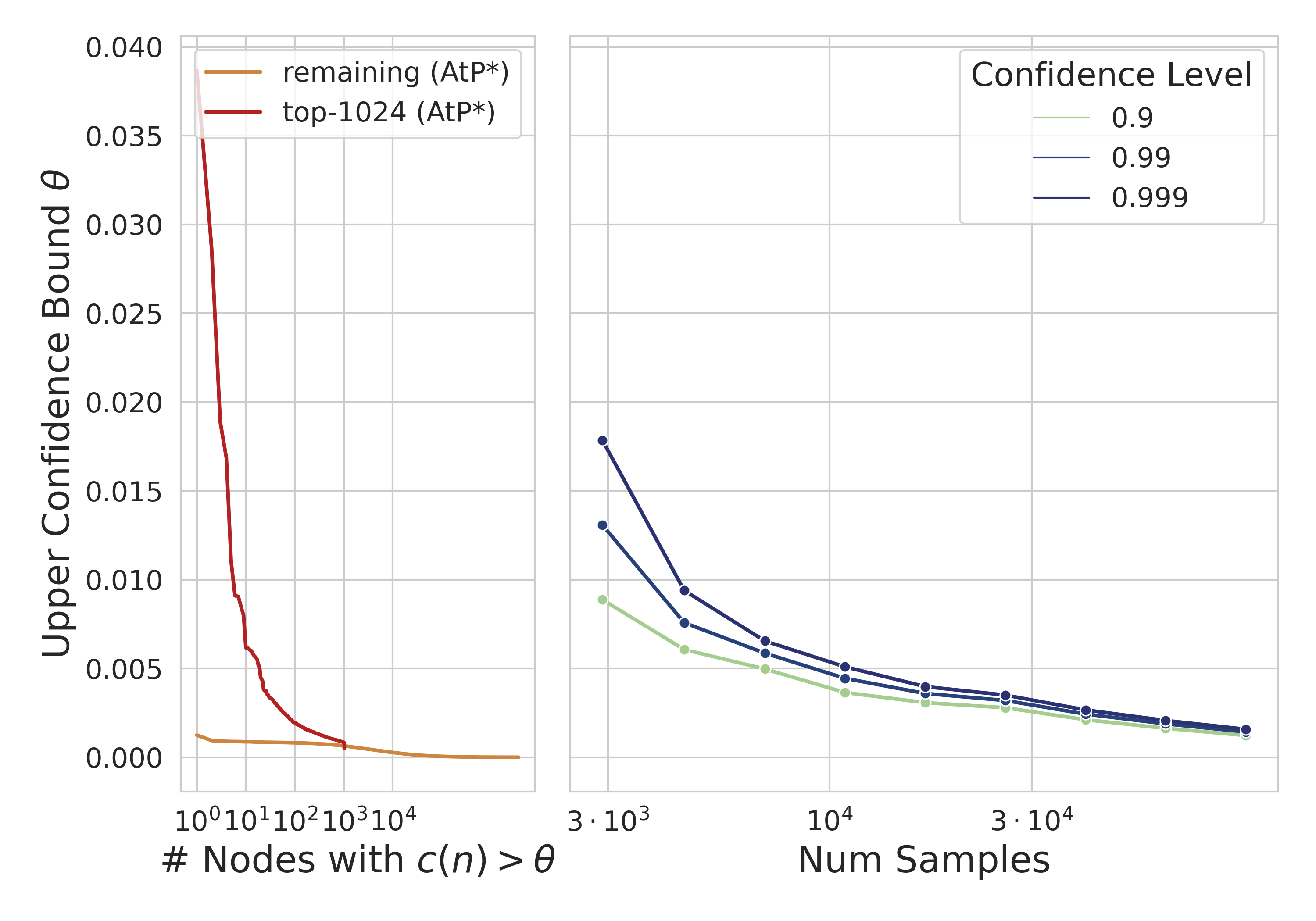}
            \caption{\texttt{CITY-PP}}
        \end{subsubfigure}
        \begin{subsubfigure}[b]{0.32\textwidth}
            \includegraphics[width=\textwidth]{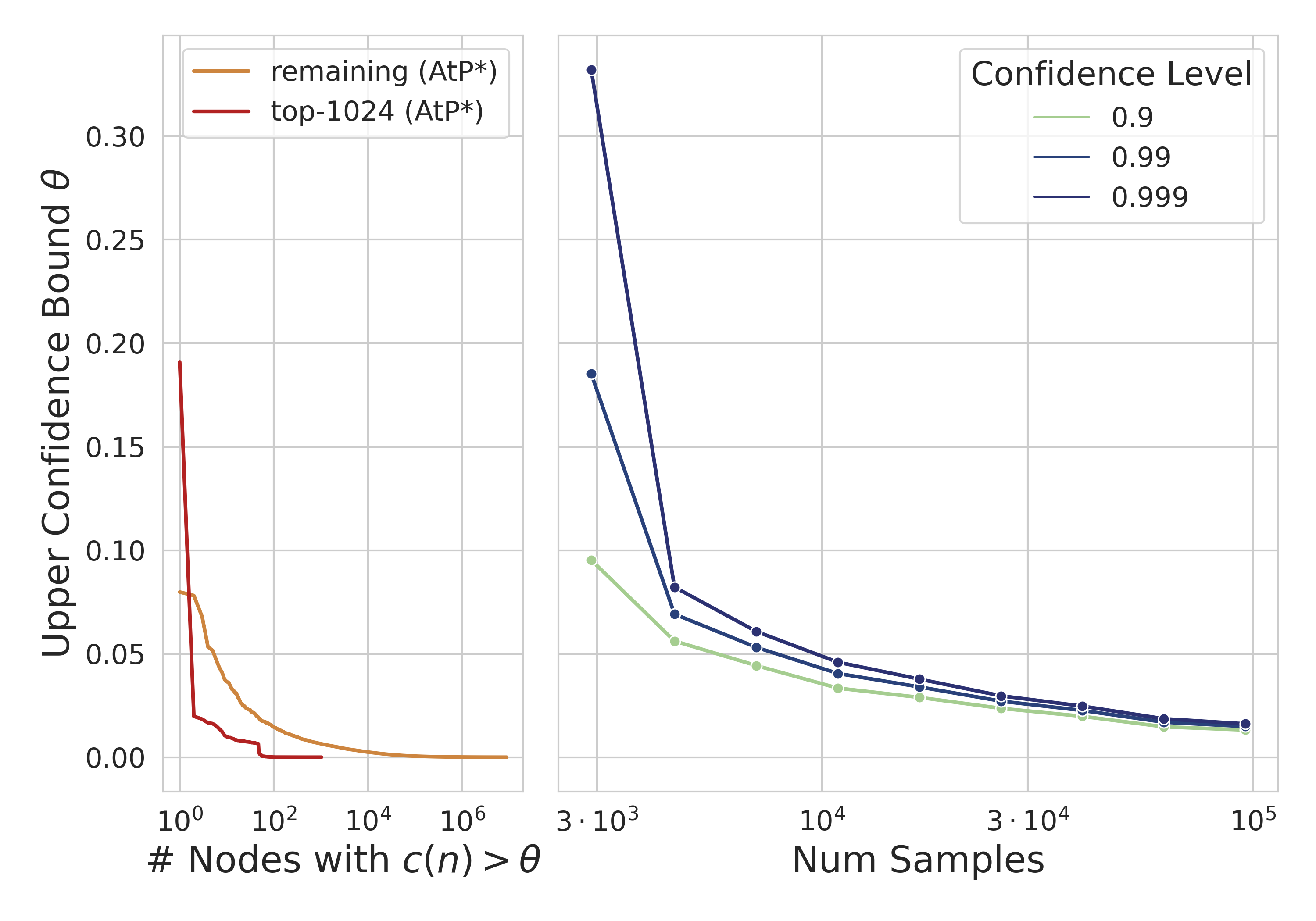}
            \caption{\texttt{RAND-PP}}
        \end{subsubfigure}
        \begin{subsubfigure}[b]{0.32\textwidth}
            \includegraphics[width=\textwidth]{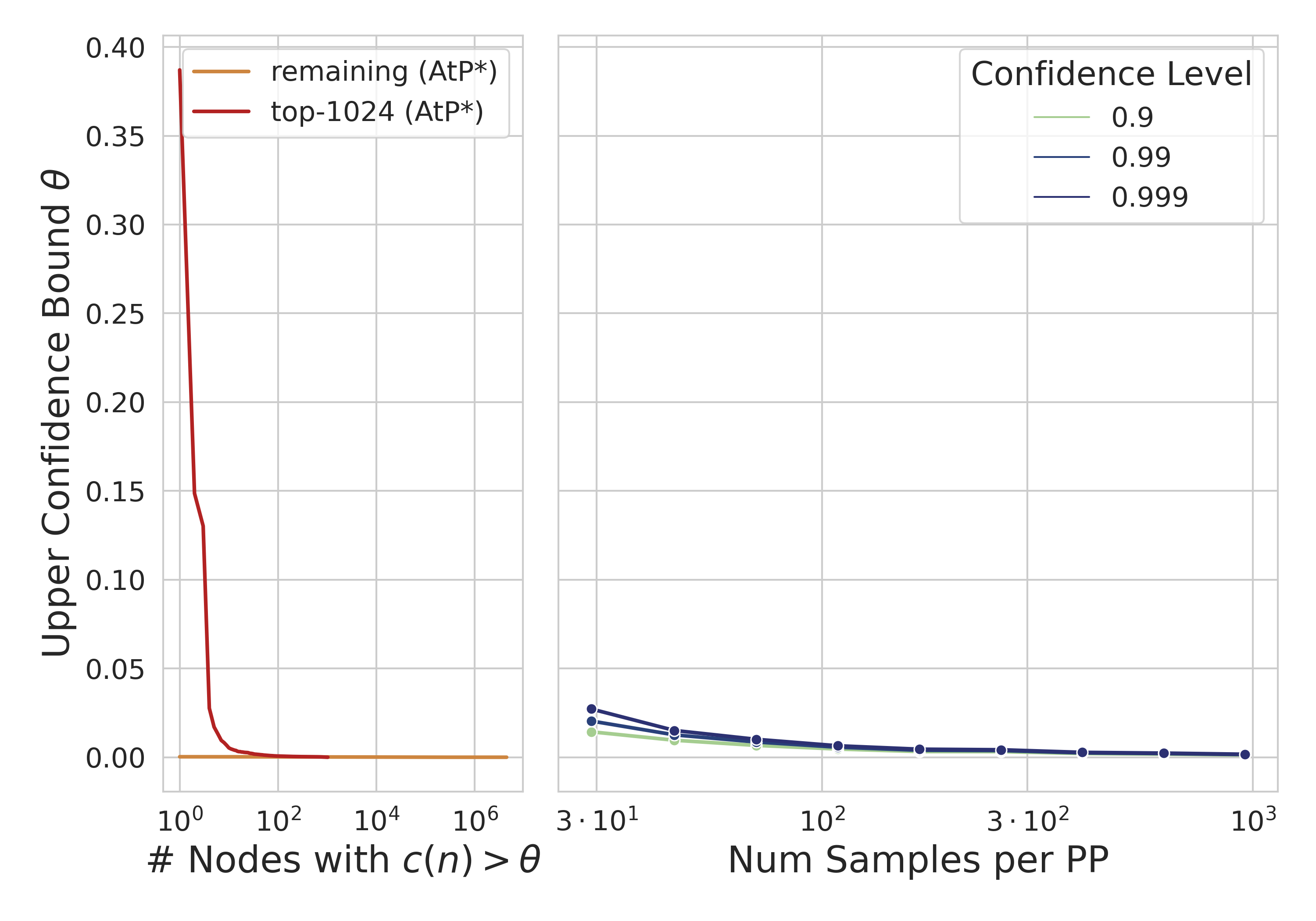}
            \caption{\texttt{A-AN}}
        \end{subsubfigure}
        \caption{\NeuronNodes{}}
    \end{subfigure}
    \label{fig:diagnostics_all_distributions}
\end{figure}

\begin{figure}
    \centering
    \caption{Cost of verified 100\% recall curves, sweeping across models and settings for \NeuronNodes{}}
    \begin{subfigure}[b]{\textwidth}
    \setcounter{subsubfigure}{0}
        \begin{subsubfigure}[b]{0.24\textwidth}
            \includegraphics[width=\textwidth]{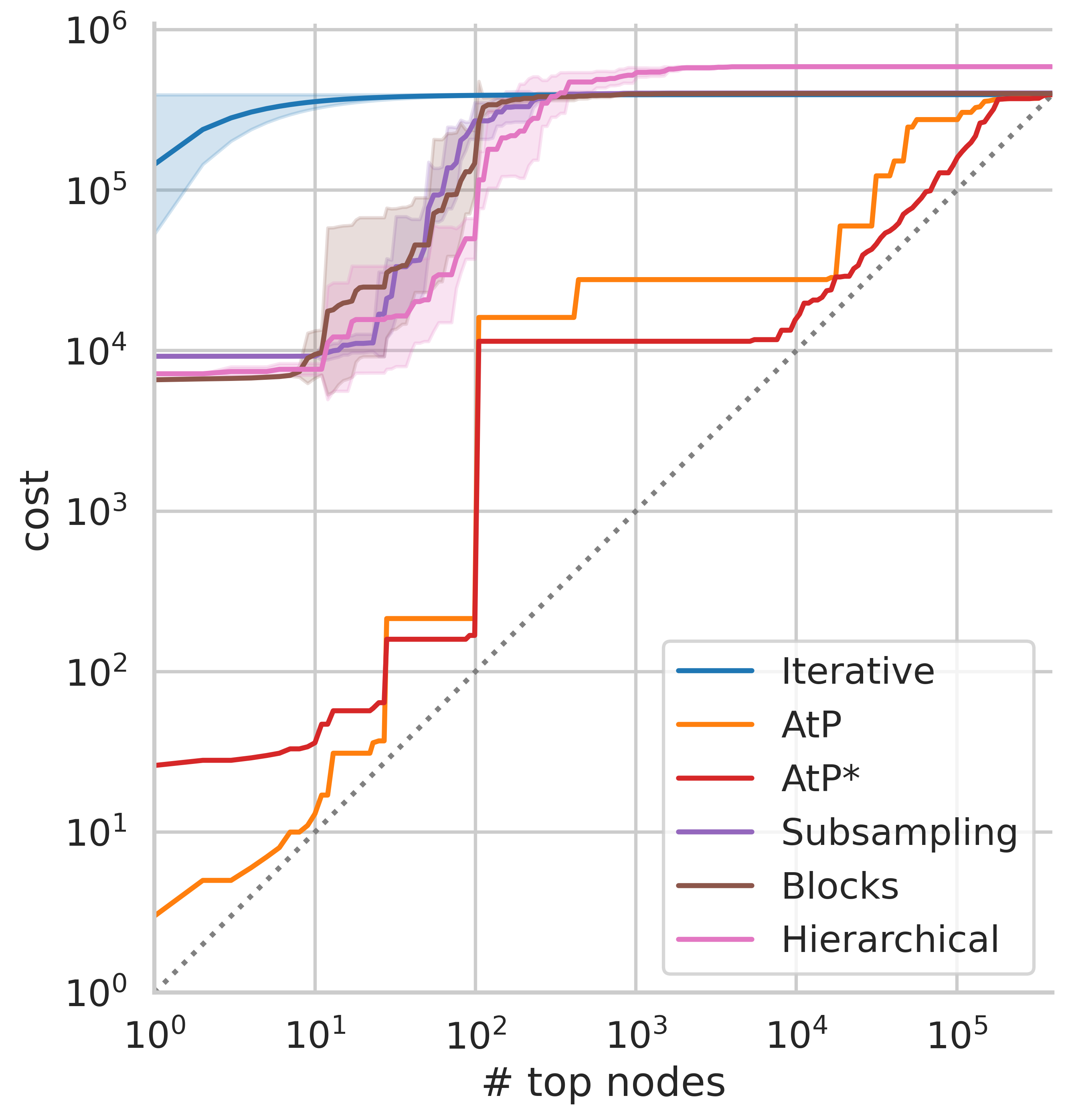}
            \caption{Pythia 410M}
        \end{subsubfigure}
        \begin{subsubfigure}[b]{0.24\textwidth}
            \includegraphics[width=\textwidth]{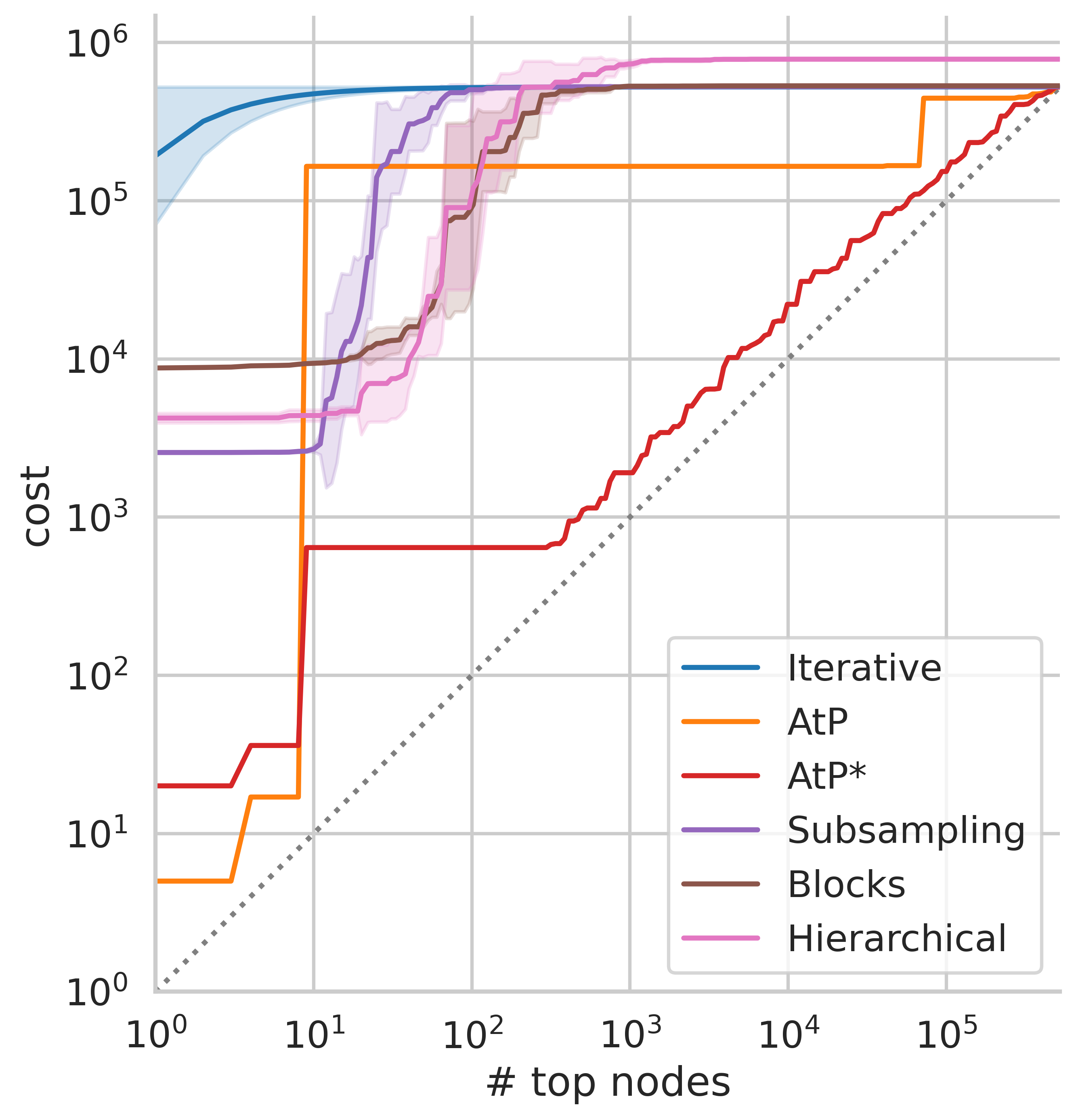}
            \caption{Pythia 1B}
        \end{subsubfigure}
        \begin{subsubfigure}[b]{0.24\textwidth}
            \includegraphics[width=\textwidth]{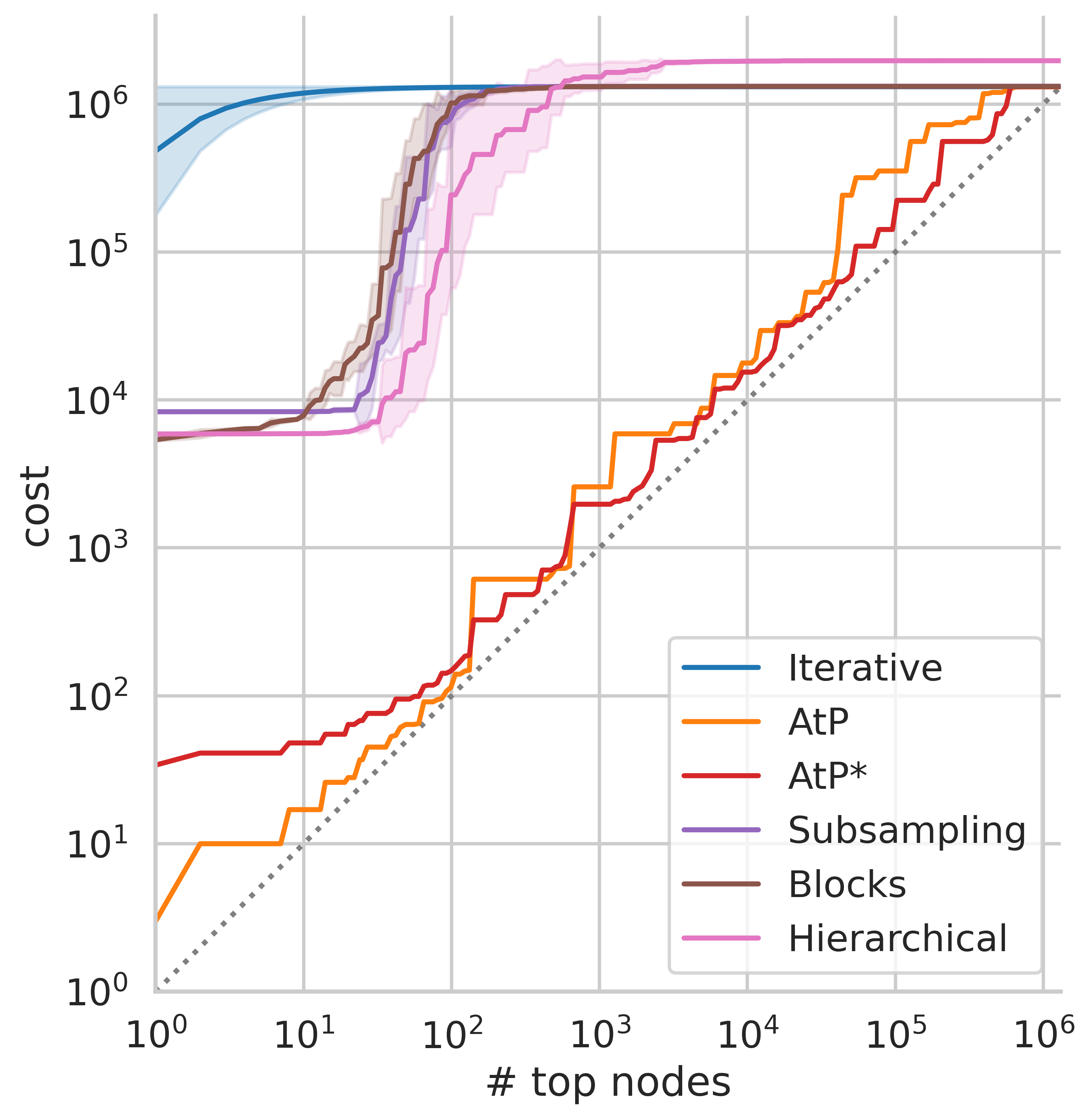}
            \caption{Pythia 2.8B}
        \end{subsubfigure}
        \begin{subsubfigure}[b]{0.24\textwidth}
            \includegraphics[width=\textwidth]{bb_cost_12b.png}
            \caption{Pythia 12B}
        \end{subsubfigure}
        \caption{\texttt{CITY-PP}}
    \end{subfigure}
    \begin{subfigure}[b]{\textwidth}
    \setcounter{subsubfigure}{0}
        \begin{subsubfigure}[b]{0.24\textwidth}
            \includegraphics[width=\textwidth]{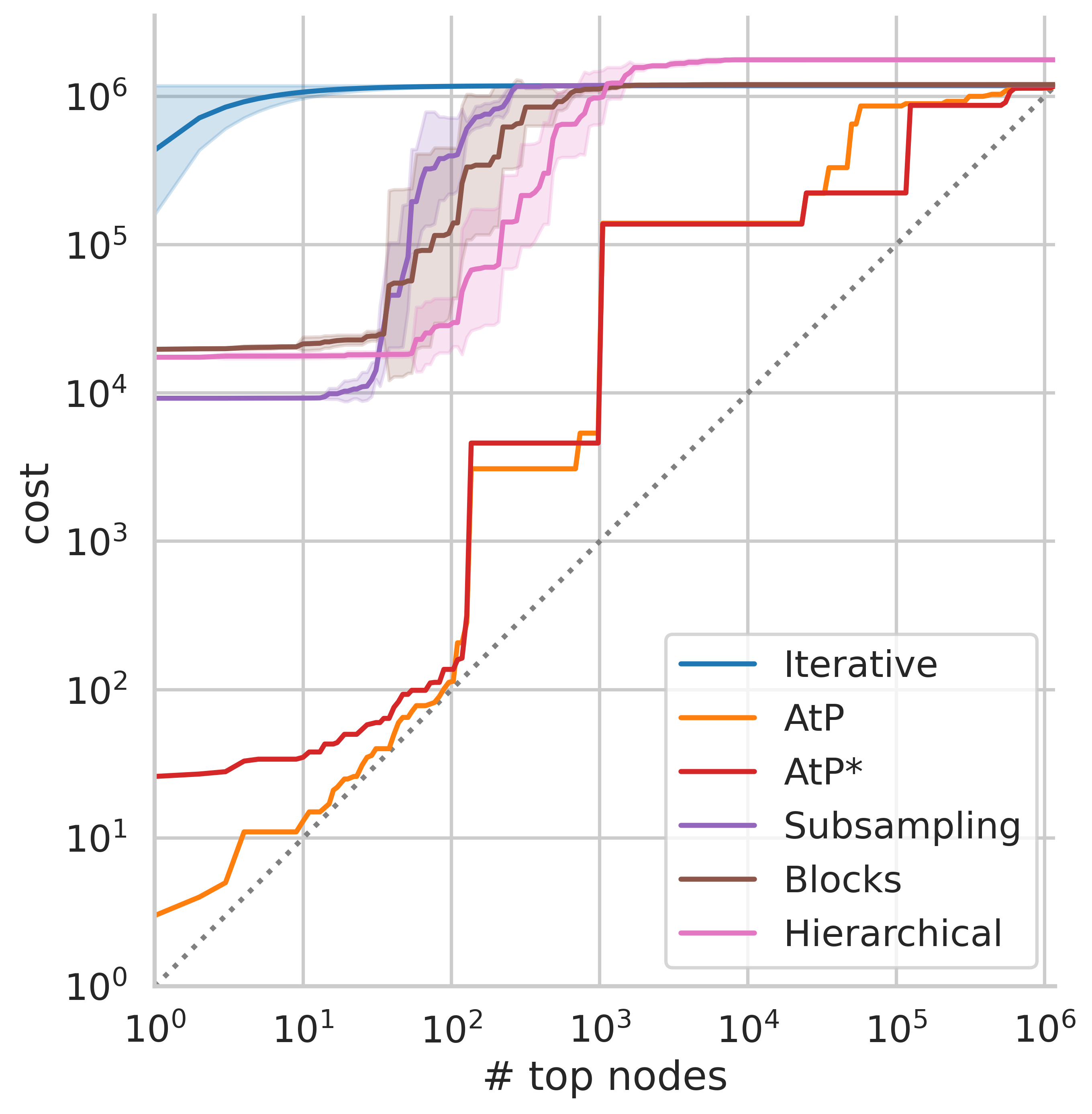}
            \caption{Pythia 410M}
        \end{subsubfigure}
        \begin{subsubfigure}[b]{0.24\textwidth}
            \includegraphics[width=\textwidth]{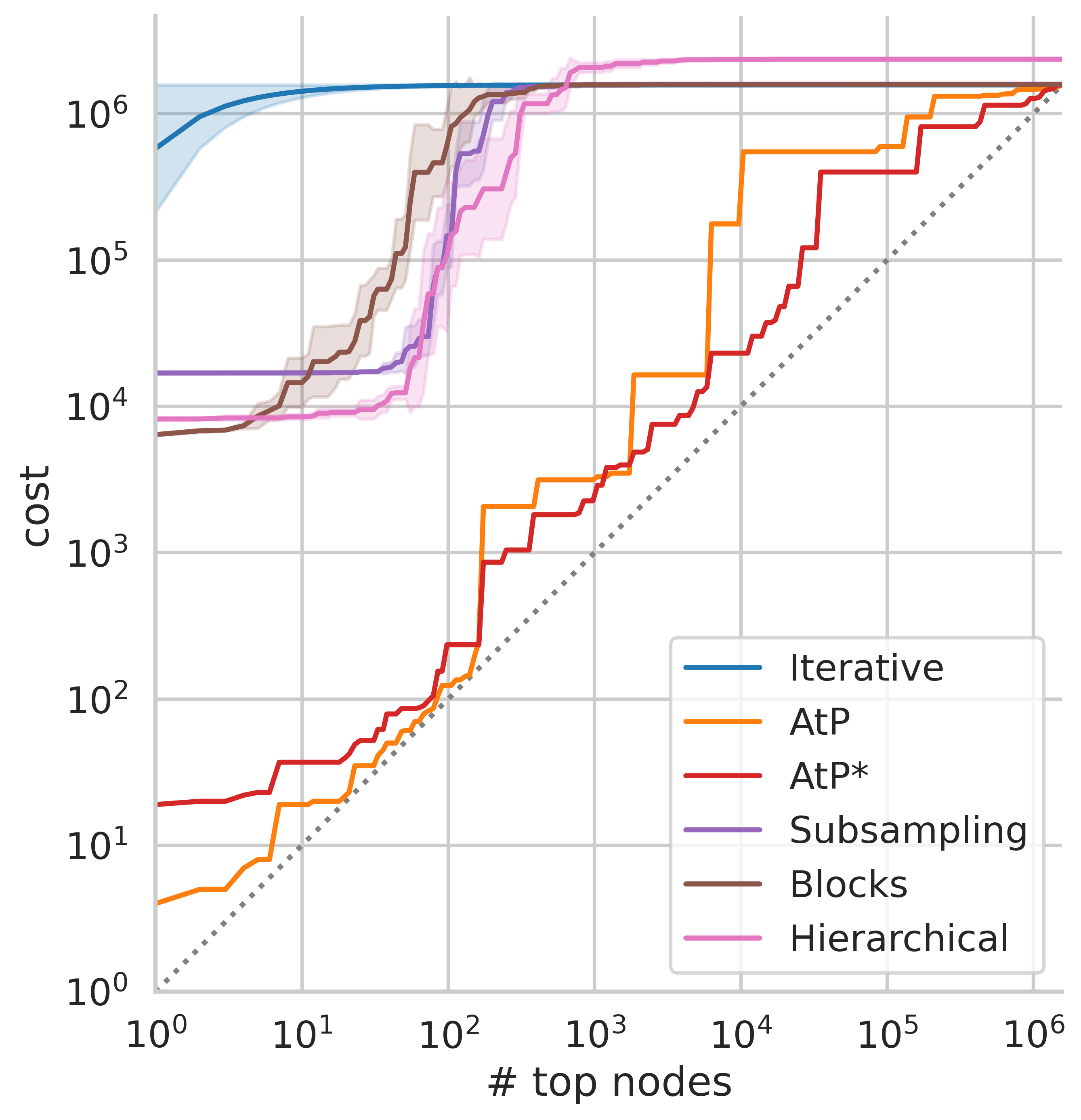}
            \caption{Pythia 1B}
        \end{subsubfigure}
        \begin{subsubfigure}[b]{0.24\textwidth}
            \includegraphics[width=\textwidth]{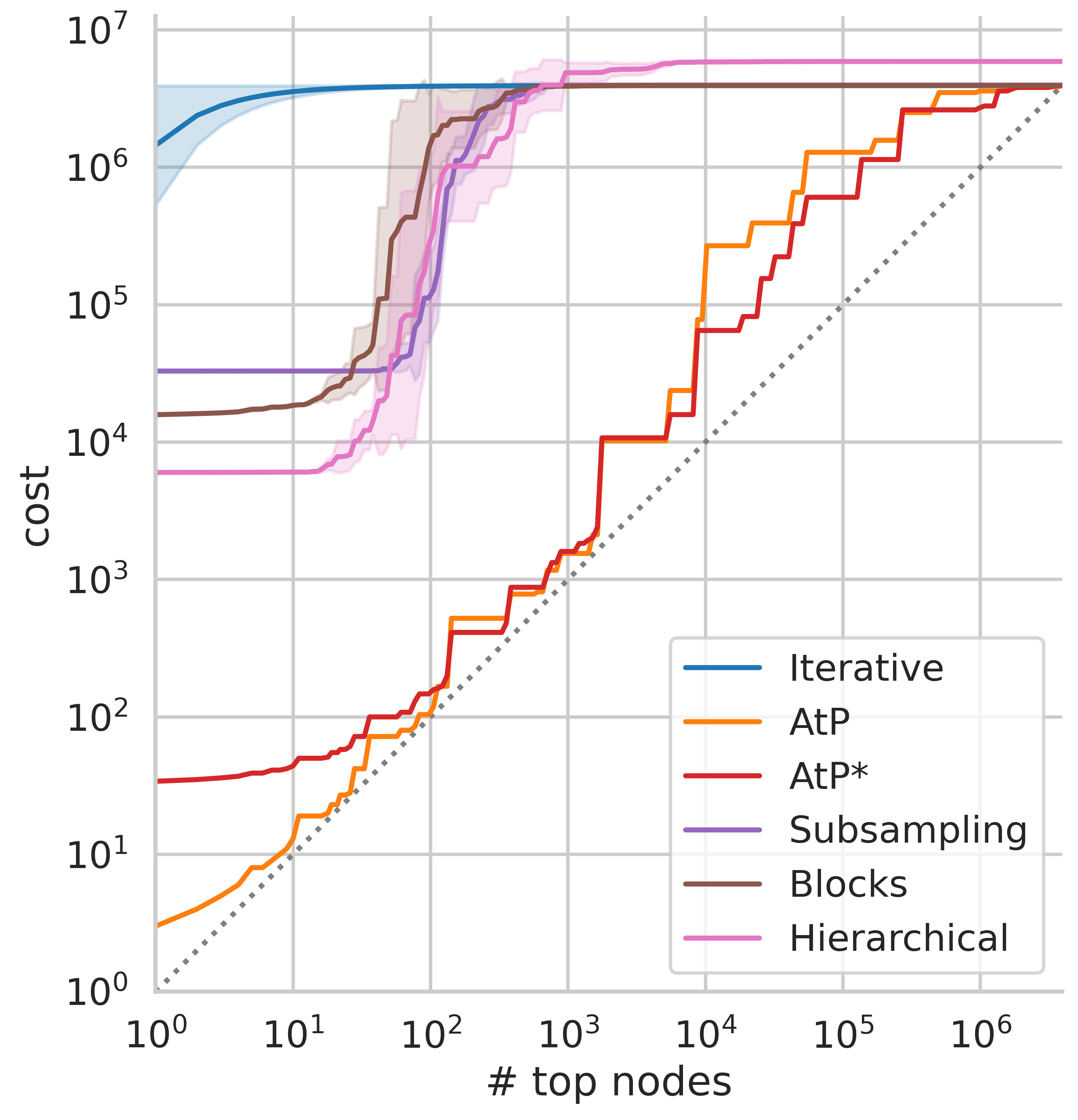}
            \caption{Pythia 2.8B}
        \end{subsubfigure}
        \begin{subsubfigure}[b]{0.24\textwidth}
            \includegraphics[width=\textwidth]{rp_mlp_cost_12b.png}
            \caption{Pythia 12B}
        \end{subsubfigure}
        \caption{\texttt{RAND-PP}}
    \end{subfigure}
    \begin{subfigure}[b]{\textwidth}
    \setcounter{subsubfigure}{0}
        \begin{subsubfigure}[b]{0.24\textwidth}
            \includegraphics[width=\textwidth]{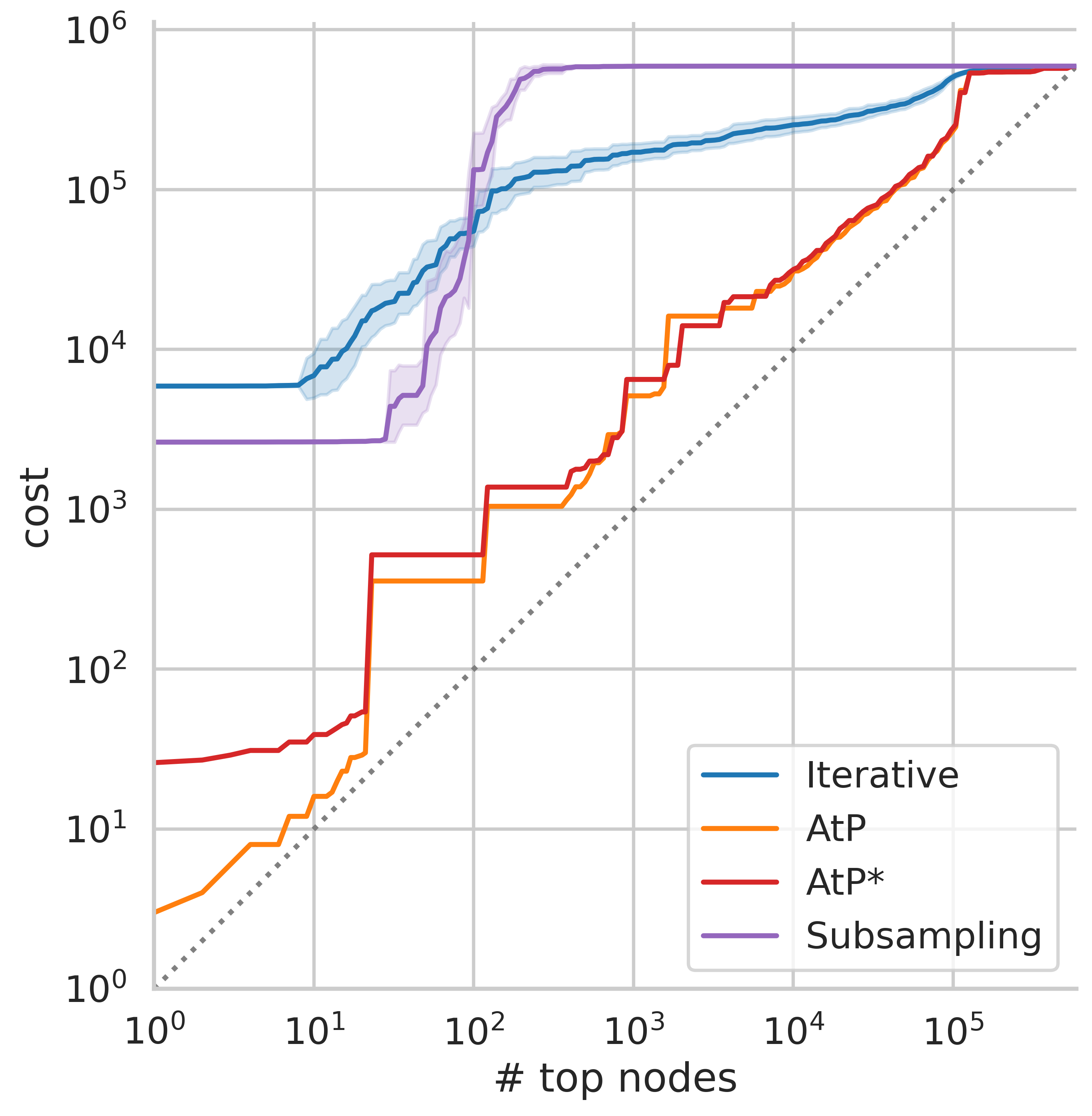}
            \caption{Pythia 410M}
        \end{subsubfigure}
        \begin{subsubfigure}[b]{0.24\textwidth}
            \includegraphics[width=\textwidth]{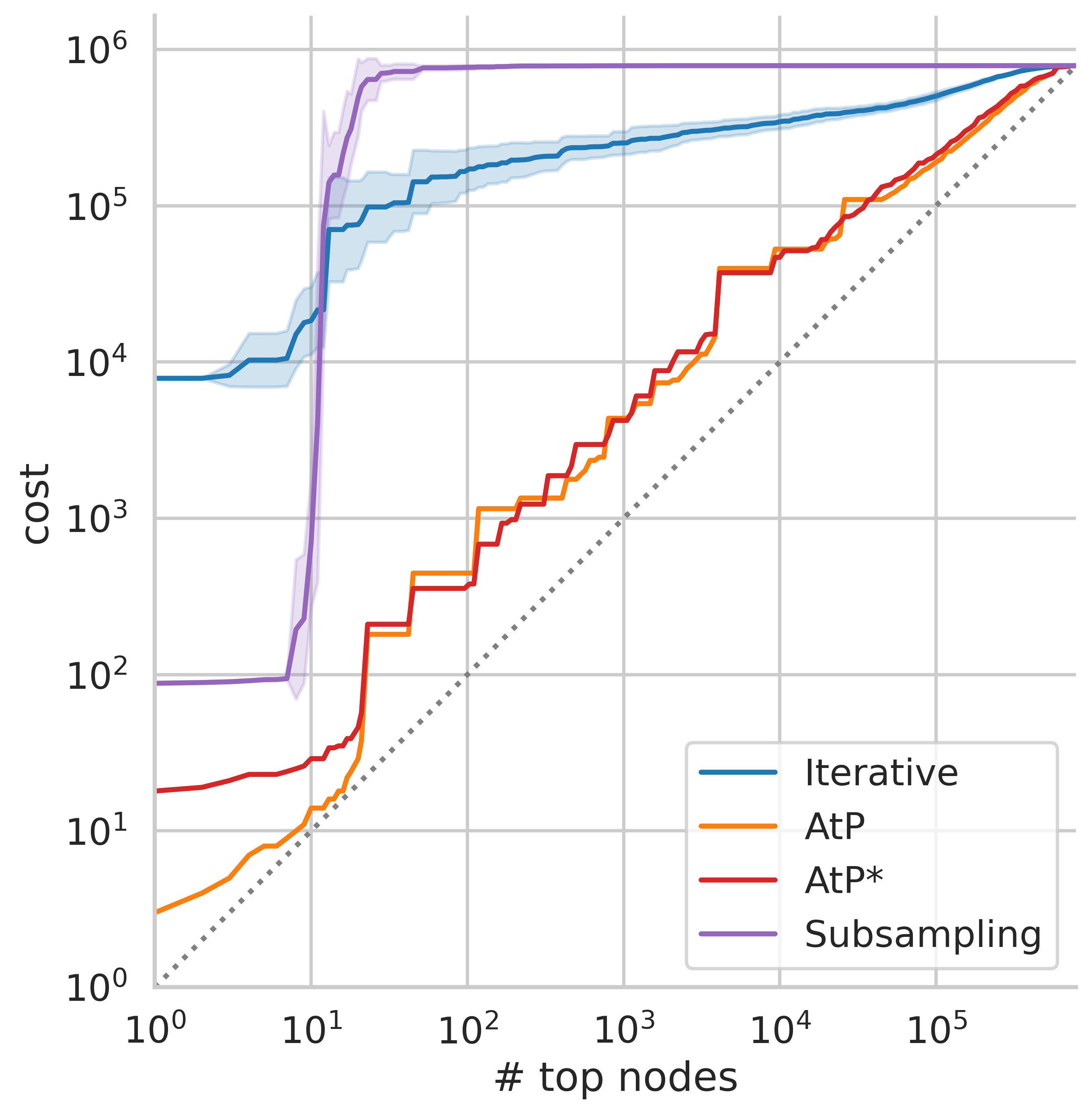}
            \caption{Pythia 1B}
        \end{subsubfigure}
        \begin{subsubfigure}[b]{0.24\textwidth}
            \includegraphics[width=\textwidth]{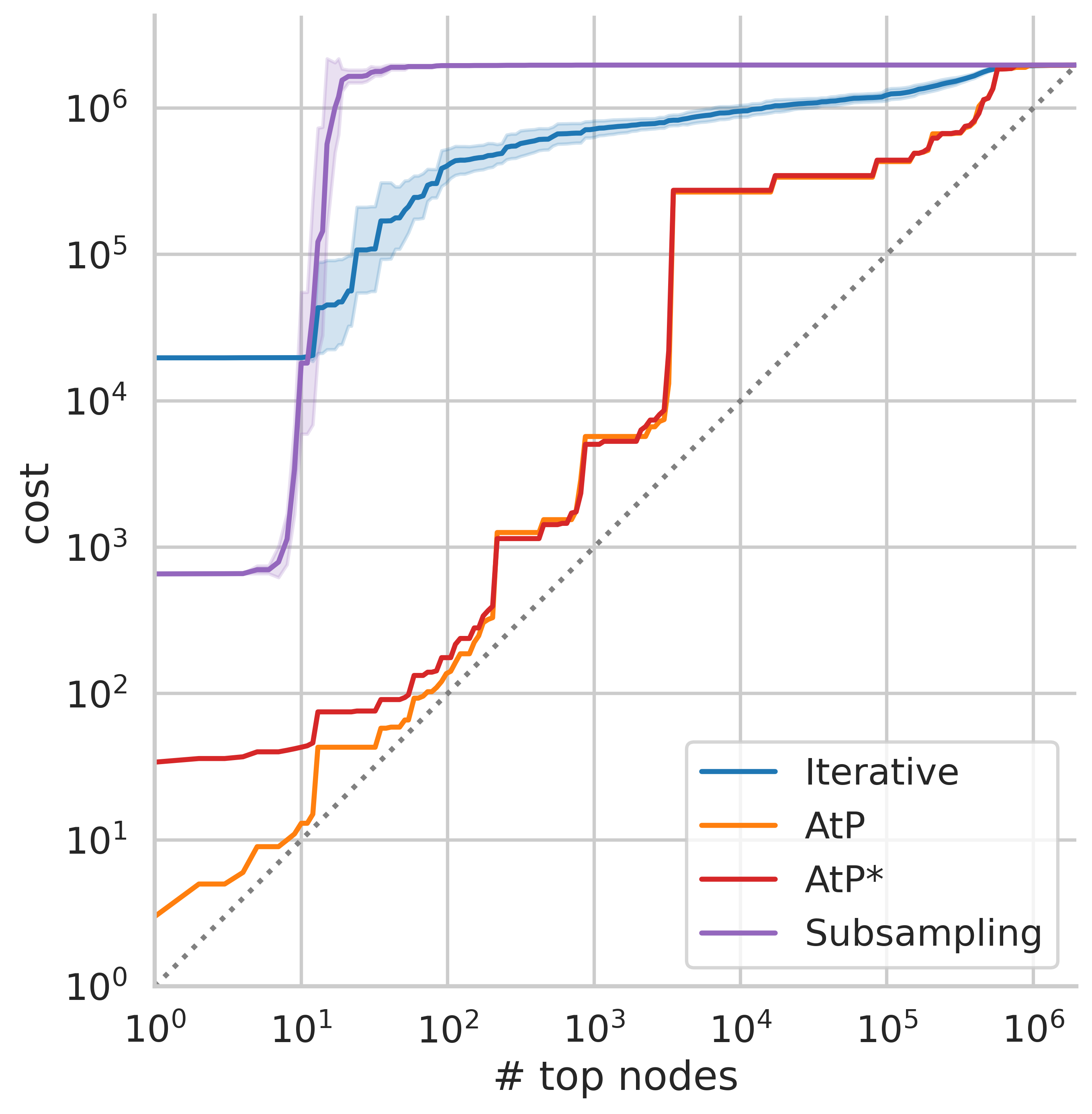}
            \caption{Pythia 2.8B}
        \end{subsubfigure}
        \begin{subsubfigure}[b]{0.24\textwidth}
            \includegraphics[width=\textwidth]{assets/a_an_cost_pythia_12b.png}
            \caption{Pythia 12B}
        \end{subsubfigure}
        \caption{\texttt{A-AN} distribution}
    \end{subfigure}
    \label{fig:mlp_costs}
\end{figure}

\begin{figure}
    \centering
    \caption{Cost of verified 100\% recall curves, sweeping across models and settings for \AttentionNodes{}}
    \begin{subfigure}[b]{\textwidth}
    \setcounter{subsubfigure}{0}
        \begin{subsubfigure}[b]{0.24\textwidth}
            \includegraphics[width=\textwidth]{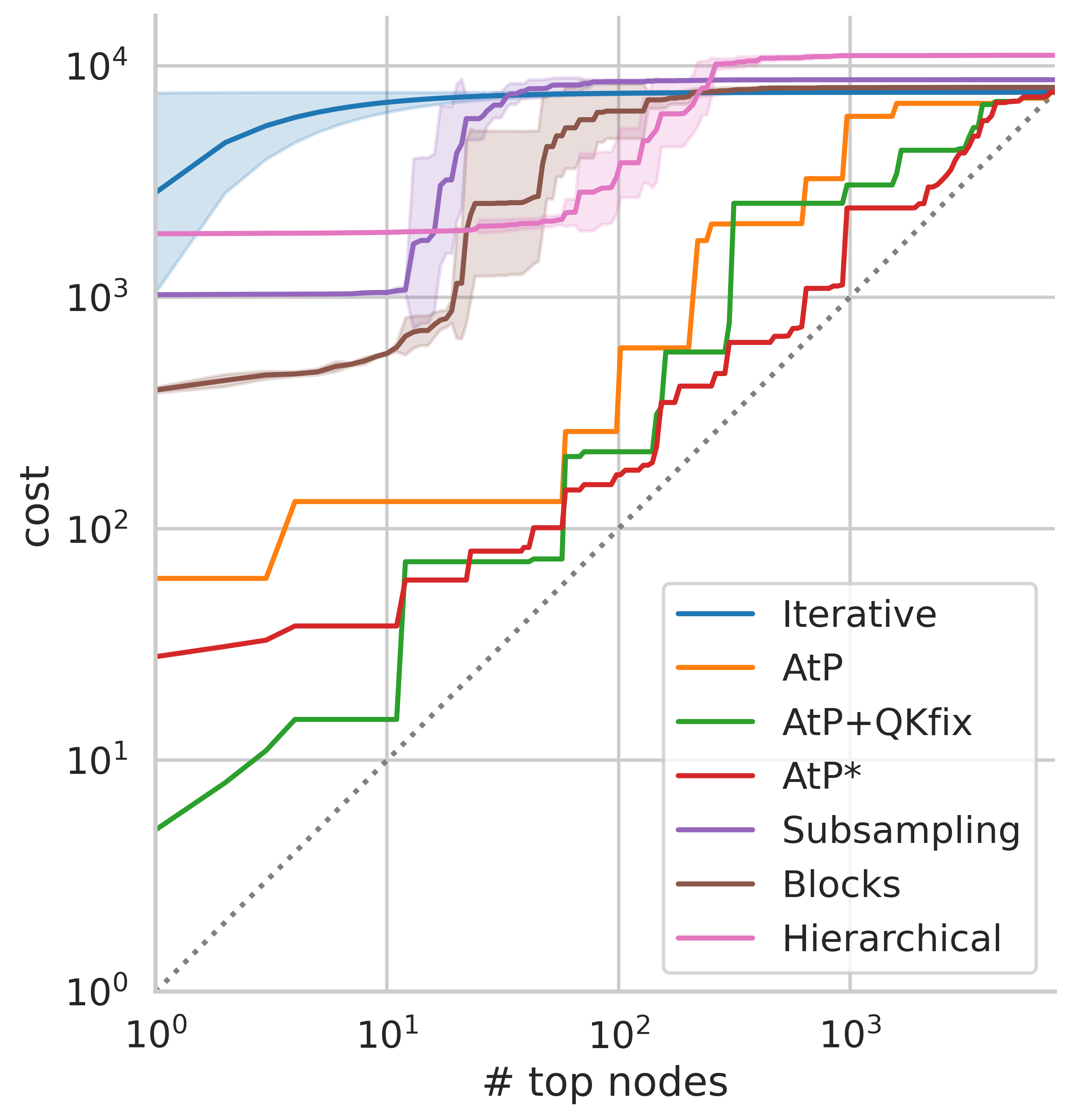}
            \caption{Pythia 410M}
        \end{subsubfigure}
        \begin{subsubfigure}[b]{0.24\textwidth}
            \includegraphics[width=\textwidth]{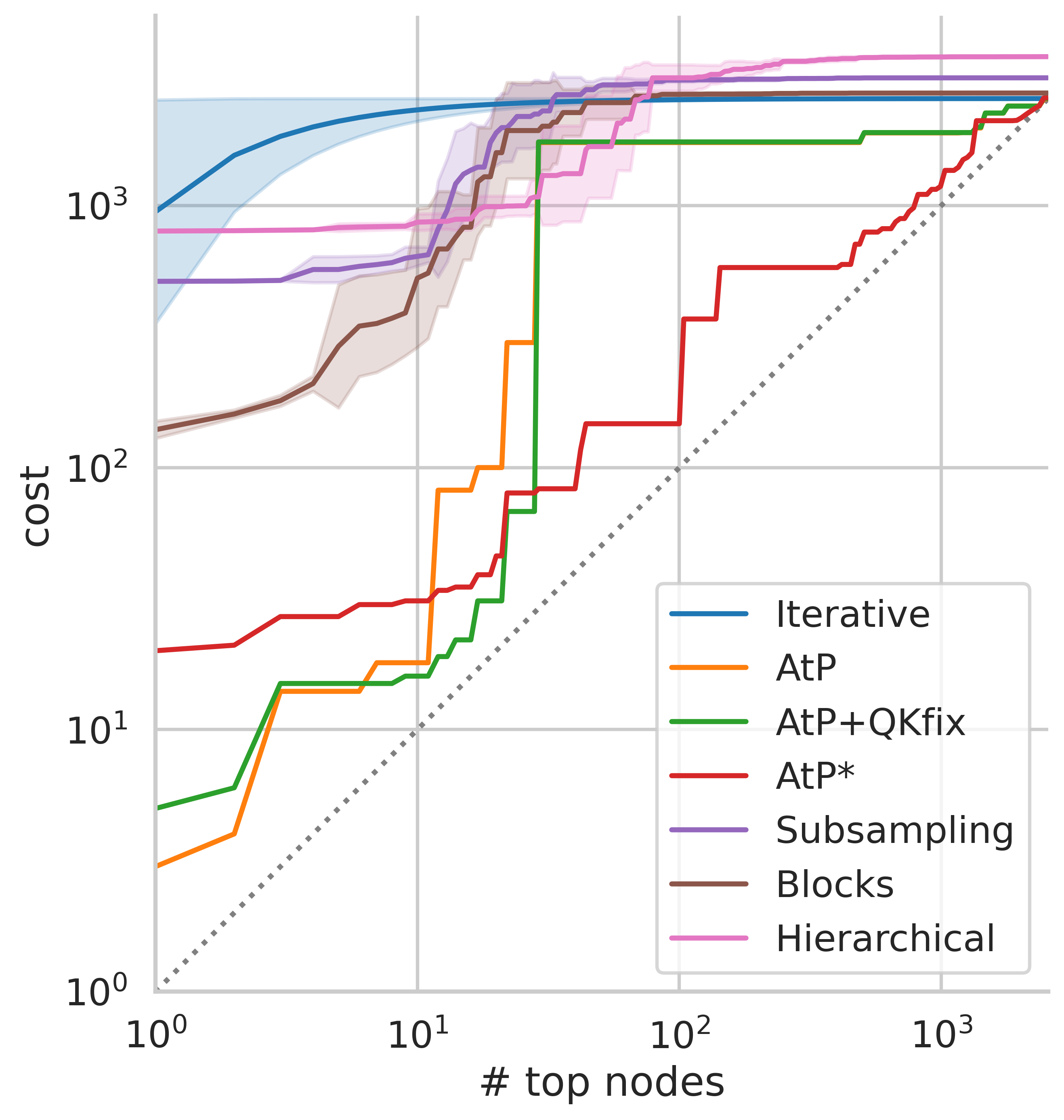}
            \caption{Pythia 1B}
        \end{subsubfigure}
        \begin{subsubfigure}[b]{0.24\textwidth}
            \includegraphics[width=\textwidth]{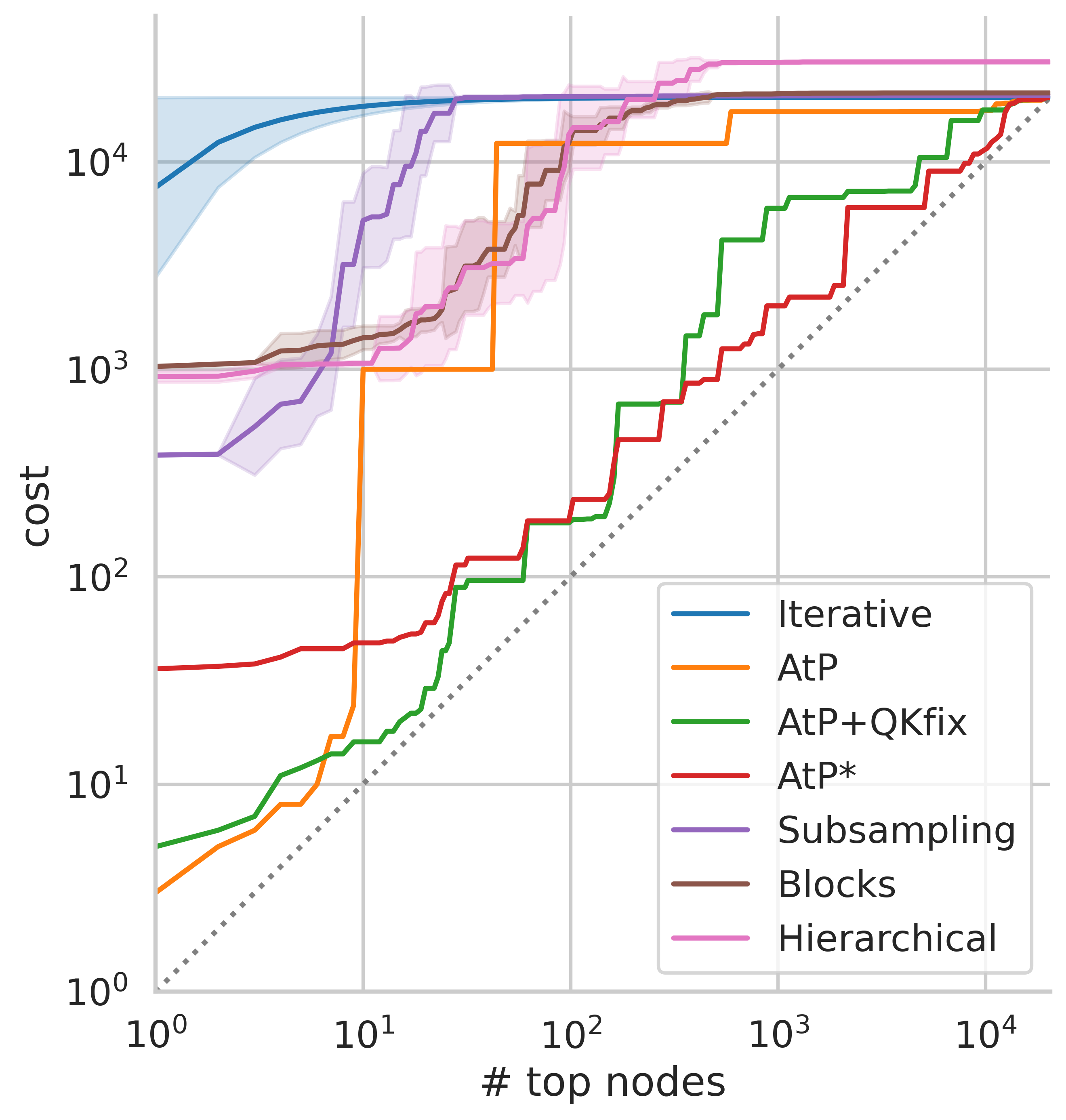}
            \caption{Pythia 2.8B}
        \end{subsubfigure}
        \begin{subsubfigure}[b]{0.24\textwidth}
            \includegraphics[width=\textwidth]{ioipp_cost_pythia-12b.png}
            \caption{Pythia 12B}
        \end{subsubfigure}
        \caption{\texttt{IOI-PP}}
    \end{subfigure}
    \begin{subfigure}[b]{\textwidth}
    \setcounter{subsubfigure}{0}
        \begin{subsubfigure}[b]{0.24\textwidth}
            \includegraphics[width=\textwidth]{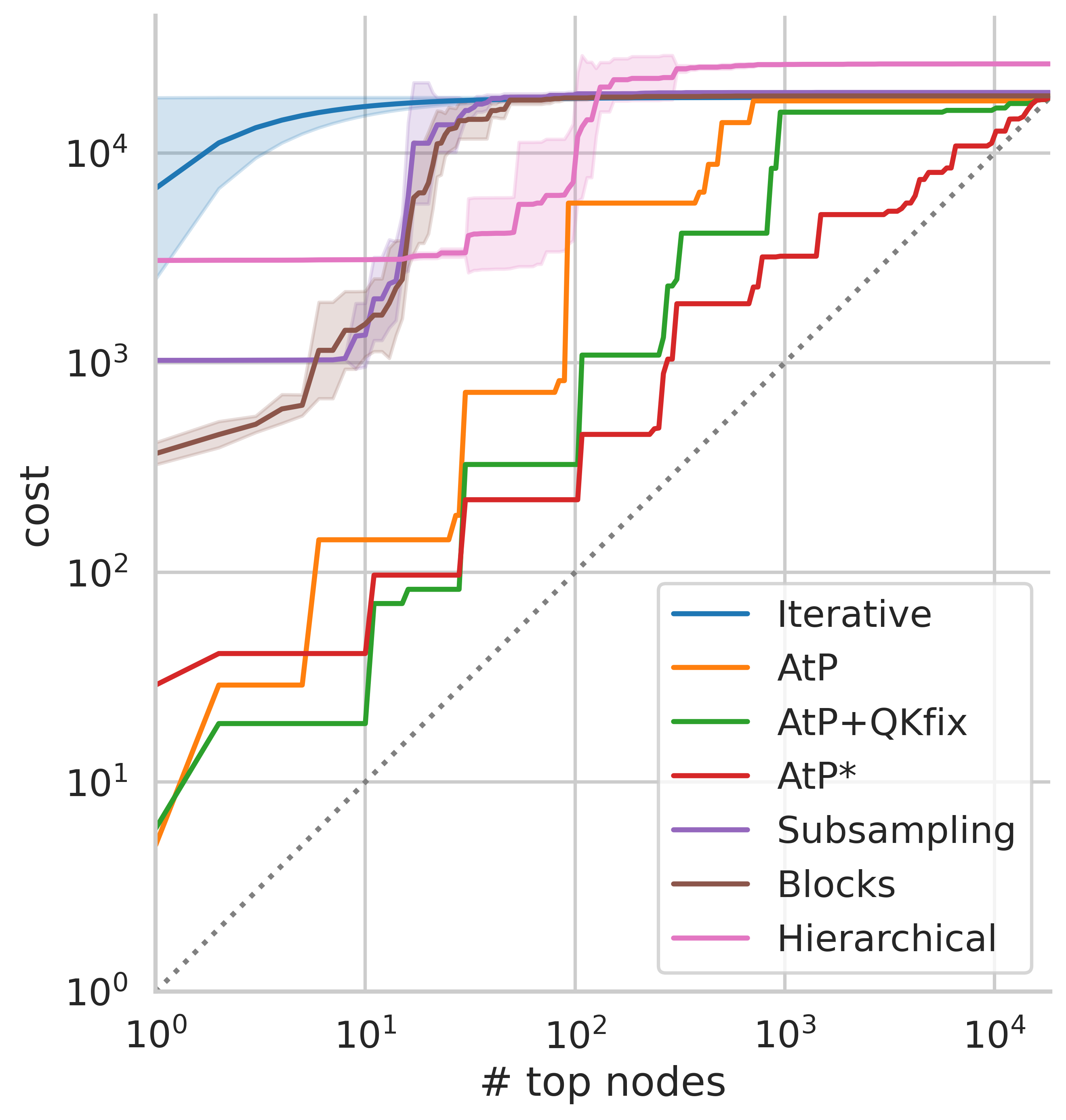}
            \caption{Pythia 410M}
        \end{subsubfigure}
        \begin{subsubfigure}[b]{0.24\textwidth}
            \includegraphics[width=\textwidth]{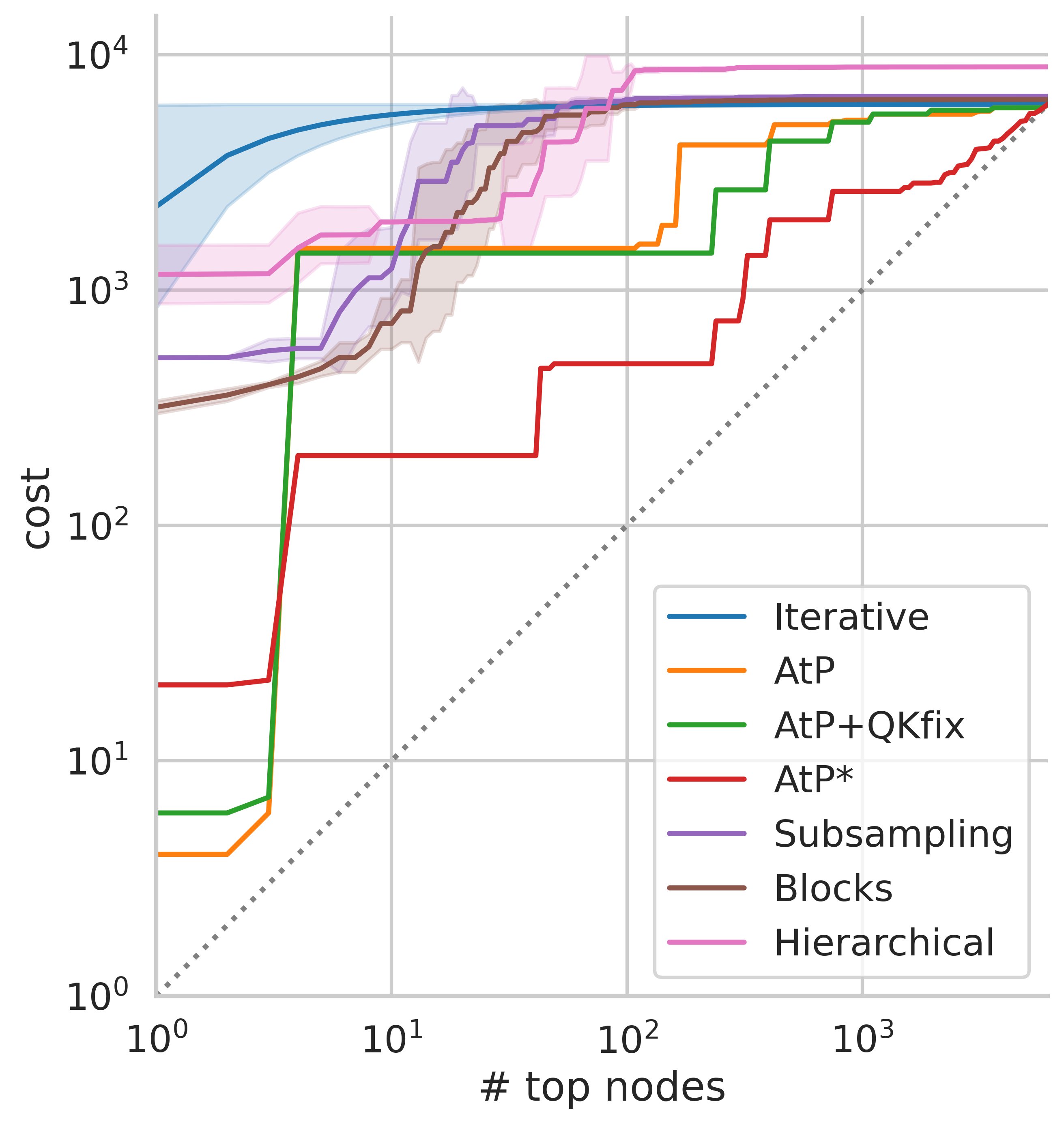}
            \caption{Pythia 1B}
        \end{subsubfigure}
        \begin{subsubfigure}[b]{0.24\textwidth}
            \includegraphics[width=\textwidth]{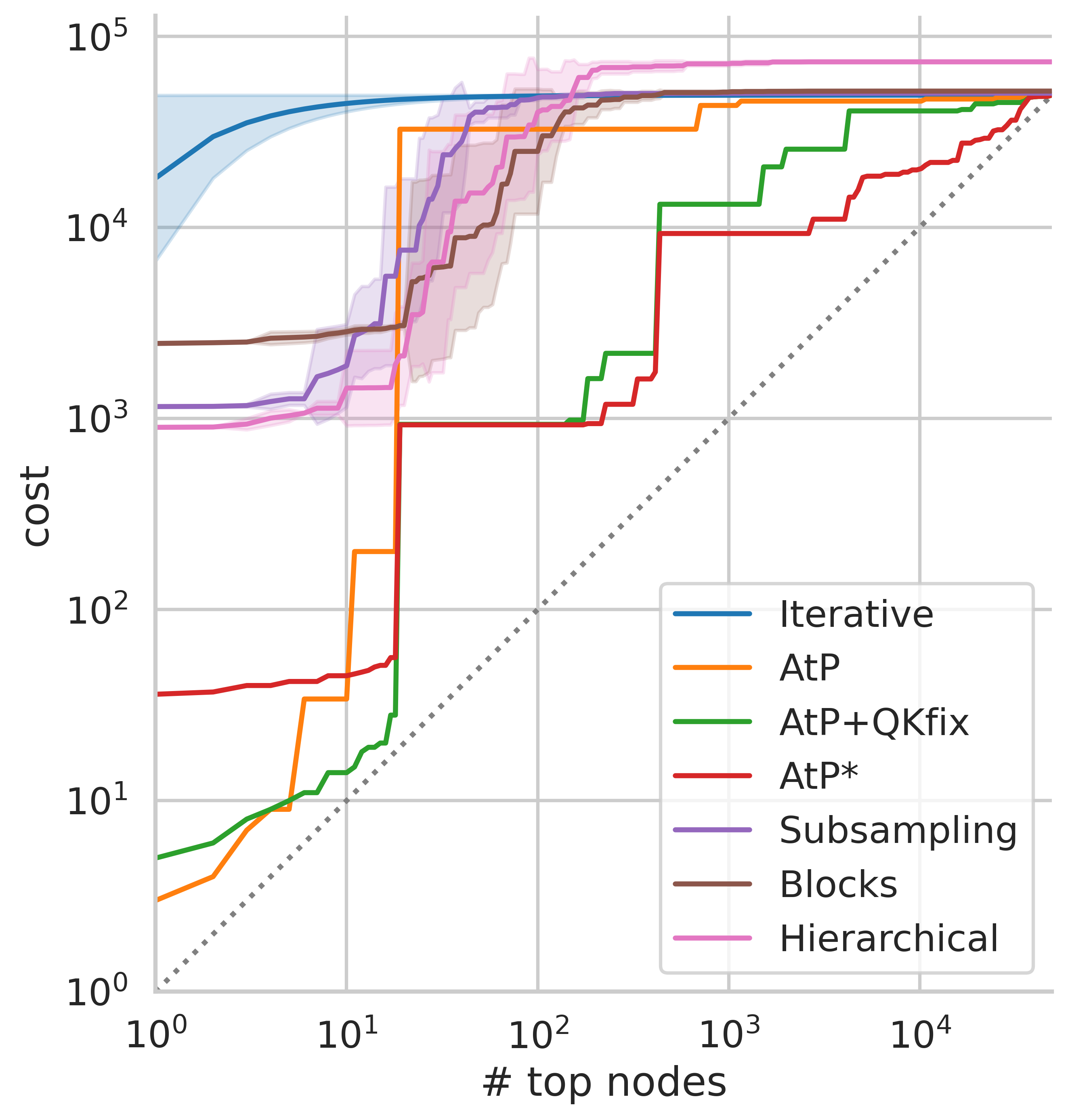}
            \caption{Pythia 2.8B}
        \end{subsubfigure}
        \begin{subsubfigure}[b]{0.24\textwidth}
            \includegraphics[width=\textwidth]{rp_attn_cost_12b.png}
            \caption{Pythia 12B}
        \end{subsubfigure}
        \caption{\texttt{RAND-PP}}
    \end{subfigure}
    \begin{subfigure}[b]{\textwidth}
    \setcounter{subsubfigure}{0}
        \begin{subsubfigure}[b]{0.24\textwidth}
            \includegraphics[width=\textwidth]{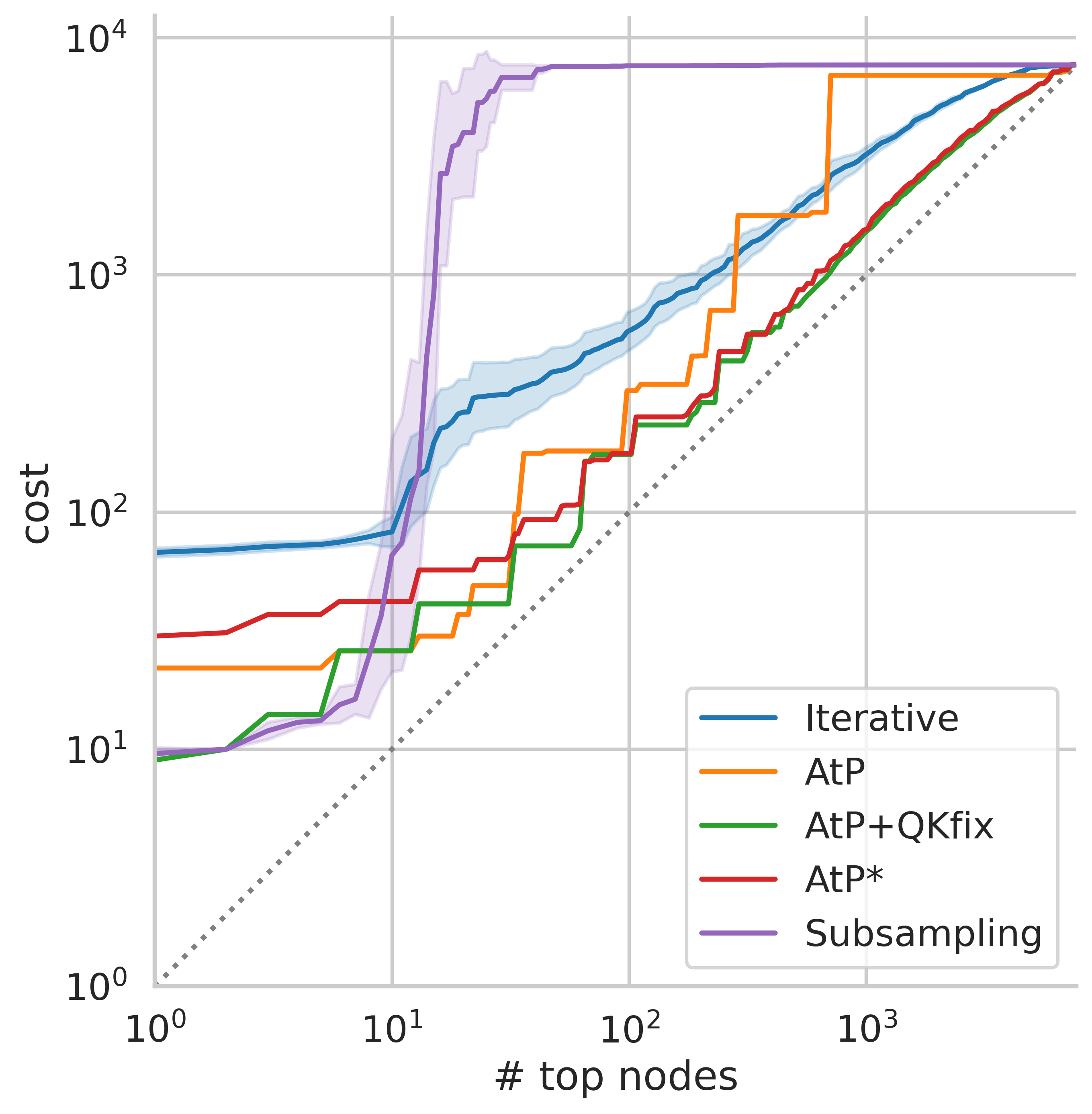}
            \caption{Pythia 410M}
        \end{subsubfigure}
        \begin{subsubfigure}[b]{0.24\textwidth}
            \includegraphics[width=\textwidth]{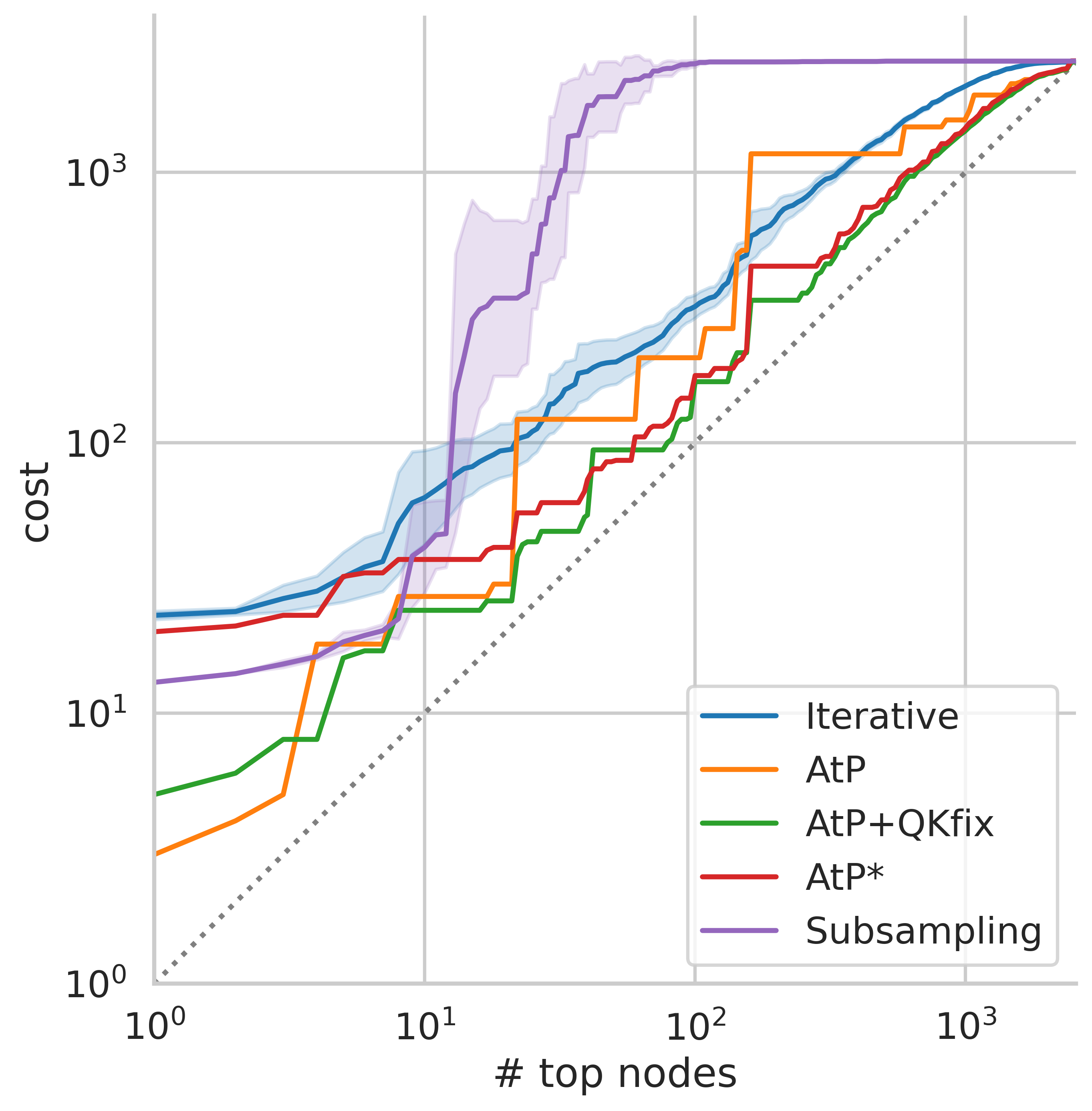}
            \caption{Pythia 1B}
        \end{subsubfigure}
        \begin{subsubfigure}[b]{0.24\textwidth}
            \includegraphics[width=\textwidth]{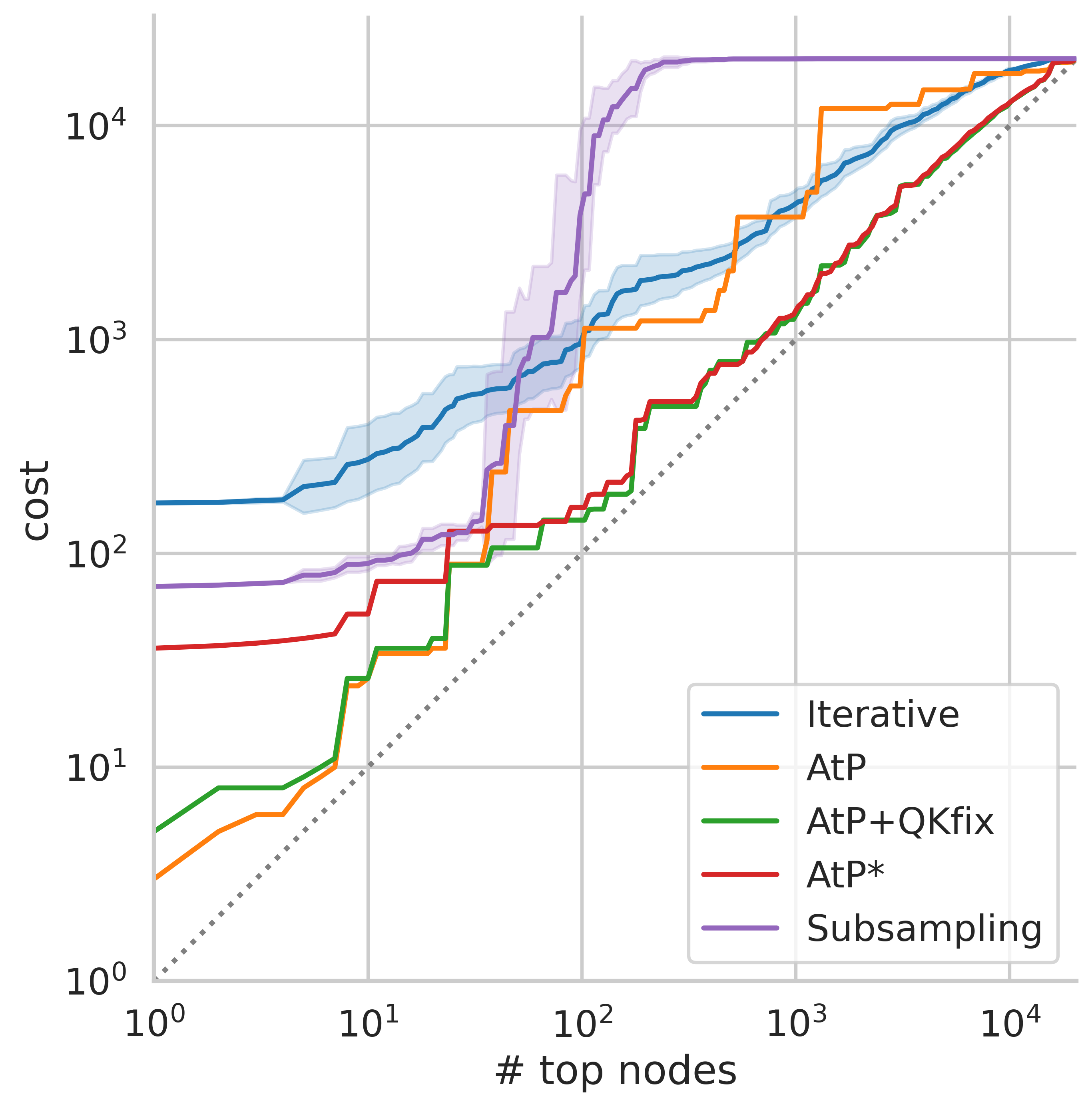}
            \caption{Pythia 2.8B}
        \end{subsubfigure}
        \begin{subsubfigure}[b]{0.24\textwidth}
            \includegraphics[width=\textwidth]{ioi_cost_12b.png}
            \caption{Pythia 12B}
        \end{subsubfigure}
        \caption{\texttt{IOI} distribution}
    \end{subfigure}
    \label{fig:attn_costs}
\end{figure}

We show the diagnostic measurements for Pythia-12B across all investigated distributions in~\Cref{fig:diagnostics_all_distributions}, and cost of verified 100\% recall curves for all models and settings in \Cref{fig:mlp_costs,fig:attn_costs}.


\subsection{Metrics} \label{app:metrics}
\begin{figure}
    \centering
    \includegraphics[width=\textwidth]{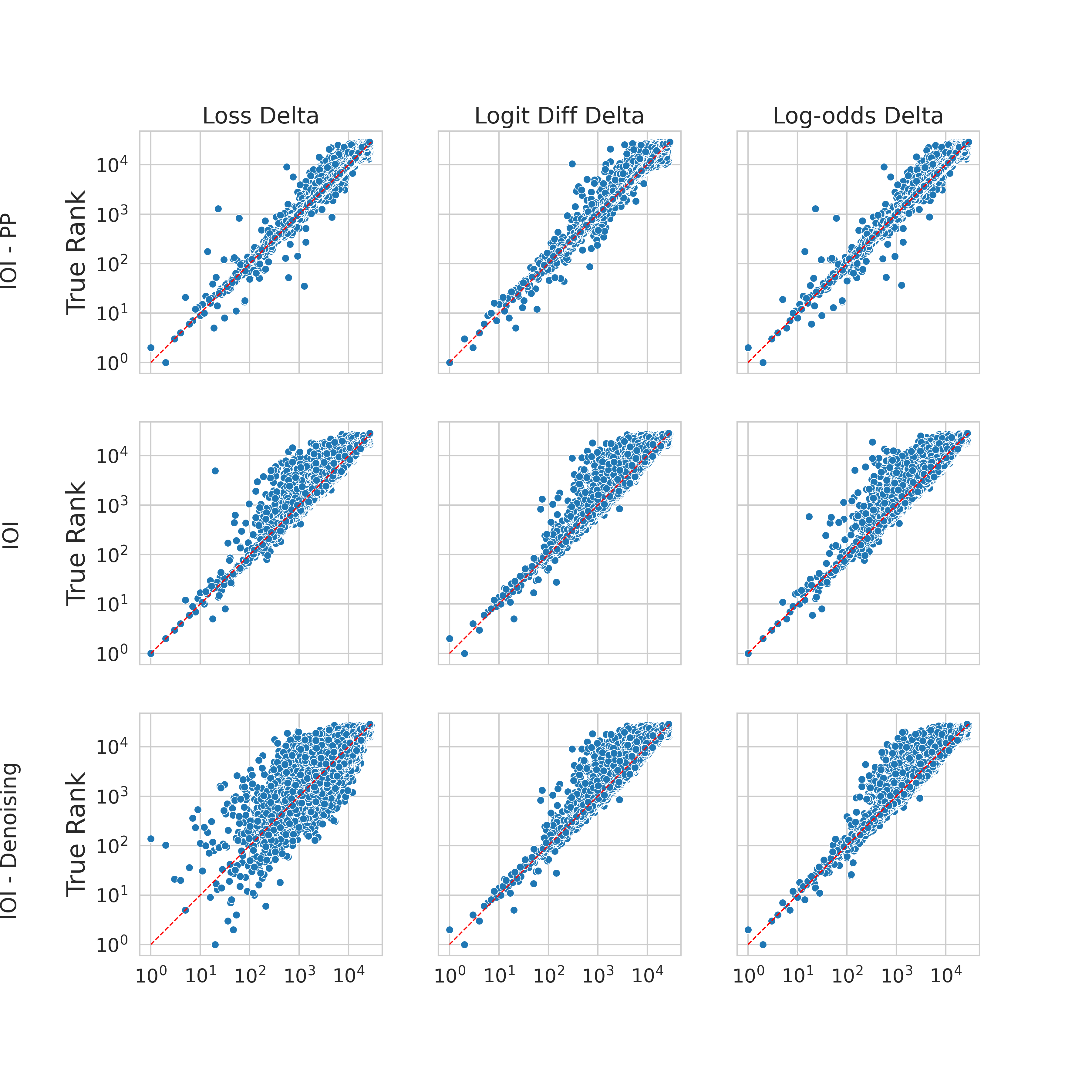}
    \caption{True ranks against AtP$^*$ ranks on Pythia-12B using various metrics $\metric$. The last row shows the effect in the denoising (rather than noising) setting; we speculate that the lower-right subplot (log-odds denoising) is similar to the lower-middle one (logit-diff denoising) because IOI produces a bimodal distribution over the correct and alternate next token.}
    \label{fig:atp_metrics}
\end{figure}

In this paper we focus on the difference in loss (negative log probability) as the metric $\metric$. We provide some evidence that AtP($^*$) is not sensitive to the choice of $\metric$. For Pythia-12B, on \texttt{IOI-PP} and \texttt{IOI}, we show the rank scatter plots in ~\Cref{fig:atp_metrics} for three different metrics.

For \texttt{IOI}, we also show that performance of AtP$^*$ looks notably worse when effects are evaluated via denoising instead of noising (cf.~\Cref{sec:problem-statement}). As of now we do not have a satisfactory explanation for this observation.

\subsection{Hyperparameter selection}\label{app:hyper_selection}

The \emph{iterative} baseline, and the \emph{AtP}-based methods, have no hyperparameters. In general, we used 5 random seeds for each hyperparameter setting, and selected the setting that produced the lowest IRWRGM cost (see~\Cref{sec:irwrgm}).

For \emph{Subsampling}, the two hyperparameters are the Bernoulli sampling probability $p$, and the number of samples to collect before verifying nodes in decreasing order of $\hat\contrib_{\text{SS}}$. $p$ was chosen from \{0.01, 0.03\}\footnote{We observed early on that larger values of $p$ were consistently underperforming. We leave it to future work to investigate more granular and smaller values for $p$.}. The number of steps was chosen among power-of-2 numbers of batches, where the batch size depended on the setting.

For \emph{Blocks}, we swept across block sizes 2, 6, 20, 60, 250. For \emph{Hierarchical}, we used a branching factor of $B=3$, because of the following heuristic argument. If all but one node had zero effect, then discovering that node would be a matter of iterating through the hierarchy levels. We'd have number of levels $\log_B |\Nodes|$, and at each level, $B$ forward passes would be required to find which lower-level block the special node is in -- and thus the cost of finding the node would be $B\log_B |\Nodes|=\frac{B}{\log B}\log|\Nodes|$. $\frac{B}{\log B}$ is minimized at $B=e$, or at $B=3$ if $B$ must be an integer. The other hyperparameter is the number of levels; we swept this from 2 to 12.

\section{AtP variants}

\subsection{Residual-site AtP and Layer normalization}\label{app:layernorm}

Let's consider the behaviour of AtP on sites that contain much or all of the total signal in the residual stream, such as residual-stream sites. \citet{neel2022attribution} described a concern about this behaviour: that linear approximation of the layer normalization would do poorly if the patched value is significantly different than the clean one, but with a similar norm. The proposed modification to AtP to account for this was to hold the scaling factors (in the denominators) fixed when computing the backwards pass. Here we'll present an analysis of how this modification would affect the approximation error of AtP. (Empirical investigation of this issue is beyond the scope of this paper.)

Concretely, let the node under consideration be $\node$, with clean and alternate values $\node^{\mathrm{clean}}$ and $\node^{\mathrm{noise}}$; and for simplicity, let's assume the model does nothing more than an unparametrized RMSNorm $\model(\node):=\node/|\node|$. Let's now consider how well $\model(\node^{\mathrm{noise}})$ is approximated, both by its first-order approximation $\hat \model_{\text{AtP}}(\node^{\mathrm{noise}}) := \model(\node^{\mathrm{clean}}) + \model(\node^{\mathrm{clean}})^\perp (\node^{\mathrm{noise}}-\node^{\mathrm{clean}})$ where $\model(\node^{\mathrm{clean}})^\perp=I-\model(\node^{\mathrm{clean}})\model(\node^{\mathrm{clean}})^\intercal$ is the projection to the hyperplane orthogonal to $\model(\node^{\mathrm{clean}})$, and by the variant that fixes the denominator: $\hat \model_{\text{AtP+frozenLN}}(\node^{\mathrm{noise}}) := \node^{\mathrm{noise}}/|\node^{\mathrm{clean}}|$.

To quantify the error in the above, we'll measure the error $\epsilon$ in terms of Euclidean distance. Let's also assume, without loss of generality, that $|\node^{\mathrm{clean}}|=1$. Geometrically, then, $\model(\node)$ is a projection onto the unit hypersphere, $\model_{\text{AtP}}(\node)$ is a projection onto the tangent hyperplane at $\node^{\mathrm{clean}}$, and $\model_{\text{AtP+frozenLN}}$ is the identity function.

Now, let's define orthogonal coordinates $(x,y)$ on the plane spanned by $\node^{\mathrm{clean}},\node^{\mathrm{noise}}$, such that $\node^{\mathrm{clean}}$ is mapped to $(1,0)$ and $\node^{\mathrm{noise}}$ is mapped to $(x,y)$, with $y\geq 0$. Then, $\epsilon_{\text{AtP}} := \left|\hat \model(\node^{\mathrm{noise}}) - \model(\node^{\mathrm{noise}})\right| = \sqrt{2+y^2-2\frac{x+y^2}{\sqrt{x^2+y^2}}}$, while $\epsilon_{\text{AtP+frozenLN}} := \left|\hat \model_{\mathrm{fix}}(\node^{\mathrm{noise}}) - \model(\node^{\mathrm{noise}})\right|=\left|\sqrt{x^2+y^2}-1\right|$.

\begin{figure}
    \centering
    \begin{subfigure}[b]{.45\textwidth}
        \centering
        \includegraphics[width=\textwidth]{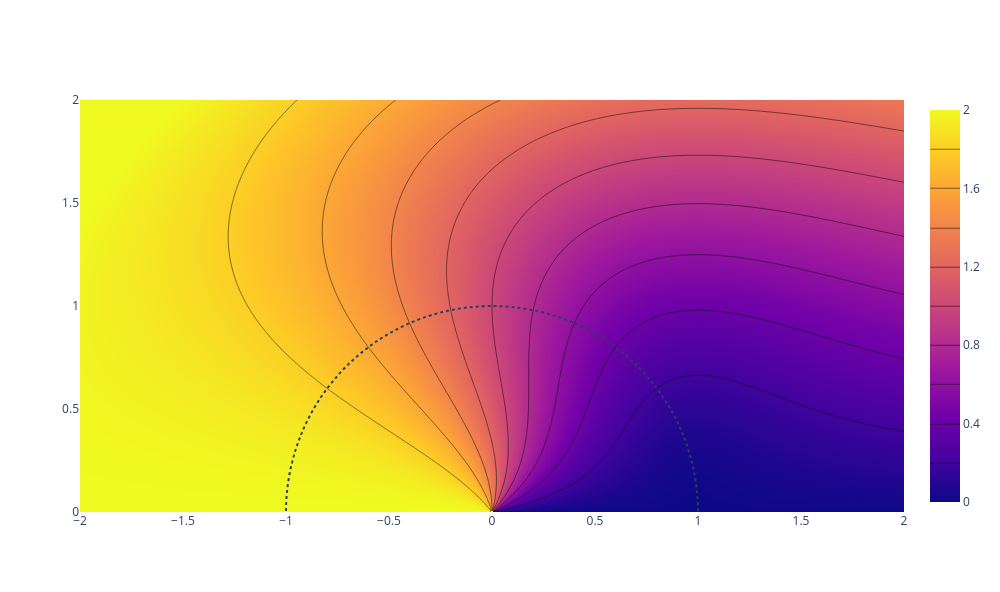}
        \caption{$\epsilon_{\text{AtP}}$}
    \end{subfigure}
    \hfill
    \begin{subfigure}[b]{.45\textwidth}
        \centering
        \includegraphics[width=\textwidth]{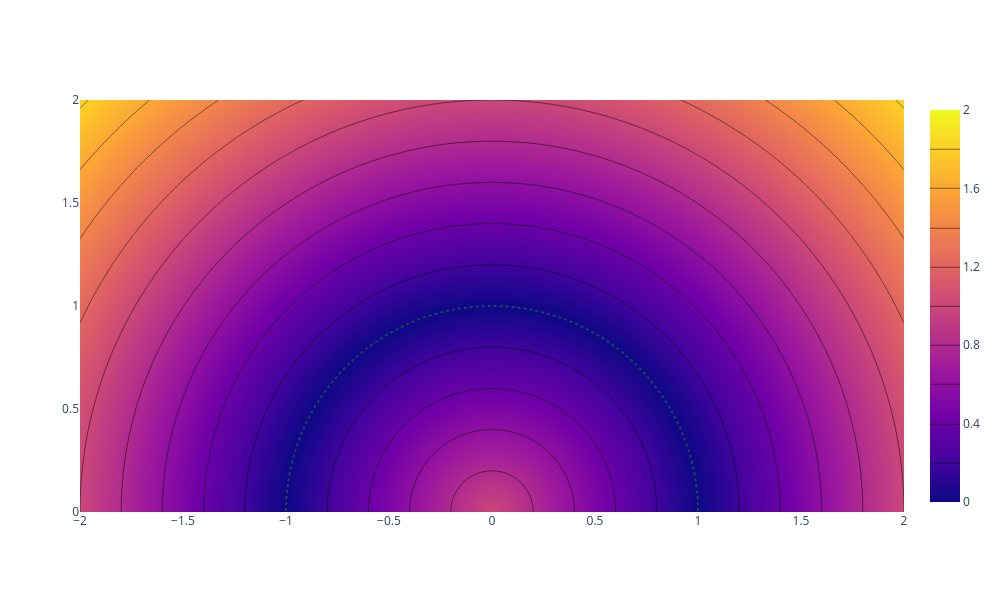}
        \caption{$\epsilon_{\text{AtP+frozenLN}}$}
    \end{subfigure}
    \begin{subfigure}[b]{\textwidth}
        \centering

        \includegraphics[width=\textwidth]{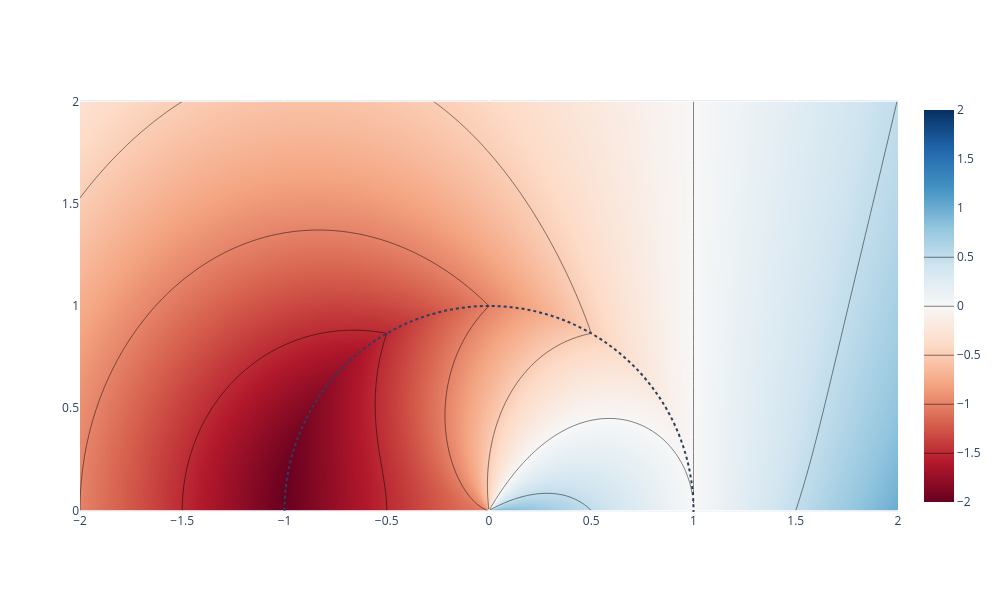}
        \caption{$\epsilon_{\text{AtP+frozenLN}}-\epsilon_{\text{AtP}}$}
    \end{subfigure}
    \caption{A comparison of how AtP and AtP with frozen layernorm scaling behave in a toy setting where the model we're trying to approximate is just $\model(\node):=\node/|\node|$. The red region is where frozen layernorm scaling helps; the blue region is where it hurts. We find that unless $x>1$, frozen layernorm scaling always has lower error when the cosine-similarity between $\node^{\mathrm{noise}}$ and $\node^{\mathrm{clean}}$ is $<\frac{1}{2}$ (in other words the angle $>60^\circ$), but often has higher error otherwise.}
    \label{fig:layernorm}
\end{figure}

Plotting the error in \Cref{fig:layernorm}, we can see that, as might be expected, freezing the layer norm denominators helps whenever $\node^{\mathrm{noise}}$ indeed has the same norm as $\node^{\mathrm{clean}}$, and (barring weird cases with $x>1$) whenever the cosine-similarity is less than $\frac{1}{2}$; but largely hurts if $\node^{\mathrm{noise}}$ is close to $\node^{\mathrm{clean}}$. This illustrates that, while freezing the denominators will generally be unhelpful when patch distances are small relative to the full residual signal (as with almost all nodes considered in this paper), it will likely be helpful in a different setting of patching residual streams, which could be quite unaligned but have similar norm.

\subsection{Edge AtP and AtP*}\label{app:edge_atp}

Here we will investigate edge attribution patching, and how the cost scales if we use GradDrop and/or QK fix. (For this section we'll focus on a single prompt pair.)

First, let's review what edge attribution patching is trying to approximate, and how it works.

\subsubsection{Edge intervention effects}

Given nodes $\node_1,\node_2$ where $\node_1$ is upstream of $\node_2$, if we were to patch in an alternate value for $\node_1$, this could impact $\node_2$ in a complicated nonlinear way. As discussed in \ref{sec:cancellation}, because LLMs have a residual stream, the ``direct effect'' can be understood as the one holding all other possible intermediate nodes between $n_1$ and $n_2$ fixed -- and it's a relatively simple function, composed of transforming the alternate value $\node_1(\xnoise)$ to a residual stream contribution $r_{\text{out},{\ell_1}}(\xclean|\doGets{\node_1}{\node_1(\xnoise)})$, then carrying it along the residual stream to an input $r_{\text{in},{\ell_2}}=r_{\text{in},{\ell_2}}(\xclean)+(r_{\text{out},{\ell_1}} - r_{\text{out},{\ell_1}}(\xclean))$, and transforming that into a value $\node_2^{\text{direct}}$.

In the above, $\ell_1$ and $\ell_2$ are the semilayers containing $n_1$ and $n_2$, respectively. Let's define $\mathbf{n}_{(\ell_1,\ell_2)}$ to be the set of non-residual nodes between semilayers $\ell_1$ and $\ell_2$. Then, we can define the resulting $\node_2^{\text{direct}}$ as: $$\node_2^{\text{direct}^{\ell_1}}(\xclean|\doGets{\node_1}{\node_1(\xnoise)}):=\node_2(\xclean|\doGets{\node_1}{\node_1(\xnoise)}, \doGets{\mathbf{n}_{(\ell_1,\ell_2)}}{\mathbf{n}_{(\ell_1,\ell_2)}(\xclean)}).$$

The residual-stream input $r_{\text{in},{\ell_2}}^{\text{direct}^{\ell_1}}(\xclean|\doGets{\node_1}{\node_1(\xnoise)})$ is defined similarly.

Finally, $\node_2$ itself isn't enough to compute the metric $\metric$ -- for that we also need to let the forward pass $\model(x^{\text{clean}})$ run using the modified $\node_2^{\text{direct}^{\ell_1}}(\xclean|\doGets{\node_1}{\node_1(\xnoise)})$, while removing all other effects of $n_1$ (i.e.\ not patching it).

Writing this out, we have edge intervention effect
\begin{align}
\inter(\node_1\rightarrow \node_2;\xclean,\xnoise)
&:=\metric(\model(\xclean|\doGets{\node_2}{\node_2^{\text{direct}^{\ell_1}}(\xclean|\doGets{\node_1}{\node_1(\xnoise)})}))\notag\\
&\mathrel{\phantom{:=}}-\metric(\model(\xclean)).
\end{align}

\subsubsection{Nodes and Edges}

Let's briefly consider what edges we'd want to be evaluating this on. In \cref{sec:setup}, we were able to conveniently separate attention nodes from MLP neurons, knowing that to handle both kinds of nodes, we'd just need to be able handle each kind of node on its own, and then combine the results. For edge interventions this of course isn't true, because edges can go from MLP neurons to attention nodes, and vice versa. For the purposes of this section, we'll assume that the node set $\Nodes$ contains the attention nodes, and for MLPs either a node per layer (as in \citet{syed2023attribution}), or a node per neuron (as in the \NeuronNodes{} setting).

Regarding the edges, the MLP nodes can reasonably be connected with any upstream or downstream node, but this isn't true for the attention nodes, which have more of a structure amongst themselves: the key, query, and value nodes for an attention head can only affect downstream nodes via the attention output nodes for that head, and vice versa. As a result, on edges between different semilayers, upstream attention nodes must be attention head outputs, and downstream attention nodes must be keys, queries, or values. In addition, there are some within-attention-head edges, connecting each query node to the output node in the same position, and each key and value node to output nodes in causally affectable positions.

\subsubsection{Edge AtP}

As with node activation patching, the edge intervention effect $\inter(\node_1\rightarrow \node_2;\xclean,\xnoise)$ is costly to evaluate directly for every edge, since a forward pass is required each time. However, as with AtP, we can apply first-order approximations: we define
\begin{align}
    \hat\inter_{\text{AtP}}(\node_1\rightarrow \node_2;\xclean,\xnoise)
    &:=\left(\Delta r_{\node_1}^{\text{AtP}}(\xclean, \xnoise)\right)^\intercal \nabla_{r_{\node_2}}^{\text{AtP}}\metric(\model(\xclean)),\\
    \text{where }\Delta r_{\node_1}^{\text{AtP}}(\xclean, \xnoise)
    &:=\operatorname{Jac}_{\node_1}(r_{\text{out},{\ell_1}})(\node_1(\xclean)) (\node_1(\xnoise) - \node_1(\xclean))\\
    \text{and }\nabla_{r_{\node_2}}^{\text{AtP}}\metric(\model(\xclean))
    &:=\left(\operatorname{Jac}_{r_{\text{in},{\ell_2}}}(\node_2)(r_{\text{in},{\ell_2}}(\xclean))\right)^\intercal
    \nabla_{\node_2}(\metric(\model(\xclean)))(\node_2(\xclean)),
\end{align}

and this is a close approximation when $\node_1(\xnoise)\approx \node_1(\xclean)$.

A key benefit of this decomposition is that the first term depends only on $\node_1$, and the second term depends only on $\node_2$; and they're both easy to compute from a forward and backward pass on $\xclean$ and a forward pass on $\xnoise$, just like AtP itself.

Then, to complete the edge-AtP evaluation, what remains computationally is to evaluate all the dot products between nodes in different semilayers, at each token position. This requires $d_{\mathrm{resid}}\numTokens (1-\frac{1}{L})|\Nodes|^2/2$ multiplications in total\footnote{This formula omits edges within a single layer, for simplicity -- but those are a small minority.}, where $L$ is the number of layers, $\numTokens$ is the number of tokens, and $|\Nodes|$ is the total number of nodes. This cost exceeds the cost of computing all $\Delta r_{\node_1}^{\text{AtP}}(\xclean, \xnoise)$ and $\nabla_{r_{\node_2}}^{\text{AtP}}\metric(\model(\xclean))$ on Pythia 2.8B even with a single node per MLP layer; if we look at a larger model, or especially if we consider single-neuron nodes even for small models, the gap grows significantly.

Due to this observation, we'll focus our attention on the quadratic part of the compute cost, pertaining to two nodes rather than just one -- i.e.\ the number of multiplications in computing all $(\Delta r_{\node_1}^{\text{AtP}}(\xclean, \xnoise))^\intercal \nabla_{r_{\node_2}}^{\text{AtP}}\metric(\model(\xclean))$. Notably, we'll also exclude within-attention-head edges from the ``quadratic cost'': these edges, from some key, query, or value node to an attention output node can be handled by minor variations of the nodewise AtP or AtP* methods for the corresponding key, query, or value node.

\subsubsection{MLPs}

There are a couple of issues that can come up around the MLP nodes. One is that, similarly to the attention saturation issue described in \cref{sec:saturation}, the linear approximation to the MLP may be fairly bad in some cases, creating significant false negatives if $\node_2$ is an MLP node. Another issue is that if we use single-neuron nodes, then those are very numerous, making the $d_{\mathrm{resid}}$-dimensional dot product per edge quite costly.

\paragraph{MLP saturation and fix}

Just as clean activations that saturate the attention probability may have small gradients that lead to strongly underestimated effects, the same is true of the MLP nonlinearity. A similar fix is applicable: instead of using a linear approximation to the function from $\node_1$ to $\node_2$, we can linearly approximate the function from $\node_1$ to the preactivation $\node_{2,\text{pre}}$, and then recompute $\node_2$ using that, before multiplying by the gradient.

This kind of rearrangement, where the gradient-delta-activation dot product is computed in $d_{\node_2}$ dimensions rather than $d_{\mathrm{resid}}$, will come up again -- we'll call it the \emph{factored} form of AtP.

If the nodes are neurons then the factored form requires no change to the number of multiplications; however, if they're MLP layers then there's a large increase in cost, by a factor of $d_{\mathrm{neurons}}$. This increase is mitigated by two factors: one is that this is a small minority of edges, outnumbered by the number of edges ending in attention nodes by $3\times (\text{\# heads per layer})$; the other is the potential for parameter sharing.

\paragraph{Neuron edges and parameter sharing}

A useful observation is that each edge, across different token\footnote{Also across different batch entries, if we do this on more than one prompt pair.} positions, reuses the same parameter matrices in $\operatorname{Jac}_{\node_1}(r_{\text{out},{\ell_1}})(\node_1(\xclean))$ and $\operatorname{Jac}_{r_{\text{in},{\ell_2}}}(\node_2)(r_{\text{in},{\ell_2}}(\xclean))$. Indeed, setting aside the MLP activation function, the only other nonlinearity in those functions is a layer normalization; if we freeze the scaling factor at its clean value as in \cref{app:layernorm}, the Jacobians are equal to the product of the corresponding parameter matrices, divided by the clean scaling factor.

Thus if we premultiply the parameter matrices then we eliminate the need to do so at each token, which reduces the per-token quadratic cost by $d_{\mathrm{resid}}$ (i.e.\ to a scalar multiplication) for neuron-neuron edges, or by $d_{\mathrm{resid}}/d_{\mathrm{site}}$ (i.e.\ to a $d_{\mathrm{site}}$-dimensional dot product) for edges between neurons and some attention site.

It's worth noting, though, that these premultiplied parameter matrices (or, indeed, the edge-AtP estimates if we use neuron sites) will in total be many times (specifically, $(L-1)\frac{d_{\mathrm{neurons}}}{4d_{\mathrm{resid}}}$ times) larger than the MLP weights themselves, so storage may need to be considered carefully. It may be worth considering ways to only find the largest estimates, or the estimates over some threshold, rather than full estimates for all edges.

\subsubsection{Edge AtP* costs}\label{app:edge_atp_star}

Let's now consider how to adapt the AtP* proposals from \cref{sec:improvements} to this setting. We've already seen that the MLP fix, which is similarly motivated to the QK fix, has negligible cost in the neuron-nodes case, but comes with a $d_{\mathrm{neurons}}/d_{\mathrm{resid}}$ overhead in quadratic cost in the case of using an MLP layer per node, at least on edges into those MLP nodes. We'll consider the MLP fix to be part of edge-AtP*. Now let's investigate the two corrections in regular AtP*: GradDrops, and the QK fix.

\paragraph{GradDrops}
GradDrops works by replacing the single backward pass in the AtP formula with $L$ backward passes; this in effect means $L$ values for the multiplicand $\nabla_{r_{\node_2}}^{\text{AtP}}\metric(\model(\xclean))$, so this is a multiplicative factor of $L$ on the quadratic cost (though in fact some of these will be duplicates, and taking this into account lets us drive the multiplicative factor down to $(L+1)/2$). Notably this works equally well with ``factored AtP'', as used for neuron edges; and in particular, if $\node_2$ is a neuron, the gradients can easily be combined and shared across $\node_1$s, eliminating the $(L+1)/2$ quadratic-cost overhead.

However, the motivation for GradDrops was to account for multiple paths whose effects may cancel; in the edge-interventions setting, these can already be discovered in a different way (by identifying the responsible edges out of $\node_2$), so the benefit of GradDrops is lessened. At the same time, the cost remains substantial. Thus, we'll omit GradDrops from our recommended procedure edge-AtP*.

\paragraph{QK fix}

The QK fix applies to the $\nabla_{\node_2}(\metric(\model(\xclean)))(\node_2(\xclean))$ term, i.e.\ to replacing the linear approximation to the softmax with a correct calculation to the change in softmax, for each different input $\Delta r_{\node_1}^{\text{AtP}}(\xclean, \xnoise)$. As in \cref{sec:saturation}, there's the simpler case of accounting for $\node_2$s that are query nodes, and the more complicated case of $\node_2$s that are key nodes using \cref{alg:kfix} -- but these are both cheap to do after computing the $\Delta\attnLogits$ corresponding to $\node_2$.

The ``factored AtP'' way is to matrix-multiply $\Delta r_{\node_1}^{\text{AtP}}(\xclean, \xnoise)$ with key or query weights and with the clean queries or keys, respectively. This means instead of the $d_{\mathrm{resid}}$ multiplications required for each edge $\node_1\rightarrow \node_2$ with AtP, we need $d_{\mathrm{resid}}d_{\mathrm{key}} + \numTokens d_{\mathrm{key}}$ multiplications (which, thanks to the causal mask, can be reduced to an average of $d_{\mathrm{key}}(d_{\mathrm{resid}}+(\numTokens+1)/2)$).

The ``unfactored'' option is to stay in the $r_{\text{in},{\ell_2}}$ space: pre-multiply the clean queries or keys with the respective key or query weight matrices, and then take the dot product of $\Delta r_{\node_1}^{\text{AtP}}(\xclean, \xnoise)$ with each one. This way, the quadratic part of the compute cost contains $d_{\mathrm{resid}}(\numTokens+1)/2$ multiplications; this will be more efficient for short sequence lengths.

This means that for edges into key and query nodes, the overhead of doing AtP+QKfix on the quadratic cost is a multiplicative factor of $\min\left(\frac{\numTokens+1}{2}, d_{\mathrm{key}}\left(1+\frac{\numTokens+1}{2d_{\mathrm{resid}}}\right)\right)$.

\paragraph{QK fix + GradDrops}
If the QK fix is being combined with GradDrops, then the first multiplication by the $d_{\mathrm{resid}}\times d_{\mathrm{key}}$ matrix can be shared between the different gradients; so the overhead on the quadratic cost of QKfix + GradDrops for edges into queries and keys, using the factored method, is $d_{\mathrm{key}}\left(1+\frac{(\numTokens+1)(L+1)}{4d_{\mathrm{resid}}}\right)$.

\subsection{Conclusion}

\begin{figure}
    \centering
    \includegraphics[width=\textwidth]{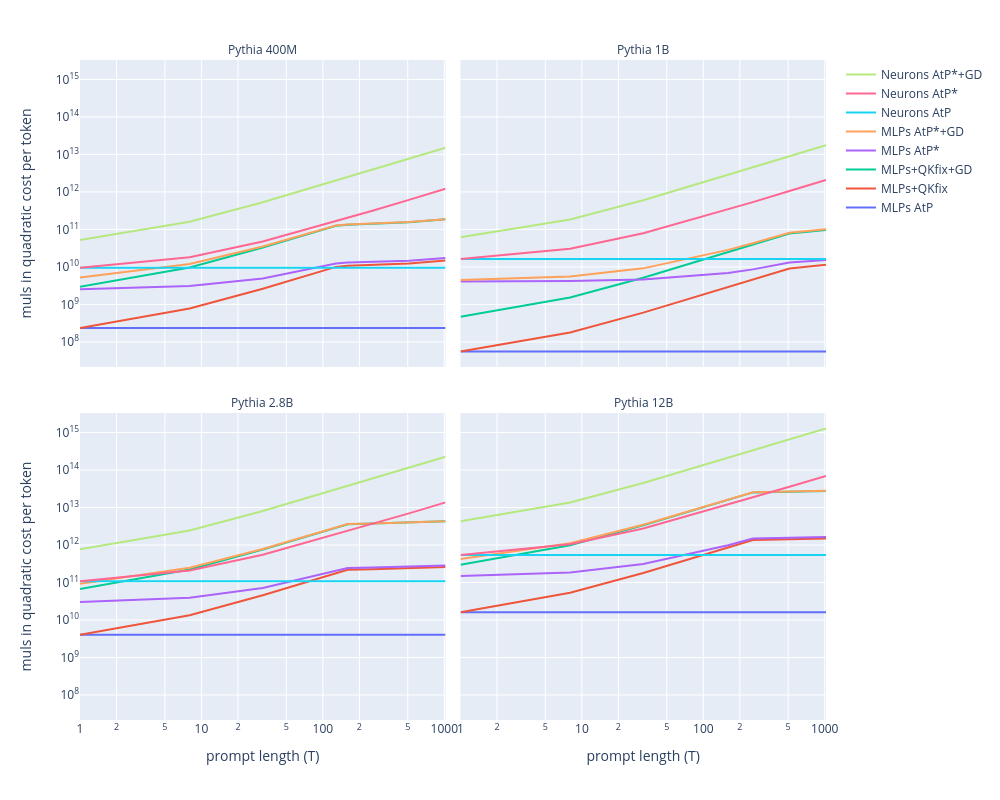}
    \caption{\small A comparison of edge-AtP variants across model sizes and prompt lengths. AtP* here is defined to include QKfix and MLPfix, but not GradDrops. The costs vary across several orders of magnitude for each setting.\\
    In the setting with full-MLP nodes, MLPfix carries substantial cost for short prompts, but barely matters for long prompts.\\
    In the neuron-nodes setting, MLPfix is costless. But GradDrops in that setting continues to impose a large cost; even though it doesn't affect MLP$\rightarrow$MLP edges, it does affect MLP$\rightarrow$Q,K edges, which come out dominating the cost with QKfix.}
    \label{fig:edge_atp_costs}
\end{figure}

Considering all the above possibilities, it's not obvious where the best tradeoff is between correctness and compute cost in all situations. In \Cref{tab:edge_costs} we provide formulas measuring the number of multiplications in the quadratic cost for each kind of edge, across the variations we've mentioned. In \Cref{fig:edge_atp_costs} we plug in the 4 sizes of Pythia model used elsewhere in the paper, such as \Cref{fig:model_sweep}, to enable numerical comparison.

\begin{table}[]
\small
    \centering
    \begin{tabular}{r|c|c|c|c|c|c}
    AtP variant& O$\rightarrow$V & O$\rightarrow$Q,K & O$\rightarrow$MLP & MLP$\rightarrow$V & MLP$\rightarrow$Q,K & MLP$\rightarrow$MLP\\
        \hline\hline
         \textbf{MLP layers} & $DH^2$ & $2DH^2$ & $DH$ & $DH$ & $2DH$ & $D$ \\
         QKfix&$DH^2$& $(T+1)DH^2$ &$DH$ & $DH$& $(T+1)DH$ & $D$ \\
         QKfix+GD&$\frac{L+1}{2}DH^2$& $\frac{(L+1)(T+1)}{2}DH^2$ &$\frac{L+1}{2}DH$ & $\frac{L+1}{2}DH$& $\frac{(L+1)(T+1)}{2}DH$ & $\frac{L+1}{2}D$ \\
         AtP*&$DH^2$& $(T+1)DH^2$ &$VNH$ & $DH$& $(T+1)DH$ & $ND$ \\
         AtP*+GD&$\frac{L+1}{2}DH^2$& $\frac{(L+1)(T+1)}{2}DH^2$ &$VNH$ & $\frac{L+1}{2}DH$& $\frac{(L+1)(T+1)}{2}DH$ & $ND$ \\
         \hline
         \textbf{QKfix (long)}&$DH^2$& $(2D+T+1)KH^2$ &$DH$ & $DH$& $(2D+T+1)KH$& $D$ \\
         QKfix+GD&$\frac{L+1}{2}DH^2$& $\frac{L+1}{2}(2D+T+1)KH^2$ &$\frac{L+1}{2}DH$ & $\frac{L+1}{2}DH$& $\frac{L+1}{2}(2D+T+1)KH$& $\frac{L+1}{2}D$ \\
         ATP*&$DH^2$& $(2D+T+1)KH^2$ &$VNH$ & $DH$& $(2D+T+1)KH$ & $ND$ \\
         AtP*+GD&$\frac{L+1}{2}DH^2$& $\frac{L+1}{2}(2D+T+1)KH^2$ &$VNH$ & $\frac{L+1}{2}DH$& $\frac{L+1}{2}(2D+T+1)KH$ & $ND$ \\
         \hline\hline
         \textbf{Neurons}&$DH^2$ & $2DH^2$ & $VNH$ & $VNH$ & $2KNH$ & $N^2$ \\
         MLPfix&$DH^2$ & $2DH^2$ & $VNH$ & $VNH$ & $2KNH$ & $N^2$ \\
         AtP*&$DH^2$ & $(T+1)DH^2$ & $VNH$ & $VNH$ & $(T+1)KNH$  & $N^2$ \\
         AtP*+GD&$\frac{L+1}{2}DH^2$ & $\frac{L+1}{2}(T+1)DH^2$ & $VNH$ & $\frac{L+1}{2}VNH$ & $\frac{(L+1)(T+1)}{2}KNH$  & $N^2$ \\
         \hline
         \textbf{ATP* (long)}&$DH^2$ & $(2D+T+1)KH^2$ & $VNH$ & $VNH$ & $(T+1)KNH$  & $N^2$ \\
         AtP*+GD&$\frac{L+1}{2}DH^2$ & $\frac{L+1}{2}(2D+T+1)KH^2$ & $VNH$ & $\frac{L+1}{2}VNH$ & $\frac{(L+1)(T+1)}{2}KNH$  & $N^2$ \\
    \end{tabular}
    \caption{\small Per-token per-layer-pair total quadratic cost of each kind of between-layers edge, across edge-AtP variants. For brevity, we omit the layer-pair $\binom{L}{2}$ factor that would otherwise be in every cell, and use $D:=d_{\mathrm{resid}},H:=\text{\# heads per layer},K:=d_{\mathrm{key}},V:=d_{\mathrm{value}},N:=d_{\mathrm{neurons}}$.}
    \label{tab:edge_costs}
\end{table}

\section{Distribution of true effects}\label{app:true_effect_distribution}

\begin{figure}
    \centering
    
    \caption{Distribution of true effects across models and prompt pair distributions}
    \begin{subfigure}[b]{\textwidth}
        \begin{subsubfigure}[b]{0.45\textwidth}
        \AttentionNodes{}
        \end{subsubfigure}
        \begin{subsubfigure}[b]{0.45\textwidth}
        \NeuronNodes{}
        \end{subsubfigure}
    \end{subfigure}
    \begin{subfigure}[b]{\textwidth}
        \setcounter{subfigure}{0}
        \setcounter{subsubfigure}{0}
        \begin{subsubfigure}[b]{0.45\textwidth}
            \includegraphics[width=\textwidth]{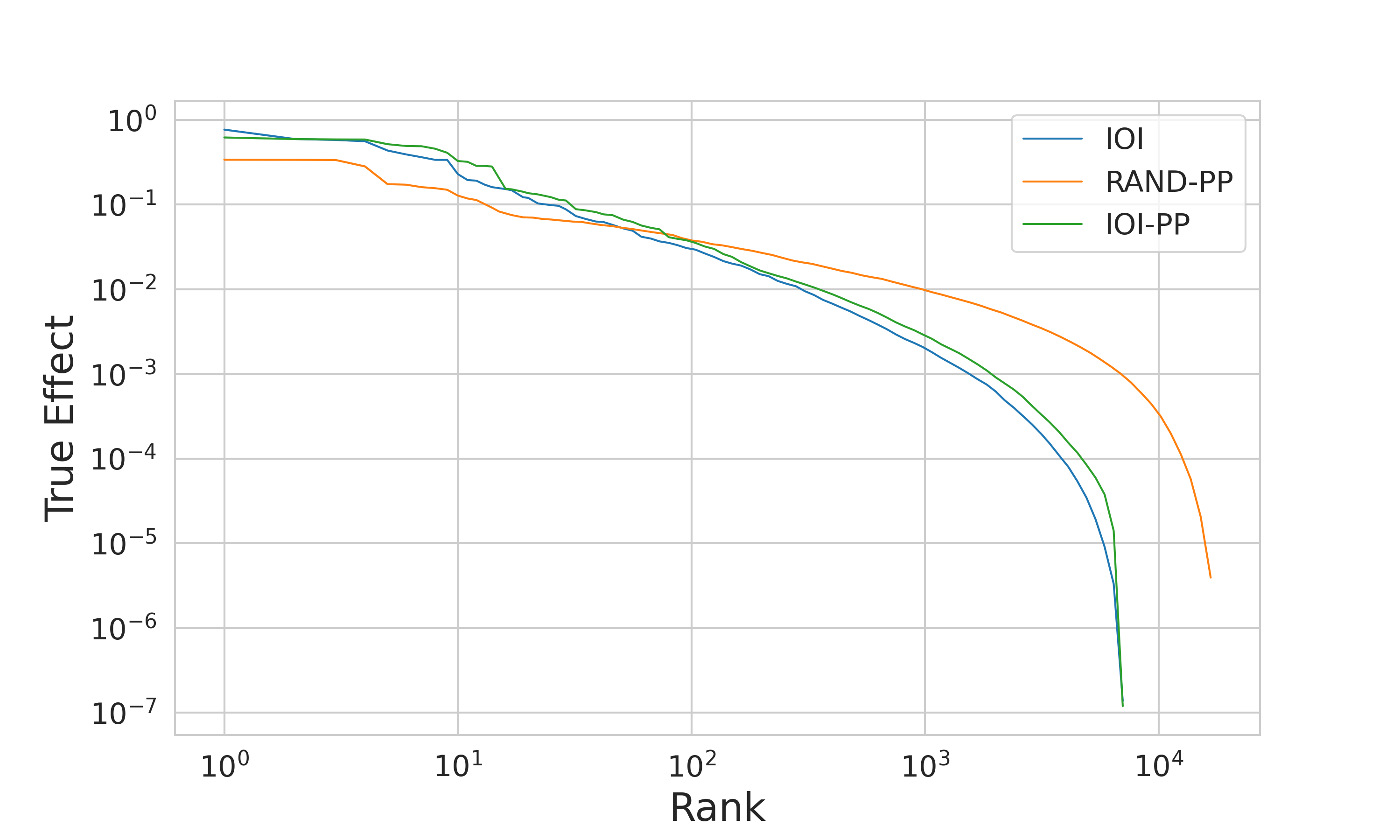}
            \caption{}
        \end{subsubfigure}
        \begin{subsubfigure}[b]{0.45\textwidth}
            \includegraphics[width=\textwidth]{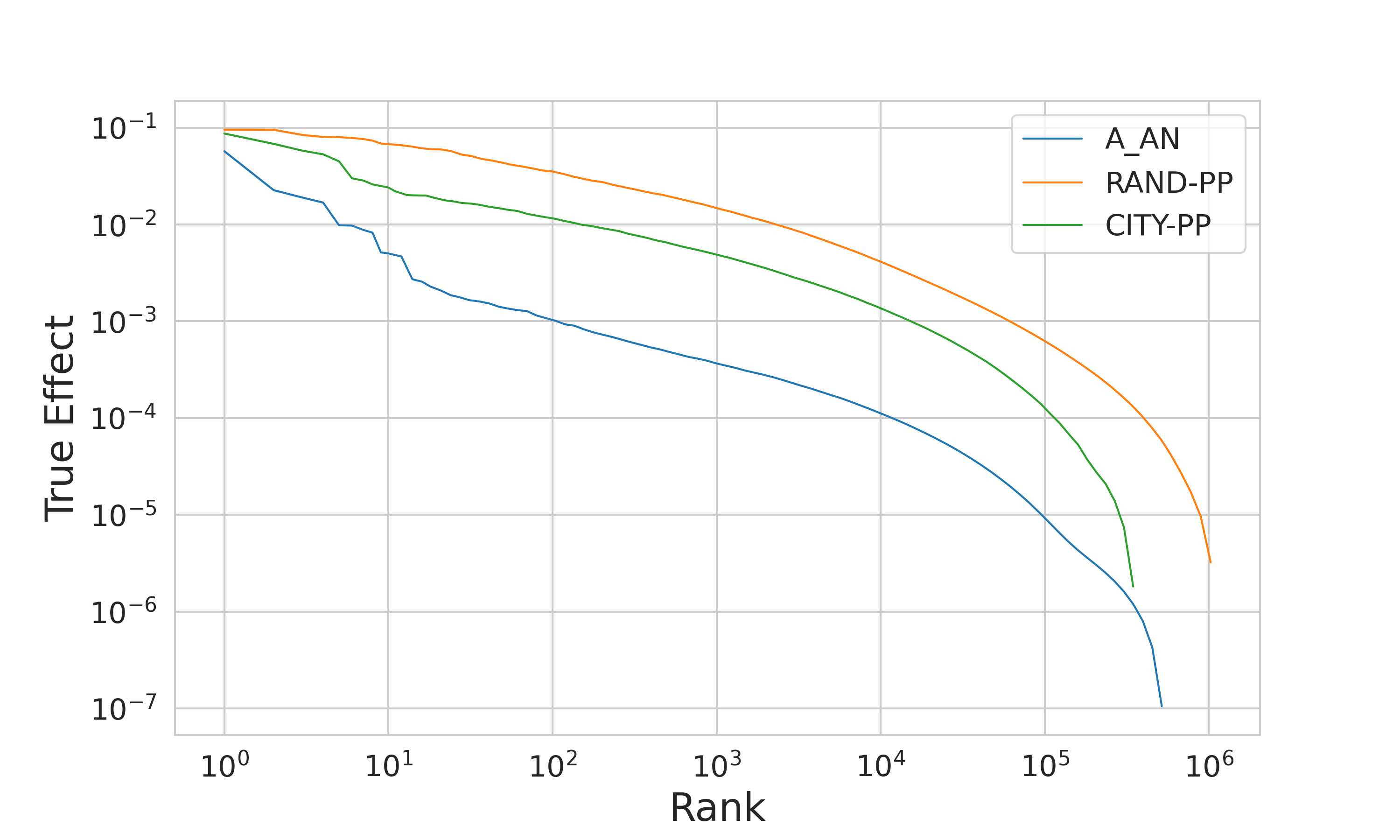}
            \caption{}
        \end{subsubfigure}
        \caption{Pythia-410M}
    \end{subfigure}
    \begin{subfigure}[b]{\textwidth}
        \setcounter{subsubfigure}{0}
        \begin{subsubfigure}[b]{0.45\textwidth}
            \includegraphics[width=\textwidth]{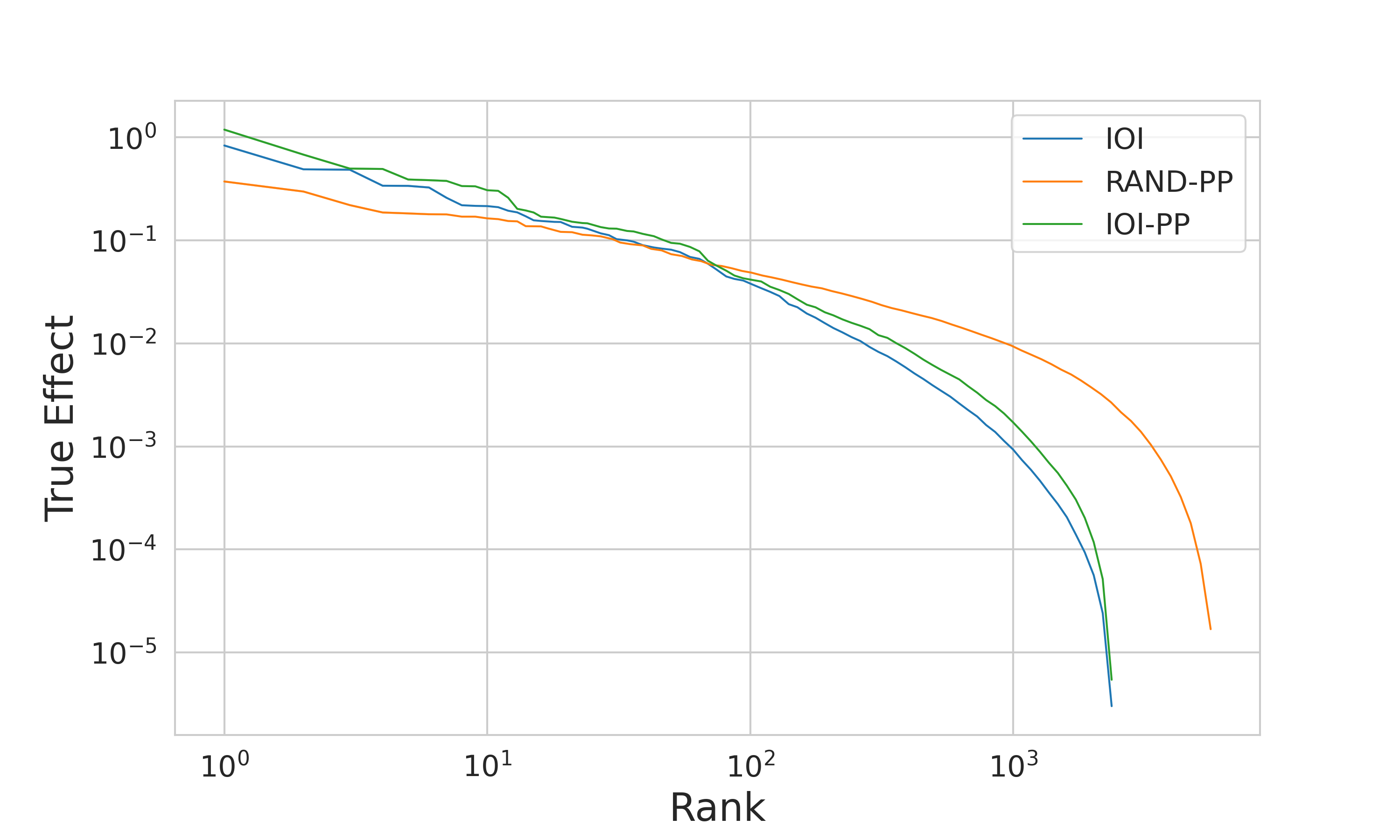}
            \caption{}
        \end{subsubfigure}
        \begin{subsubfigure}[b]{0.45\textwidth}
            \includegraphics[width=\textwidth]{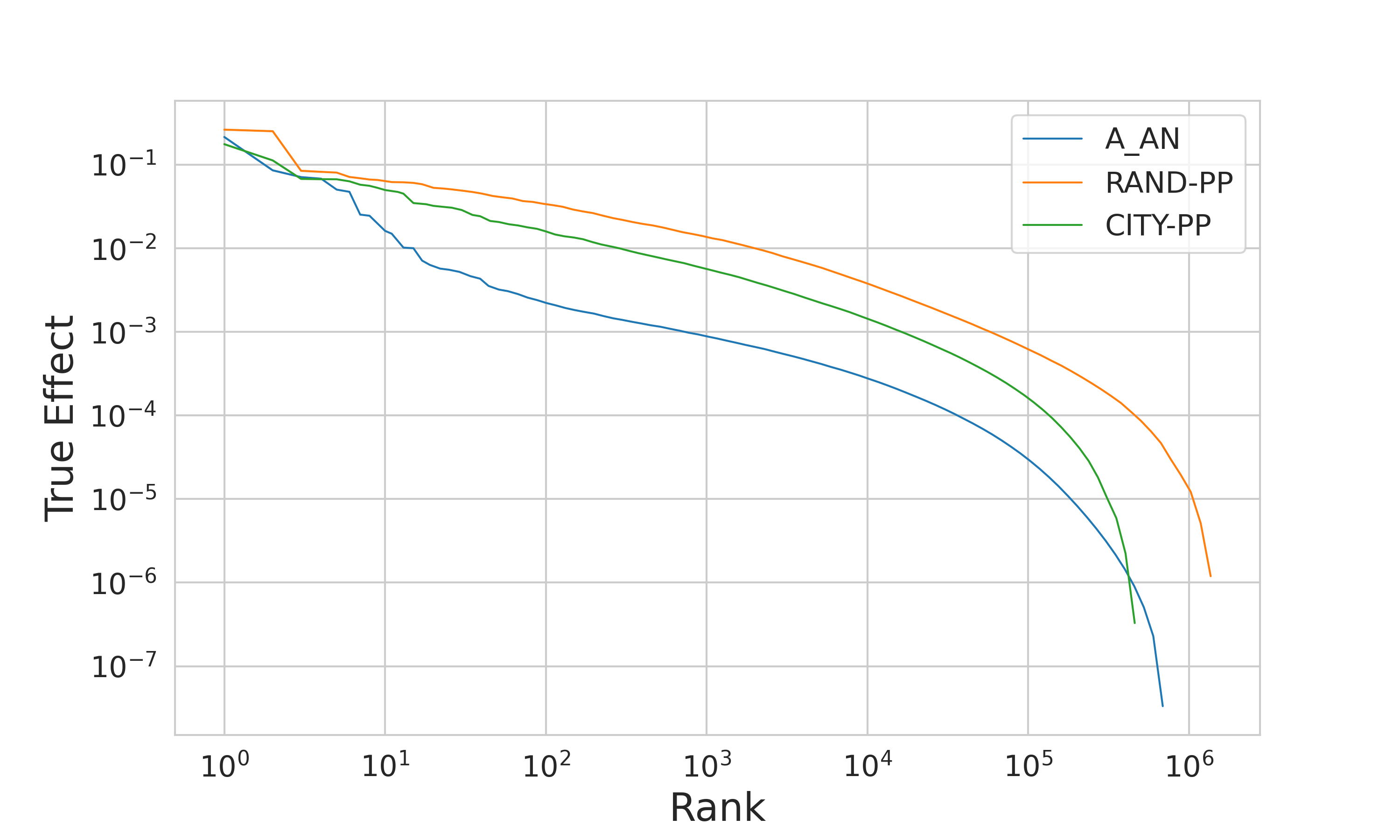}
            \caption{}
        \end{subsubfigure}
        \caption{Pythia-1B}
    \end{subfigure}
    \begin{subfigure}[b]{\textwidth}
        \setcounter{subsubfigure}{0}
        \begin{subsubfigure}[b]{0.45\textwidth}
            \includegraphics[width=\textwidth]{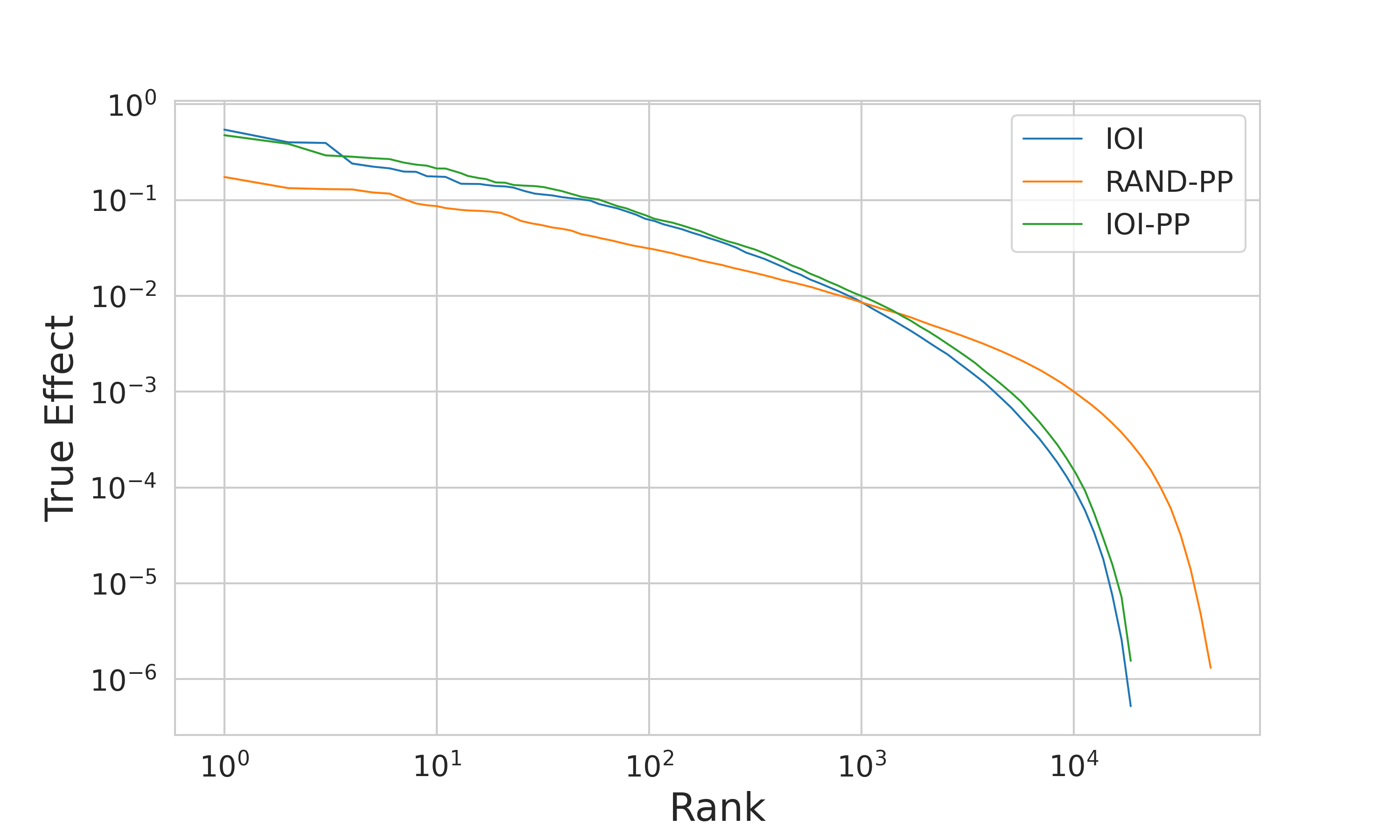}
            \caption{}
        \end{subsubfigure}
        \begin{subsubfigure}[b]{0.45\textwidth}
            \includegraphics[width=\textwidth]{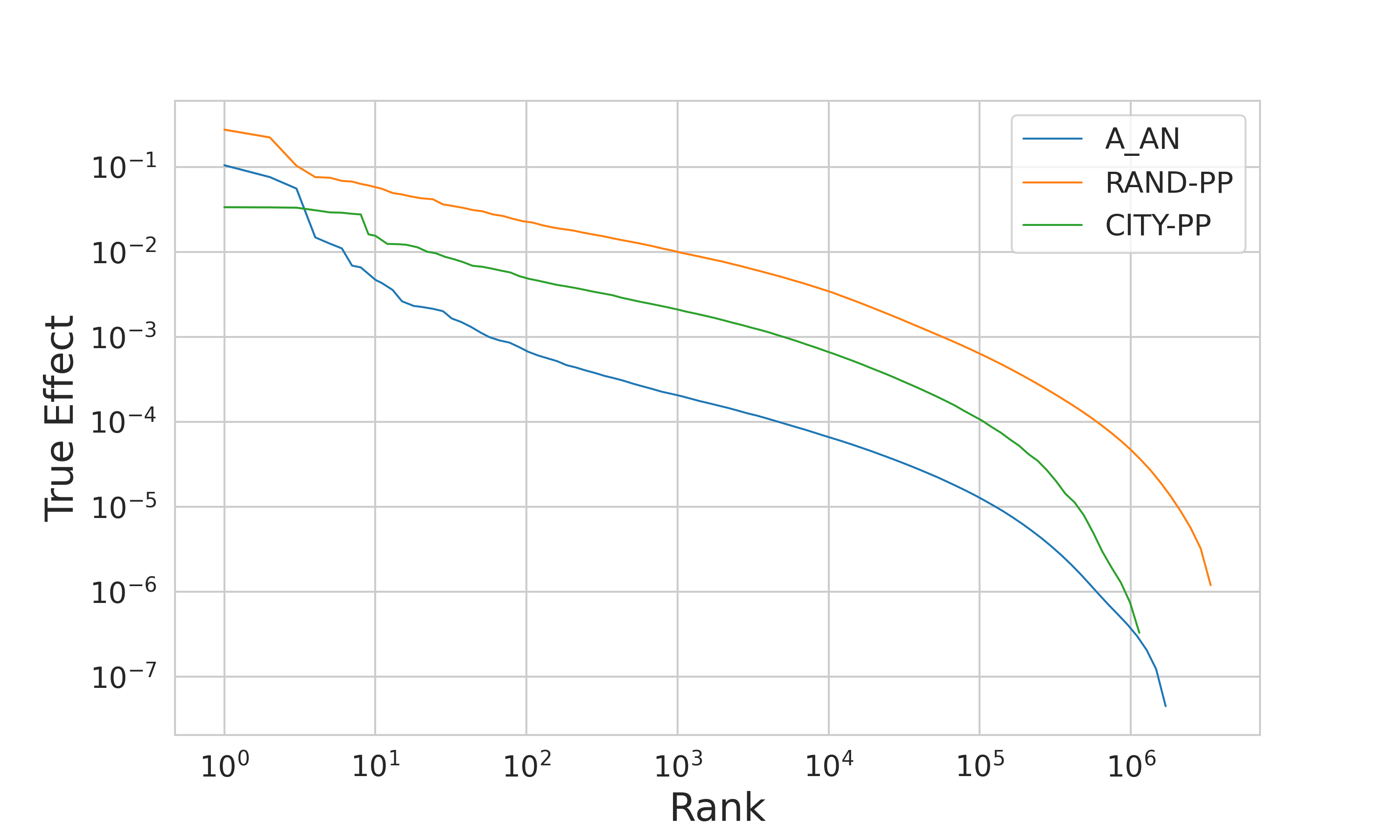}
            \caption{}
        \end{subsubfigure}
        \caption{Pythia-2.8B}
    \end{subfigure}
    \begin{subfigure}[b]{\textwidth}
        \setcounter{subsubfigure}{0}
        \begin{subsubfigure}[b]{0.45\textwidth}
            \includegraphics[width=\textwidth]{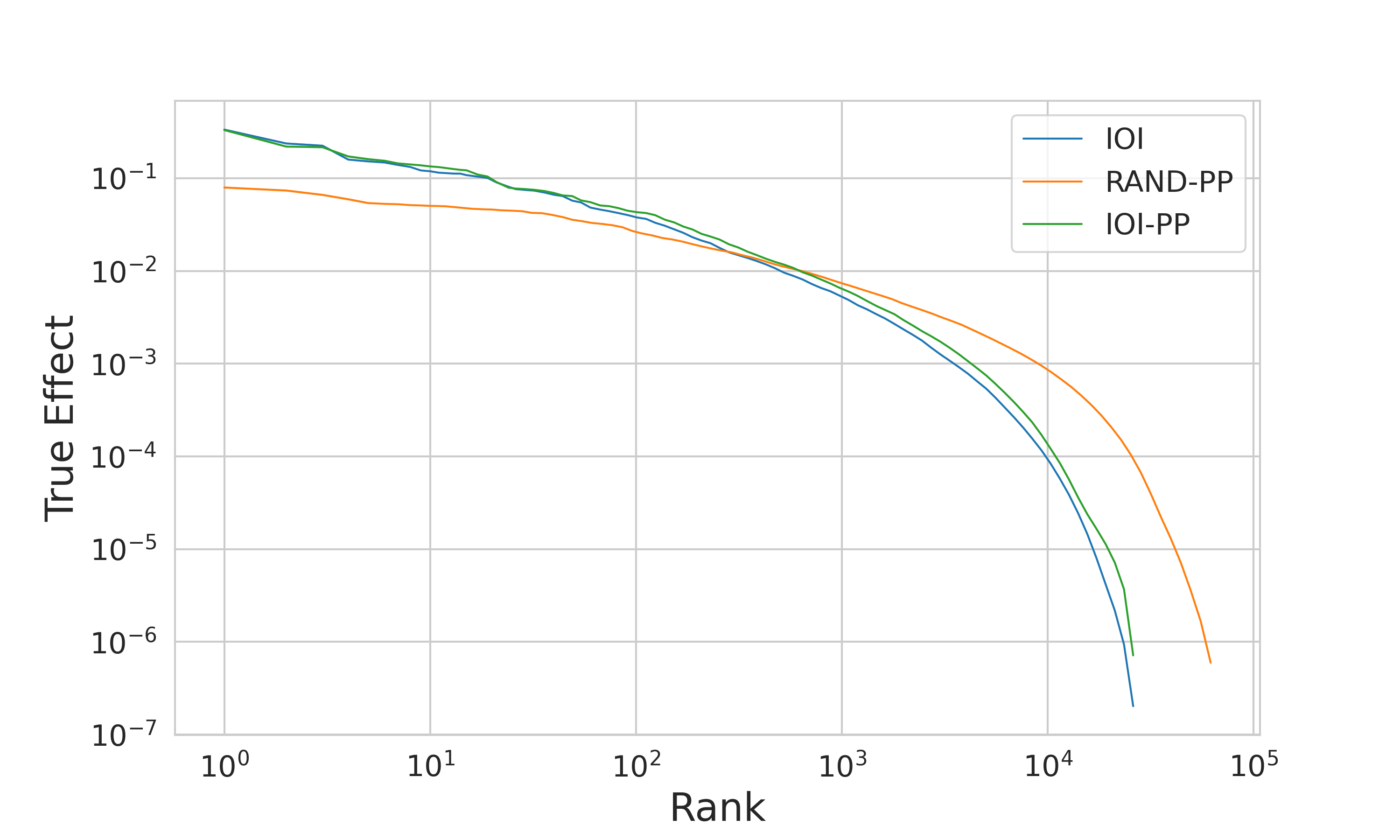}
            \caption{}
        \end{subsubfigure}
        \begin{subsubfigure}[b]{0.45\textwidth}
            \includegraphics[width=\textwidth]{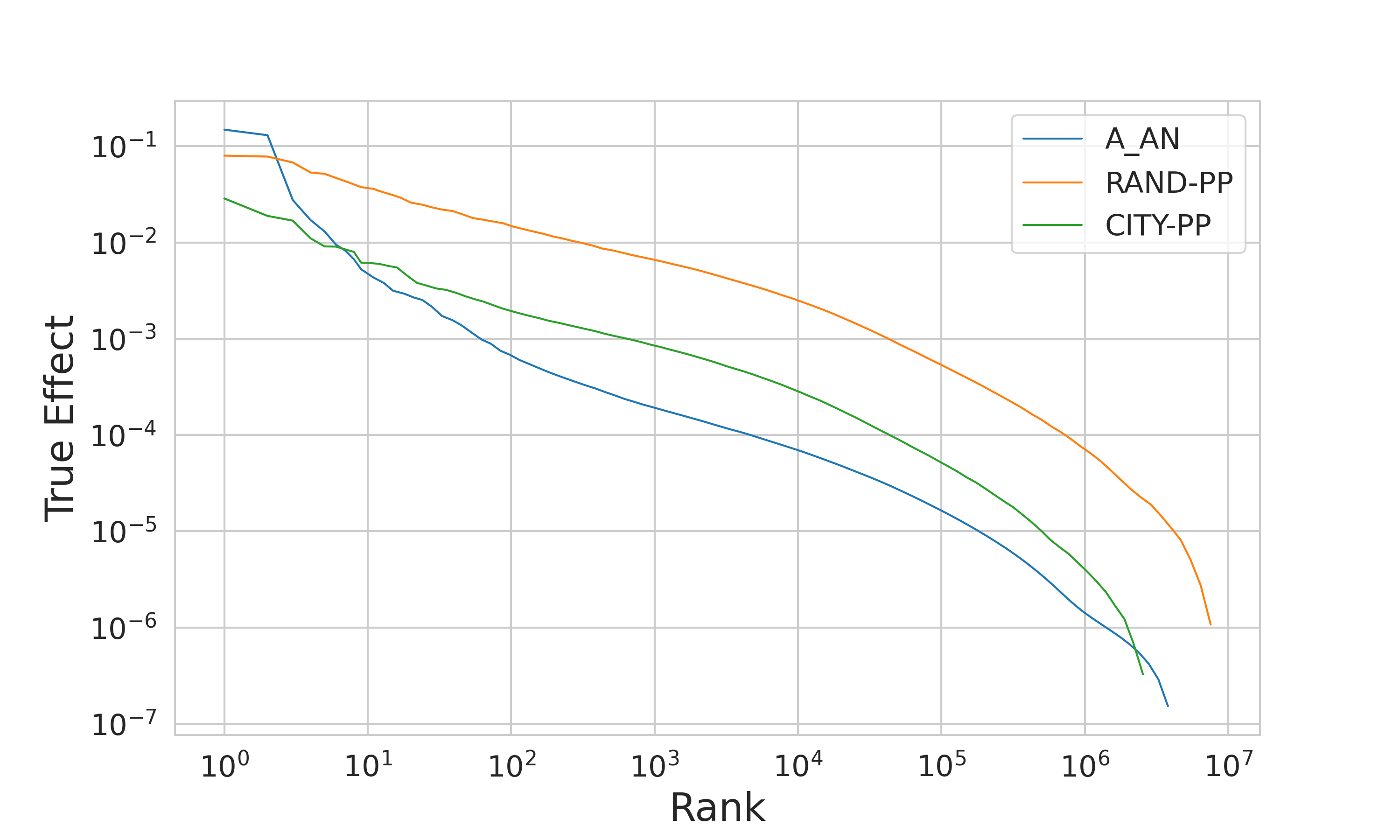}
            \caption{}
        \end{subsubfigure}
        \caption{Pythia-12B}
    \end{subfigure}
    \label{fig:true_effect_distribution}
\end{figure}

In~\Cref{fig:true_effect_distribution}, we show the distribution of $\contrib(\node)$ across models and distributions.

\end{document}